%% file: main.tex
\documentclass[10pt,journal,compsoc]{IEEEtran}

\ifCLASSOPTIONcompsoc
  \usepackage[nocompress]{cite}
\else
  \usepackage{cite}
\fi
\ifCLASSINFOpdf
\else
\fi
\pdfoutput=1

\usepackage{graphicx}
\usepackage[utf8]{inputenc} 
\usepackage[T1]{fontenc}    
\usepackage{url}            
\usepackage{booktabs}       
\usepackage{amsfonts}       
\usepackage{nicefrac}       
\usepackage{microtype}      
\usepackage{epsfig}
\usepackage{float}
\usepackage{multicol}
\usepackage{multirow}
\usepackage{amssymb}
\usepackage{amsmath}
\usepackage{wrapfig}
\usepackage{xspace}
\usepackage{color}
\usepackage{colortbl}
\usepackage[dvipsnames, svgnames, x11names]{xcolor}
\usepackage[misc]{ifsym}
\usepackage{makecell}
\usepackage{soul}
\usepackage{bm}
\usepackage{wrapfig}

\definecolor{linkcolor}{RGB}{255,0,0}
\definecolor{urlcolor}{RGB}{255,105,180}
\definecolor{citecolor}{RGB}{66,168,235}
\usepackage[pagebackref=true,breaklinks=true,colorlinks=true,bookmarks=false,linkcolor=linkcolor,urlcolor=urlcolor,citecolor=citecolor]{hyperref}

\usepackage{adjustbox}
\usepackage{pifont}
\usepackage{array}
\usepackage{enumitem}
\newcolumntype{C}[1]{>{\centering\arraybackslash}p{#1}} 

\definecolor{lightgray}{rgb}{0.8, 0.8, 0.8}
\definecolor{lgray}{rgb}{0.66, 0.66, 0.66}
\definecolor{whit_tab}{RGB}{255, 255, 255}
\definecolor{gray_tab}{RGB}{235, 235, 235}
\definecolor{oran_tab}{RGB}{254, 247, 241}
\definecolor{blue_tab}{RGB}{200, 227, 245}
\definecolor{lblu_tab}{RGB}{231, 239, 248}

\definecolor{ours}{RGB}{225, 235, 246}       
\definecolor{ourslight}{RGB}{241, 246, 252}  
\definecolor{oursmid}{RGB}{232, 241, 250}    
\definecolor{tab_others}{RGB}{247, 247, 247} 

\definecolor{gaincolor}{RGB}{40, 130, 95}    
\definecolor{warncolor}{RGB}{180, 105, 60}   
\newcommand{\default}[1]{\cellcolor{ourslight}\textbf{#1}}

\usepackage[ruled,vlined,linesnumbered]{algorithm2e}

\usepackage[capitalize]{cleveref}
\crefname{section}{Sec.}{Secs.}
\Crefname{section}{Section}{Sections}
\Crefname{table}{Table}{Tables}
\crefname{table}{Tab.}{Tabs.}

\newlength\savewidth

\renewcommand{\paragraph}[1]{\vspace{1.25mm}\noindent\textbf{#1}}

\newcommand{\fsours}[1]{\cellcolor{ourslight}#1}
\newcommand{\fsbest}[1]{\cellcolor{ours}#1}

\usepackage{xcolor}
\usepackage{colortbl}
\definecolor{ours}{RGB}{225, 235, 246}

\usepackage{rotating}

\colorlet{tab_ours}{ours!50} 
\definecolor{tab_others}{RGB}{235, 235, 235}

\begin{document}
%

\title{MambaADv2: Evolving Duality-enhanced \\ State Space Model  for \\ Unsupervised  Anomaly Detection}

%
%
%

\author{Xiaobin Hu,
        ~Haoyang He,
        ~Bo Yin,
        ~Yu He, 
        ~Lei Xie,
         ~Jiangning Zhang, \\
        ~Yu-Gang Jiang, ~\textit{Fellow}, IEEE, 
        ~Shuicheng Yan, ~\textit{Fellow}, IEEE
        
\IEEEcompsocitemizethanks{
\IEEEcompsocthanksitem Xiaobin Hu, Haoyang He, Lei Xie and Jiangning Zhang are with Zhejiang University, Hangzhou, China.
\IEEEcompsocthanksitem Xiaobin Hu, Bo Yin, and Shuicheng Yan are with the School of Computing, National University of Singapore, Singapore. 
\IEEEcompsocthanksitem Yu He is with College of Computing and Data Science, Nanyang Technological University, Singapore
\IEEEcompsocthanksitem Yu-Gang Jiang is with the Institute of Trustworthy Embodied AI, Fudan University University, Shanghai, China.
}
}

%
%

\markboth{IEEE TRANSACTIONS ON PATTERN ANALYSIS AND MACHINE INTELLIGENCE}
{Shell \MakeLowercase{\textit{et al.}}: Bare Advanced Demo of IEEEtran.cls for IEEE Computer Society Journals}
%



\input{secs/0_abstract.tex}

\maketitle

\IEEEdisplaynontitleabstractindextext

%
\IEEEpeerreviewmaketitle

\input{secs/1_introduction}

\input{secs/2_related_work}

\input{secs/3_method}

\input{secs/4_experiments}

\input{secs/5_conclusion}
\ifCLASSOPTIONcaptionsoff
  \newpage
\fi



{
\bibliographystyle{IEEEtran}
\bibliography{IEEEabrv,main}
}

\input{secs/6_appendix.tex}

%








\end{document}

%% file: secs/0_abstract.tex
\IEEEtitleabstractindextext{
\begin{abstract}
While recent advancements in anomaly detection have demonstrated the efficacy of CNN- and Transformer-based approaches, these architectures face inherent limitations: CNNs struggle to capture long-range dependencies, whereas Transformers suffer from quadratic computational complexity. Consequently, Mamba-based architectures have attracted considerable attention, as they successfully combine superior long-range dependency modeling with linear computational complexity. By critically rethinking the structural evolution across the Mamba lineage 1-3 series, this paper proposes MambaADv2, a framework tailored for multi-class unsupervised anomaly detection. MambaADv2 comprises a pre-trained encoder and a Mamba-inspired decoder, equipped with Duality-enhanced State Space (DSS) modules across multiple scales. 
The proposed DSS module effectively models both global dependencies and local representations by integrating parallel-cascaded Hybrid State Space (HSS) blocks and frequency-enhanced convolution operations. 
The structure of the Hybrid State Space (HSS) block is tailored by following the SSD-based Mamba lineage and incorporating Mamba3-style position-aware state-space modeling, leveraging the dual computational paths of linear recurrence and parallel matrix formulation to model local continuity and global contextual comparison, thereby better serving the core anomaly detection objective of precisely reconstructing normal representations while magnifying anomalous deviations.
Additionally, we propose a semantics-adaptive progressive scanning strategy that decays scanning complexity along the feature pyramid. This design is inspired by our observation that optimal Hilbert scanning heavily depends on feature semantics: while shallow features require diverse directions, excessive scanning dilutes deep abstract features. This alignment of scanning with hierarchical semantics improves both accuracy and efficiency. Extensive evaluations across six diverse anomaly detection datasets and seven metrics yield state-of-the-art results, confirming the efficacy of the proposed approach. 





\end{abstract}

}

%% file: secs/1_introduction.tex
\section{Introduction} \label{section:intro}
The advent of smart manufacturing has significantly underscored the importance of visual Anomaly Detection (AD) in industrial production~\cite{pang2021deep,bergmann2022mvtec,bergmann2021mvtec,zou2022spot,wang2024real}. This technology holds the promise of enhancing efficiency, reducing manual inspection costs, and bolstering both product quality and production line stability. Currently, the majority of existing methods rely heavily on a single-class paradigm~\cite{deng2022anomaly,liu2023simplenet,zhang2023destseg,batzner2024efficientad}, necessitating a distinct model for each class and thereby incurring substantial computational and memory overheads. While recent advancements have introduced multi-class AD techniques~\cite{you2022unified,he2024diffusion,zhang2023exploring,invad,zhao2025unimmad}, there remains considerable room for improvement, particularly regarding the optimal trade-off between accuracy and efficiency.

\begin{figure*}[t!]
\centering
\includegraphics[width=\textwidth]{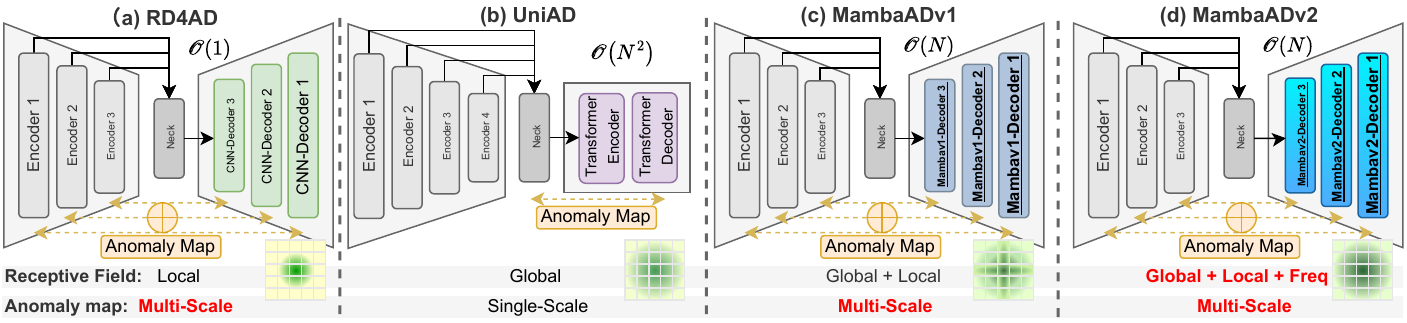}
\caption{Compared with (a) local CNN-based RD4AD~\cite{deng2022anomaly} and (b) global Transformer-based UniAD~\cite{you2022unified}, and (c) the preliminary MambaAD~\cite{he2024mambaad}, our proposed MambaADv2 maintains linear complexity while comprehensively integrating global, local, and frequency information. This expanded receptive field, along with multi-scale features, improves anomaly localization and multi-class detection performance.}
\label{fig:motivation}
\end{figure*}

Contemporary unsupervised anomaly detection (AD) algorithms predominantly fall into three paradigms~\cite{zhang2023exploring,cao2024survey,zhang2024ader}: \textit{Embedding-based}~\cite{roth2022towards, defard2021padim, bergmann2020uninformed, deng2022anomaly,cao2024bias}, \textit{Synthesizing-based}~\cite{zavrtanik2021draem, li2021cutpaste, zhang2023destseg, liu2023simplenet}, and \textit{Reconstruction-based}~\cite{liang2023omni,he2024diffusion,cao2022informative}. Although synthesizing and embedding-based methods yield promising results, they are typically hindered by intricate heuristic designs and rigid architectures. Conversely, reconstruction-based approaches, notably RD4AD~\cite{deng2022anomaly} and UniAD~\cite{you2022unified}, exhibit superior scalability and performance. As illustrated in \cref{fig:motivation} (a), RD4AD leverages a pre-trained teacher-student framework to compare features across multiple scales. However, inherently constrained by its CNN backbone, \textit{it struggles to capture long-range spatial dependencies.} UniAD, pioneering multi-class AD, adopts a transformer-based encoder-decoder architecture~\cite{vaswani2017attention,dosovitskiy2020image} (\cref{fig:motivation} (b)). While adept at global contextual modeling, \textit{transformers suffer from quadratic computational complexity. Consequently, UniAD is restricted to operating on the lowest-resolution feature maps, inevitably compromising its detection efficacy.}

Recently, state space models, notably Mamba~\cite{gu2023mamba,dao2024transformers}, have achieved remarkable success in large language models by matching the representational power of transformers while operating with linear computational complexity. This breakthrough has catalyzed a rapid wave of adaptations within the visual domain~\cite{liu2024vmamba, zhu2024vision, shi2024vmambair, huang2024localmamba, ruan2024vm, wang2024mamba, li2024mamba, hatamizadeh2025mambavision}. 
Motivated by these advancements, preliminary work MambaAD~\cite{he2024mambaad} attempts the integration of Mamba into visual anomaly detection, as illustrated in \cref{fig:motivation} (c). 
{By unifying global contextual awareness with local feature extraction, MambaAD capitalizes on its linear complexity to efficiently compute high-resolution anomaly maps across multiple scales.} 
\textit{However, such an early-stage design suffers from restricted representational capacity and suboptimal efficiency when handling dense, multi-class visual features.}

\begin{figure*}[t]
\centering
\includegraphics[width=\textwidth]{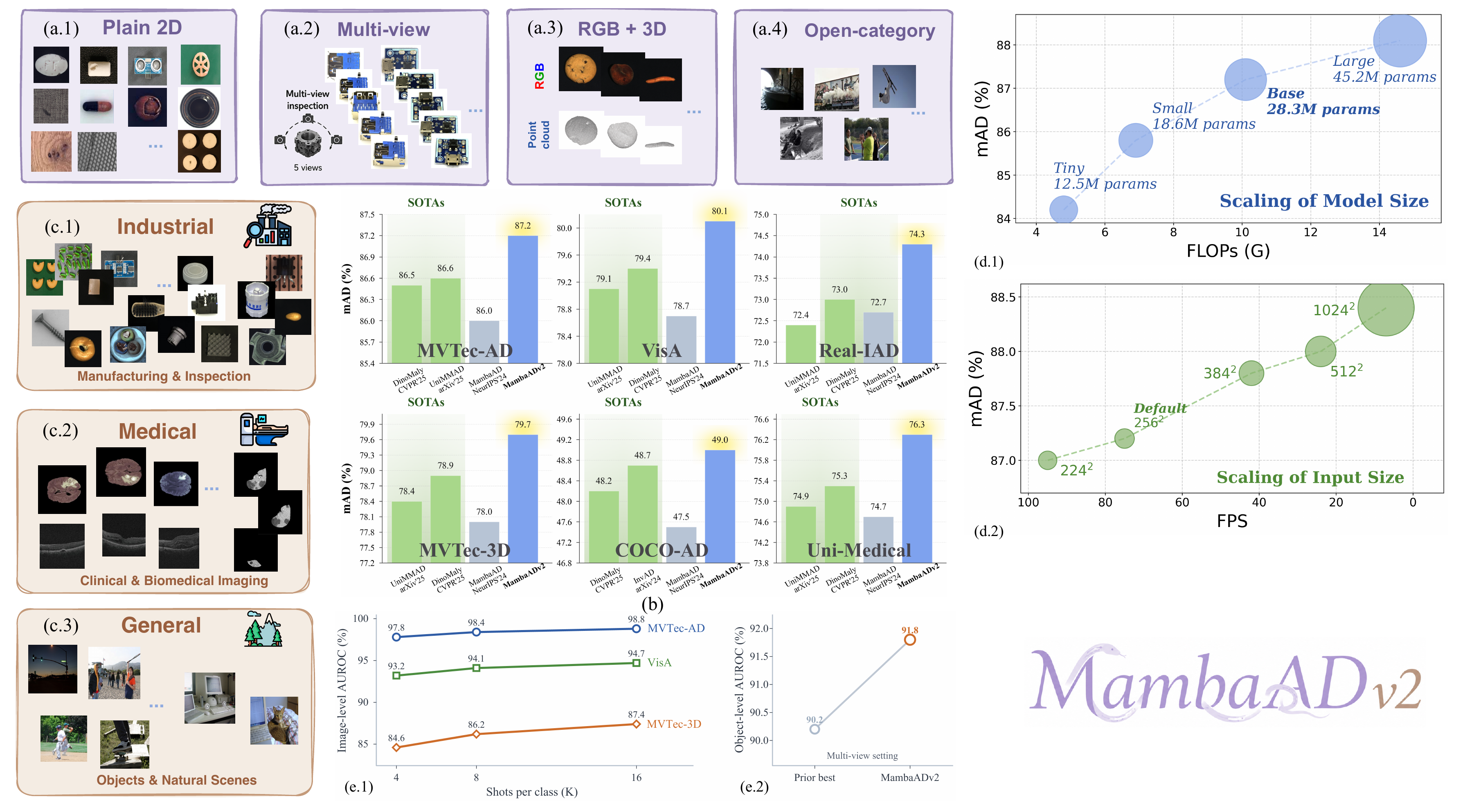}
\caption{Overview: modalities, benchmarks, and scalability of MambaADv2.
(a) Multi-modal capabilities span 2D images, multi-view 2D inputs, RGB+3D point cloud, and Cross-domain.
(b) MambaADv2 achieves strong performance across multiple benchmarks, including MVTec-AD, VisA, Real-IAD, MVTec-3D, COCO-AD, and Uni-Medical, surpassing corresponding state-of-the-art methods.
(c) Cross-domain validation spans industrial manufacturing and inspection, clinical and biomedical imaging, and general objects and natural scenes.
(d) Scalability analysis shows consistent improvements with larger model capacity and higher input resolution.
(e) Additional evaluations demonstrate robust few-shot transfer and multi-view anomaly detection performance compared with prior SOTAs.}
\label{fig:overview}
\end{figure*}

To address the limitations, 
we rethink the evolution characteristics of Mamba series
and then propose MambaADv2 as shown in \cref{fig:motivation} (d), a novel and powerful framework comprising a pre-trained encoder and a Mamba-inspired decoder. The core of our decoder is equipped with the proposed Duality-enhanced State Space (DSS) modules across multiple scales, which elegantly model both global dependencies and local representations. 
Specifically, our methodology introduces three pivotal advancements: First, we systematically evolve the core sequence modeling unit to Mamba-3, stepping beyond the preliminary explorations of Mamba-1 in visual anomaly detection. Rather than a simple architectural substitution, the internal structure of the Hybrid State Space (HSS) block is tailored by following the SSD-based Mamba lineage~\cite{dao2024transformers, lahoti2026mamba} and incorporating Mamba-3-style position-aware state-space modeling. This tailored design leverages the dual computational paths of linear recurrence and parallel matrix formulation to model local continuity and global contextual comparison. Such a design better serves the core objective of anomaly detection, precisely reconstructing normal representations while magnifying anomalous deviations. Alongside this structural evolution, we incorporate Rotary Position Embedding (RoPE)~\cite{su2024roformer} and provide comprehensive ablations on critical SSM design choices to establish effective practices.

Second, we revisit local feature modeling by introducing a spatial-frequency joint enhancement. Previous local branches typically relied on standard depthwise convolutions, which exclusively model local patterns in the spatial domain and overlook discriminative frequency cues. To overcome this, the proposed DSS module integrates parallel-cascaded HSS blocks with frequency-enhanced convolution operations, advancing the module to a robust "\textit{spatial-frequency-sequence}" architecture. Specifically, we replace standard convolutions with Wavelet Convolutions~\cite{mallat2002theory,finder2024wavelet}, which endow the feature extraction process with multi-resolution spatial-frequency coupled analysis capabilities without introducing noticeable computational overhead. 
Furthermore, we incorporate a frequency-domain attention mechanism
to explicitly model the periodic structures and frequency distribution patterns. This dual enhancement significantly bolsters the model's capacity to distinguish both high-frequency textural perturbations and low-frequency structural anomalies.

Finally, we propose a semantics-adaptive progressive scanning strategy. While previous SSM-based vision models predominantly employ a fixed, computationally heavy scanning configuration (\textit{e.g.,} eight-direction Hilbert scanning~\cite{hilbert1935stetige,he2024mambaad}) across all layers, we reveal a critical insight: \textit{the optimal number of scanning directions is intrinsically coupled with the semantic hierarchy of features~\cite{lin2017feature}}. Shallow features are spatially rich and direction-sensitive, demanding diverse scanning directions for complementary spatial perspectives. Conversely, highly abstract deep features suffer from representational redundancy, where excessive scanning inadvertently weakens effective semantic signals. Motivated by this, our strategy dynamically decays scanning complexity along the feature pyramid. This approach pioneers the alignment of SSM scanning configurations with hierarchical feature semantics, achieving an optimal paradigm that improves detection accuracy while substantially reducing computational costs.

By calculating and aggregating anomaly maps of Duality-enhanced State Space across multiple feature scales, MambaADv2 achieves state-of-the-art (SoTA) performance on six representative AD datasets. Notably, it delivers superior results across seven distinct image- and pixel-level metrics while maintaining a low parameter count and minimal computational overhead. This is an extension of the previous conference version
(MambaAD~\cite{he2024mambaad} in NeurIPS’24). In the conference paper, we mainly
introduce MambaAD, a multi-class unsupervised anomaly detection method based on the Hybrid State Space of Mamba-1, which outperforms
the state-of-the-art diverse anomaly detection baselines. 
As shown in Fig.~\ref{fig:overview}, in this extended journal version, 
we make the following four contributions:

\begin{itemize}
\item We propose MambaADv2, which inherits the capability of State Space Duality~\cite{dao2024transformers} with Rotary Position Embedding (RoPE)~\cite{su2024roformer} to significantly enhance representational capacity and preserve crucial spatial structures, overcoming MambaADv1's structural limitations of restricted multi-class modeling, suboptimal training parallelism, and severe spatial information loss.
\item We design a spatial-frequency joint enhancement for local feature modeling, which explicitly captures multi-resolution frequency patterns and structural periodicity~\cite{mallat2002theory,finder2024wavelet}, resolving MambaADv1's critical perception blind spot caused by relying exclusively on spatial-domain depthwise convolutions that ignore frequency-level anomaly cues.
\item We present a semantics-adaptive progressive scanning strategy, which dynamically aligns the scanning complexity with the feature pyramid's semantic hierarchy, addressing MambaADv1's representational redundancy and semantic interference caused by applying a rigid, fixed eight-direction scanning configuration uniformly across all abstract layers.
\item We demonstrate the exceptional performance and efficiency of MambaADv2 in multi-class anomaly detection (AD). Our approach establishes new state-of-the-art (SoTA) results on \textit{\textbf{six}} distinct AD datasets across \textit{\textbf{seven}} evaluation metrics, all while maintaining a remarkably low parameter count and minimal computational overhead.
\end{itemize}


%% file: secs/2_related_work.tex
\section{Related Work} \label{section:related}
\noindent\textbf{Unsupervised Anomaly Detection.} 
Existing AD methods can be broadly categorized into three main paradigms:

\noindent \textbf{\textit{1) Embedding-based methods}} extract multi-channel feature representations from RGB images~\cite{roth2022towards, defard2021padim, bergmann2020uninformed, deng2022anomaly, cohen2020sub, rudolph2021same, gudovskiy2022cflow, yu2021fastflow, zhou2024msflow}. These approaches typically rely on networks pre-trained on ImageNet~\cite{deng2009imagenet}. For instance, PatchCore~\cite{roth2022towards} extracts nominal patch features and stores them in a memory bank to measure anomaly scores via Mahalanobis distance. Alternatively, approaches like~\cite{bergmann2020uninformed, wang2021student, salehi2021multiresolution, tien2023revisiting} employ a student-teacher framework where student networks are trained to regress the outputs of a teacher network. Other variants adapt pre-trained features to the one-class target distribution~\cite{reiss2021panda, lee2022cfa, hyun2024reconpatch}, formulate patch-level one-class objectives~\cite{ruff2018deep, yi2020patch}, incorporate position and neighborhood statistics~\cite{bae2023pni}, or pursue latency-oriented deployment~\cite{batzner2024efficientad}. However, a significant distribution shift often exists between the natural datasets used for pre-training and industrial target images.

\noindent \textbf{\textit{2) Synthesizing-based methods}} generate pseudo-anomalies on normal samples~\cite{zavrtanik2021draem, li2021cutpaste, schluter2022natural, anomalydiffusion, chen2024unified}. For example, DRAEM~\cite{zavrtanik2021draem} generates pseudo-anomalies using Perlin noise and external texture images. Its architecture consists of a joint reconstruction and discriminative network that outputs an anomaly mask. Recent synthesis strategies further expand anomaly diversity with naturalistic perturbations, diffusion priors, feature-space generation, and gradient-guided global-local synthesis~\cite{schluter2022natural, anomalydiffusion, zavrtanik2022dsr, chen2024unified}. Despite their competitive performance, a noticeable domain gap between these synthesized artifacts and real-world anomalies remains a fundamental limitation.

\noindent \textbf{\textit{3) Reconstruction-based methods}}~\cite{deng2022anomaly,liang2023omni,invad,cao2023collaborative} focus on training auto-encoding architectures to reconstruct normal images, thereby mitigating the reliance on pre-trained weights. Various architectures, including Autoencoders~\cite{bergmann2018improving,zavrtanik2021reconstruction,ristea2022self,gong2019memorizing}, Transformers~\cite{pirnay2022inpainting}, Generative Adversarial Networks (GANs)~\cite{schlegl2019f,akcay2018ganomaly,liang2023omni, yan2021learning}, and diffusion models~\cite{he2024diffusion,wyatt2022anoddpm}, have been adopted as backbones. Memory-augmented designs also attempt to restrict reconstruction to normal patterns~\cite{gong2019memorizing}. However, the excessive generalization capacity of these models can inadvertently allow them to reconstruct anomalous regions, leading to inaccurate defect localization.

\noindent\textbf{Multi-class Anomaly Detection.} Most AD methods are predominantly trained on individual categories. As the number of categories scales, this one-model-per-class paradigm incurs prohibitive time and memory costs and struggles to handle large intra-class diversity. To overcome these limitations, multi-class unsupervised anomaly detection (MUAD) has recently garnered significant attention. UniAD~\cite{you2022unified} pioneers a unified reconstruction framework capable of handling multiple classes simultaneously. DiAD~\cite{he2024diffusion} introduces a diffusion-based framework featuring a semantic-guided network to preserve the semantic consistency of reconstructed images. Furthermore, ViTAD~\cite{zhang2023exploring} explores the efficacy of vanilla Vision Transformers (ViTs) in the MUAD setting. Hierarchical vector-quantized reconstruction~\cite{lu2023hierarchical}, feature inversion under the COCO-AD benchmark~\cite{invad}, prompt-based adaptation~\cite{li2024promptad}, and MoE-driven decompression~\cite{zhao2025unimmad} further broaden the design space, but efficient global-local modeling for high-resolution localization remains underexplored.

\noindent\textbf{State Space Models.} State Space Models (SSMs)~\cite{gu2021combining, gu2021efficiently, smith2022simplified, mehta2022long, fu2022hungry} have garnered considerable attention for their exceptional efficacy in long-sequence modeling. Notably, the Structured State Space sequence model (S4)~\cite{gu2021efficiently} effectively captures long-range dependencies (LRDs) via diagonal structural parameterization, thereby alleviating the computational bottlenecks inherent in prior works. Early multidimensional extensions such as S4ND~\cite{nguyen2022s4nd} and Mamba-ND~\cite{li2024mamba} also demonstrated the promise of applying SSMs to images and videos. Building upon the S4 framework, several variants have emerged, including S5~\cite{smith2022simplified}, H3~\cite{fu2022hungry}, and most prominently, Mamba~\cite{gu2023mamba}. By integrating a data-dependent selection mechanism, Mamba establishes a novel paradigm distinct from conventional CNNs and Transformers, achieving robust context modeling while preserving strictly linear scalability for processing long sequences. 

The immense potential of Mamba has catalyzed a surge of pioneering research within the visual domain~\cite{liu2024vmamba, zhu2024vision, ruan2024vm, hu2024zigma, huang2024localmamba, shi2024vmambair, wu2024h, wang2024mamba, DGmamba, zhang2024pointmamba, OMGSeg,hatamizadeh2025mambavision, yang2024plainmamba, yu2025mambaout}. For instance, VMamba~\cite{liu2024vmamba} introduces a Cross-Scan Module (CSM) to bridge the gap between non-causal 2D spatial images and ordered 1D sequences, effectively mitigating direction sensitivity. Furthermore, Mamba has been extensively adopted in medical image segmentation~\cite{ruan2024vm, liu2024swin, wu2024h, wang2024mamba, li2023transformer, ma2024u, xing2401segmamba}, seamlessly embedding Mamba blocks into UNet-like architectures to formulate task-specific networks. Video-oriented Mamba models further verify the scalability of SSMs for long spatiotemporal sequences~\cite{li2024videomamba, chen2024video, yang2024vivim}. Beyond standalone vision tasks, VL-Mamba~\cite{qiao2024vl} and Cobra~\cite{zhao2024cobra} delve into the potential of SSMs for Multimodal Large Language Models (MLLMs). Additionally, ZigMa~\cite{hu2024zigma} addresses spatial continuity challenges during the scanning process by incorporating Mamba into a Stochastic Interpolation framework~\cite{albergo2023stochastic}. MambaAD~\cite{he2024mambaad} exploits the Mamba-1's long-range modeling capacity via combining SSM's structure with CNNs' modeling prowess for multi-class unsupervised anomaly detection. 

In contrast, MambaADv2 further investigates the evolution from Mamba-1 to Mamba-3-style state-space modeling and proposes Duality-enhanced State Space (DSS) modules by integrating parallel-cascaded Hybrid State Space (HSS) blocks and frequency-enhanced convolution operations. This design better serves the core objective of anomaly detection: precisely reconstructing normal representations while magnifying anomalous deviations.

%% file: secs/3_method.tex
\section{Methodology}
\label{section:method}

This section presents MambaADv2. We begin with the mathematical foundations of SSMs and their evolution to Mamba-3 (\cref{sec:prelim}), describe the overall framework (\cref{sec:framework}), and then detail each proposed component: the Locality-Enhanced State Space Block (\cref{sec:lss}), the redesigned Hybrid State Space Block (\cref{sec:hss}), the Inception Mixer (\cref{sec:fim}), and the Semantics-Adaptive Progressive Scanning strategy (\cref{sec:scan}).

\subsection{Preliminaries}
\label{sec:prelim}

\noindent\textbf{State Space Models.}
Rooted in control theory, State Space Models (SSMs)~\cite{gu2021combining, gu2021efficiently} define a mapping from a one-dimensional input sequence $x(t) \in \mathbb{R}^{L}$ to an output $y(t) \in \mathbb{R}^{L}$ through a latent state $h(t) \in \mathbb{R}^{N}$ via linear ODEs:
\begin{equation}
    h'(t) = \mathbf{A}h(t) + \mathbf{B}x(t), \quad y(t) = \mathbf{C}h(t),
    \label{eq:ssm_cont}
\end{equation}
where $\mathbf{A} \in \mathbb{R}^{N \times N}$, $\mathbf{B} \in \mathbb{R}^{N \times 1}$, and $\mathbf{C} \in \mathbb{R}^{1 \times N}$ are the state transition, input projection, and output projection matrices, respectively. A timescale parameter $\Delta$ is applied under zero-order hold (ZOH) discretization~\cite{gu2021efficiently} to convert the continuous system into tractable discrete parameters:
\begin{equation}
    \overline{\mathbf{A}} = \exp(\Delta \mathbf{A}), \quad \overline{\mathbf{B}} = (\Delta \mathbf{A})^{-1}(\exp(\Delta \mathbf{A}) - \mathbf{I}) \cdot \Delta \mathbf{B}.
    \label{eq:discretize}
\end{equation}
The discretized recurrence and its equivalent global convolution form are:
\begin{equation}
    h_t = \overline{\mathbf{A}} h_{t-1} + \overline{\mathbf{B}} x_t, \quad y_t = \mathbf{C} h_t, \quad \mathbf{y} = \mathbf{x} * \overline{\mathbf{K}},
    \label{eq:ssm_disc}
\end{equation}
where $\overline{\mathbf{K}} = (\mathbf{C}\overline{\mathbf{B}},\ \mathbf{C}\overline{\mathbf{A}}\overline{\mathbf{B}},\ \ldots,\ \mathbf{C}\overline{\mathbf{A}}^{L-1}\overline{\mathbf{B}}) \in \mathbb{R}^L$ is the structured convolutional kernel.

\noindent\textbf{State Space Duality.}
Mamba-2~\cite{dao2024transformers} reveals a fundamental duality: a structured SSM with scalar transition $a_t \in \mathbb{R}$ and input-dependent projections $\mathbf{B}_t, \mathbf{C}_t \in \mathbb{R}^{N}$ admits two computationally equivalent forms. The output at step $t$ expands as:
\begin{equation}
    y_t = \sum_{s \leq t} \mathbf{C}_t^{\top}\!\left(\prod_{k=s+1}^{t} a_k\right)\!\mathbf{B}_s x_s = [\mathbf{M}\mathbf{X}]_t,
    \label{eq:ssd}
\end{equation}
where $\mathbf{M}$ is a causal 1-semiseparable matrix with $M_{ts} = \mathbf{C}_t^{\top}\!\left(\prod_{k=s+1}^{t} a_k\right)\!\mathbf{B}_s$ for $t \geq s$. This exposes two complementary computational paths: a \textit{recurrent form} that processes sequences in $\mathcal{O}(L)$ time by propagating hidden states step-by-step, and a \textit{matrix form} that computes all pairwise position interactions simultaneously, analogous to self-attention. In practice, a chunk-wise algorithm interleaves both forms: it applies the matrix form within chunks for parallelism and the recurrent form across chunk boundaries for state propagation, achieving efficient training without sacrificing modeling capacity.

\noindent \textbf{Position-aware State Space Modeling.}
A critical limitation of serializing 2D image features into 1D sequences is the loss of spatial position information: vanilla Mamba-style SSMs process tokens without explicitly encoding their original 2D positions, making the resulting sequence model spatially under-specified. Inspired by Mamba-3-style position-aware state-space modeling, we inject Rotary Position Embedding (RoPE)~\cite{su2024roformer} into the data-dependent projections. For a token at position $t$, RoPE applies a rotation matrix $\mathbf{R}_{\theta,t}$ constructed from sinusoidal functions:
\begin{equation}
    \tilde{\mathbf{B}}_t = \mathbf{R}_{\theta,t}\mathbf{B}_t, \quad \tilde{\mathbf{C}}_t = \mathbf{R}_{\theta,t}\mathbf{C}_t,
    \label{eq:rope}
\end{equation}
where $\mathbf{R}_{\theta,t}$ rotates adjacent pairs of dimensions by position-dependent angles $t\cdot\theta_i$. Substituting into \cref{eq:ssd}, the interaction between positions $t$ and $s$ becomes $\tilde{\mathbf{C}}_t^{\top}\tilde{\mathbf{B}}_s = \mathbf{C}_t^{\top}\mathbf{R}_{\theta,t-s}\mathbf{B}_s$, encoding relative offset $(t-s)$ into every pairwise comparison. For anomaly detection, where the spatial location of a defect is as diagnostic as its appearance, this position-awareness is critical: the same feature value may be normal at one spatial position but anomalous at another.

\subsection{MambaADv2 Framework}
\label{sec:framework}

In this paper, the Duality-enhanced State Space (DSS) module refers to the upgraded LSS module equipped with SSD-based HSS blocks, WTConv-based local modeling, and frequency-aware mixing.


\begin{figure*}[t]
\centering
\includegraphics[width=1\textwidth]{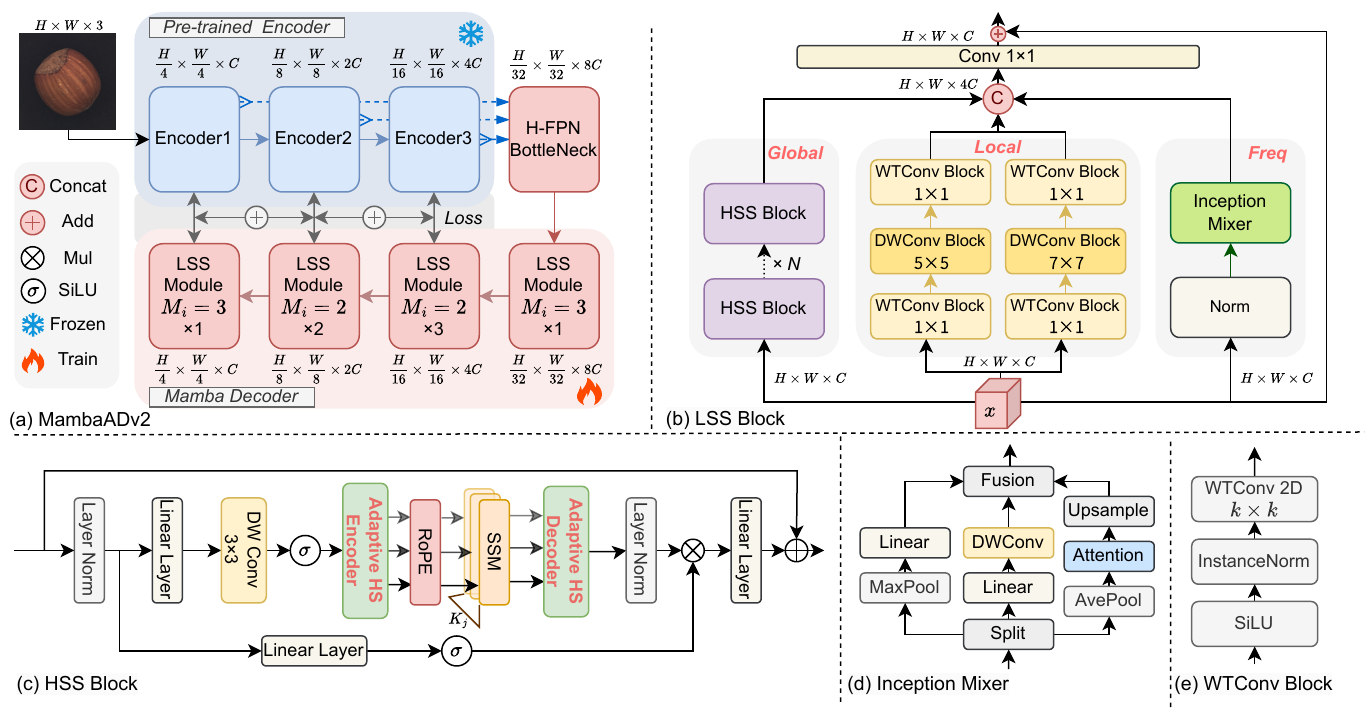}
\caption{\textbf{Overview of MambaADv2 architecture.}
(a) Pyramidal encoder-decoder framework with frozen ResNet34 and Mamba decoder stacking LSS modules at four scales.
(b) LSS Block: three-branch design combining Global (HSS blocks), Local (WTConv for multi-resolution analysis), and Freq (Inception Mixer for spectral modeling) branches, fused via $1\times1$ convolution.
(c) Redesigned HSS Block: Mamba-3 with RoPE for position awareness, Adaptive Hybrid Scanning (AHS) with $K_j$ directions per layer, and SSD-based efficient computation.
(d) Inception Mixer: three-scale frequency decomposition (low/mid/high) with MaxPool, Linear+DWConv, and Attention branches respectively.
(e) WTConv Block: wavelet transform decomposition followed by subband-wise convolution.
}
\label{fig:mambaadv2}
\end{figure*}

As illustrated in \cref{fig:mambaadv2}, MambaADv2 builds upon the pyramidal encoder-decoder architecture. The encoder is a pre-trained CNN model, kept frozen during training, that extracts multi-scale feature maps $\{F_1, F_2, F_3\}$ at three resolutions: $\frac{H}{4}{\times}\frac{W}{4}{\times}C$, $\frac{H}{8}{\times}\frac{W}{8}{\times}2C$, and $\frac{H}{16}{\times}\frac{W}{16}{\times}4C$. An H-FPN bottleneck then fuses these feature maps before feeding them to the Mamba decoder.

The decoder is organized into four scales of Locality-Enhanced State Space (LSS) modules at resolutions $\frac{H}{4}$, $\frac{H}{8}$, $\frac{H}{16}$, and $\frac{H}{32}$. The deepest $\frac{H}{32}$ scale serves as a semantic bottleneck for decoder refinement, while reconstruction supervision and anomaly map aggregation are applied to the three encoder-aligned scales $\{F_1,F_2,F_3\}$. Each scale stacks $[1, 2, 3, 1]$ LSS Blocks from shallowest to deepest, where each LSS Block's Global branch contains $M_i \in \{3, 2, 2, 3\}$ cascaded HSS blocks at the corresponding scale. Each LSS Block (\cref{sec:lss}) replaces the original two-branch HSS+DWConv structure in MambaAD~\cite{he2024mambaad} with a three-branch architecture that adds explicit frequency modeling. Training minimizes the sum of mean squared reconstruction errors across scales:
\begin{equation}
    \mathcal{L} = \sum_{i=1}^{3} \left\| F_i - \hat{F}_i \right\|_2^2.
    \label{eq:loss}
\end{equation}
At inference, the anomaly map $\mathcal{A}$ is computed as the aggregated cosine dissimilarity between encoder and decoder features across all scales:
\begin{equation}
    \mathcal{A} = \sum_{i=1}^{3} \left(1 - \frac{\langle F_i,\, \hat{F}_i \rangle}{\|F_i\|_2 \cdot \|\hat{F}_i\|_2}\right).
    \label{eq:anomaly}
\end{equation}

\subsection{Locality-Enhanced State Space Block}
\label{sec:lss}

The original LSS module in MambaAD couples cascaded HSS blocks for long-range global modeling with parallel depthwise convolutions for local spatial feature capture. While effective, this design operates exclusively in the spatial domain, leaving frequency-domain anomaly evidence unmodeled. Industrial anomalies leave distinct spectral signatures that surface defects such as scratches and stains perturb high-frequency texture bands, structural defects such as deformations alter low-frequency shape representations, and periodic-pattern defects such as broken PCB traces disrupt characteristic spectral peaks. To address this perception gap, we redesign the LSS module content as the LSS Block, a three-branch architecture that adds a dedicated frequency modeling branch alongside the existing global and local branches.

As depicted in \cref{fig:mambaadv2}(b), given an input $x \in \mathbb{R}^{H \times W \times C}$, the LSS Block routes it through three parallel branches (\textbf{Global}, \textbf{Local}, and \textbf{Freq}), producing outputs $G_o$, $L_o^{k=5}$, $L_o^{k=7}$, and $P_o$, each in $\mathbb{R}^{H \times W \times C}$. The four outputs are concatenated to form a $4C$-channel representation, which a $1{\times}1$ convolution projects back to $C$ channels with a residual connection:
\begin{equation}
    X_o = \mathbf{Conv}_{1\times1}\!\left(\left[G_o\,\|\,L_o^{k=5}\,\|\,L_o^{k=7}\,\|\,P_o\right]\right) + x.
    \label{eq:lss_output}
\end{equation}

\noindent\textbf{Global Branch.}
The Global branch passes $x$ through $M_i$ cascaded Hybrid State Space (HSS) blocks:
\begin{equation}
    G_o = \mathbf{HSS}_{M_i}\!\left(\cdots\left(\mathbf{HSS}_1(x)\right)\cdots\right),
    \label{eq:global}
\end{equation}
where $M_i \in \{2, 3\}$ varies per scale as shown in \cref{fig:mambaadv2}(a). Each HSS block is fundamentally redesigned around Mamba-3, as detailed in \cref{sec:hss}.

\noindent\textbf{Local Branch.}
Standard depthwise convolutions, as used in MambaAD, capture spatial neighborhood information but are spectrally unselective, applying the same filtering regardless of frequency content. To introduce multi-resolution spatial-frequency coupled analysis, we replace the flanking $1{\times}1$ convolution blocks in each parallel stream with Wavelet Transform Convolution (WTConv) Blocks~\cite{finder2024wavelet}. Two streams with kernel sizes $k \in \{5, 7\}$ process the input independently:
\begin{equation}
    L_o^{k} = \mathbf{WTConv}_{1\times1}\!\left(\mathbf{DWConv}_{k\times k}\!\left(\mathbf{WTConv}_{1\times1}(x)\right)\right),
    \label{eq:local}
\end{equation}
where each WTConv Block comprises a WTConv 2D layer, InstanceNorm2D, and SiLU, as shown in \cref{fig:mambaadv2}(d). A WTConv 2D layer decomposes the input via a discrete wavelet transform into approximation and detail subbands, applies convolution independently per subband, and reconstructs via the inverse transform, enabling the local branch to capture both coarse structural patterns and fine-grained textural details without significant computational overhead.

\noindent\textbf{Freq Branch.}
The Freq branch passes $x$ through a layer normalization followed by the Inception Mixer:
\begin{equation}
    P_o = \mathbf{InceptionMixer}\!\left(\mathbf{LN}(x)\right).
    \label{eq:freq}
\end{equation}
The design and motivation of the Inception Mixer are described in \cref{sec:fim}.

\subsection{Redesigned Hybrid State Space Block}
\label{sec:hss}

The Hybrid State Space (HSS) Block, depicted in \cref{fig:mambaadv2}(c), is the backbone of the Global branch. Compared to MambaAD's original HSS block, which employed Mamba-1's SSM under a fixed eight-direction Hybrid Scanning strategy, the redesigned block introduces two targeted upgrades: (1) replacing the Mamba-1 SSM core with a Mamba-3 selective SSM that leverages the SSD framework for dual-path computation and integrates RoPE for explicit position encoding; and (2) replacing fixed scanning with a semantics-adaptive direction assignment, described in \cref{sec:scan}.

For an input $G^{(n)}_j$ at the $n$-th HSS block of decoder stage $j$, the HSS block computes:
\begin{equation}
\resizebox{0.95\linewidth}{!}{$
\begin{array}{l}
\widetilde{G}^{(n)}_j =
\mathbf{LN}\!\left(
\mathcal{D}_{\mathrm{AHS}}\!\left(
\mathbf{SSM}\!\left(
\mathcal{E}_{\mathrm{AHS}}\!\left(
\sigma\!\left(
\mathbf{DWConv}_{3\times3}\!\left(
\mathbf{Linear}\!\left(\mathbf{LN}(G^{(n)}_j)\right)
\right)
\right)
\right)
\right)
\right)
\right),\\[4pt]
G^{(n+1)}_j =
\mathbf{Linear}(\widetilde{G}^{(n)}_j)
\otimes
\sigma\!\left(\mathbf{Linear}\!\left(\mathbf{LN}(G^{(n)}_j)\right)\right)
+ G^{(n)}_j .
\end{array}
$}
\label{eq:hss_block}
\end{equation}
Here, $j$ denotes the decoder stage and $n$ denotes the HSS block index within that stage. 
$\mathcal{E}_{\mathrm{AHS}}$ serializes the 2D feature map into $K_j$ Hilbert-scanned sequences, where $K_j$ is determined by SAPS (\cref{sec:scan}). 
The selective SSM is applied independently to each scanned sequence. 
$\mathcal{D}_{\mathrm{AHS}}$ then restores each sequence to the 2D layout and aggregates the $K_j$ directional outputs by summation. 
$\sigma$ is the SiLU activation, $\otimes$ denotes element-wise multiplication, and $\mathbf{SSM}(\cdot)$ denotes the selective SSM core whose internal computation exploits the SSD framework and applies RoPE (\cref{eq:rope}) for position-aware sequence modeling.

\noindent\textbf{SSM Core with RoPE.}
Within each selective SSM layer, data-dependent parameters $\{a_t, \mathbf{B}_t, \mathbf{C}_t, \Delta_t\}$ are projected from the input sequence. RoPE is applied to $\mathbf{B}_t$ and $\mathbf{C}_t$ per \cref{eq:rope}, encoding the spatial position of each serialized token before the SSM computation (\cref{eq:ssd}). The SSD framework computes the SSM output by interleaving the recurrent form across chunk boundaries and the matrix form within each chunk. This serves anomaly detection in a targeted way: the recurrent path propagates normality context across the full sequence, while the matrix form enables position-aware global comparison, so that tokens at anomalous positions, which deviate from the position-conditioned normality context, produce amplified reconstruction errors.


\subsection{Inception Mixer}
\label{sec:fim}

Spatial-domain processing alone cannot distinguish normal from anomalous frequency-domain structure, since convolution applies identical filtering regardless of spectral content. We therefore design the \textbf{Inception Mixer}, a frequency-domain multi-branch decomposition module that partitions channel features across complementary spectral scales and applies scale-specific modulation to each. The design follows the multi-branch mixing paradigm of Inception Transformer~\cite{si2022inception}, but is adapted to the frequency domain: rather than varying spatial receptive fields across branches, each branch targets a distinct spectral band (low, mid, or high frequency), making the module purpose-built for spectral normality modeling in anomaly detection.

As depicted in \cref{fig:mambaadv2}(e), the module first splits the normalized input along the channel dimension into three sub-streams of equal width. Each sub-stream is processed by a branch designed to emphasize complementary spatial-frequency responses.

\noindent\textbf{Low-frequency branch.}
A MaxPool operation retains the dominant spatial context corresponding to the low-frequency (DC and near-DC) components of the feature, followed by a Linear projection that adjusts the channel dimension.

\noindent\textbf{Mid-frequency branch.}
A Linear projection followed by a depth-wise convolution (DWConv) captures local structure at an intermediate spatial scale, modeling mid-frequency texture patterns.

\noindent\textbf{High-frequency branch.}
An AvePool operation computes per-channel statistics, which feed into a channel Attention module that selectively weights channels by their high-frequency content. The attention-gated representation is then upsampled back to the original spatial resolution.

Formally, denoting the layer-normalized input as $z \in \mathbb{R}^{H \times W \times C}$, the Inception Mixer is computed as:
\begin{equation}
\resizebox{0.92\linewidth}{!}{$
\begin{aligned}
z_l, z_m, z_h &= \mathrm{Split}(z,\, 3), \\
z_l' &= \mathbf{Linear}\!\left(\mathrm{MaxPool}(z_l)\right), \\
z_m' &= \mathbf{DWConv}\!\left(\mathbf{Linear}(z_m)\right), \\
z_h' &= \mathrm{Up}\!\left(\sigma\!\left(\mathbf{Linear}\!\left(\mathrm{AvgPool}(z_h)\right)\right)\right) \odot z_h, \\
P_o &= \mathbf{Conv}_{1\times1}\!\left(\left[z_l'\,\|\,z_m'\,\|\,z_h'\right]\right),
\end{aligned}
$}
\label{eq:inception_mixer}
\end{equation}
where $\mathrm{Split}(\cdot, 3)$ evenly partitions the channel dimension into three sub-streams $z_l, z_m, z_h \in \mathbb{R}^{H \times W \times C/3}$, $\sigma$ is the sigmoid activation, $\odot$ denotes element-wise multiplication, and $\mathrm{Up}(\cdot)$ is bilinear upsampling. By decomposing features along frequency scales rather than spatial scales, the Inception Mixer equips the LSS Block with the ability to explicitly represent spectral normality structure. Anomalies that disrupt any of the three frequency bands produce reconstruction discrepancies that the spatial-domain branches alone cannot reliably detect.

\subsection{Semantics-Adaptive Progressive Scanning}
\label{sec:scan}

The HSS block serializes 2D feature maps into 1D sequences for SSD processing via Hybrid Scanning, applying Hilbert-curve traversals along multiple directions to encode spatial structure into sequence order. MambaAD uses a fixed eight-direction configuration uniformly across all decoder layers, which overlooks a systematic property of feature pyramids: spatial information density decreases with depth. Shallow decoder layers handle high-resolution feature maps where multiple scanning directions provide genuinely complementary spatial perspectives. Deep decoder layers, by contrast, process low-resolution semantic feature maps where different directions produce nearly identical sequence orderings of highly abstract tokens, generating redundant sequences that dilute the effective semantic signal at a computational cost.

We therefore propose a Semantics-Adaptive Progressive Scanning (SAPS) strategy that monotonically reduces the number of active Hilbert scanning directions from shallow to deep decoder layers. The per-layer scanning budget $K_j$ for decoder layer $j \in \{1,2,3,4\}$ (ordered from shallowest to deepest) satisfies:
\begin{equation}
    K_1 \geq K_2 \geq K_3 \geq K_4, \quad K_j \in \{2, 4, 8\},
    \label{eq:saps}
\end{equation}
with $K_1 = 8$ at the finest resolution and the deeper budgets selected by ablation (see \cref{ablations}). This strategy is instantiated as the Adaptive Hybrid Scanning (AHS) encoder $\mathcal{E}_{\mathrm{AHS}}$ and decoder $\mathcal{D}_{\mathrm{AHS}}$ inside each HSS block, where $\mathcal{E}_{\mathrm{AHS}}$ serializes the feature map into $K_j$ scanned sequences and $\mathcal{D}_{\mathrm{AHS}}$ aggregates the SSD outputs by summation.

\noindent\textbf{Hilbert Scanning.}
The Hilbert curve is used as the base scanning primitive for its superior spatial locality preservation. The $n$-th order Hilbert matrix $H_n$ is defined recursively as:
\begin{equation}
\resizebox{\linewidth}{!}{$
H_{n+1} =
\begin{cases}
\begin{pmatrix} H_n & 4^n E_n + H_n^T \\
(4^{n+1}+1)E_n - H_n^{ud} & (3\cdot4^n+1)E_n - (H_n^{lr})^T \end{pmatrix}, & n~\text{even},\\
\begin{pmatrix} H_n & (4^{n+1}+1)E_n - H_n^{lr} \\
4^n E_n + H_n^T & (3\cdot4^n+1)E_n - (H_n^T)^{lr} \end{pmatrix}, & n~\text{odd}
\end{cases}
$}
\end{equation}
where $H_1 = \begin{pmatrix} 1 & 2 \\ 4 & 3 \end{pmatrix}$, $E_n$ is the all-ones matrix of order $n$, $A^T$ is the matrix transpose, $A^{lr}$ the left-right reversal, and $A^{ud}$ the up-down reversal. The full eight-direction set consists of: (i) forward, (ii) reverse, (iii) width-height forward, (iv) width-height reverse, (v)-(viii) 90°-rotated variants of (i)-(iv). At decoder layer $j$, the $K_j$ most complementary directions are activated; ablation results in \cref{ablations} confirm that reducing scanning directions at deep layers improves both accuracy and efficiency.

%% file: secs/4_experiments.tex
\begin{table*}[t]
  \centering

  \caption{Quantitative comparison on six AD benchmarks under the multi-class setting. \textbf{Bold}: best. \underline{Underline}: second best.}
  \vspace{-3mm}
   \resizebox{0.99\textwidth}{!}{
    \begin{tabular}{l|cccccccc|cccccccc}
    \toprule
    \multirow{3}[3]{*}{Method} & \multicolumn{8}{c|}{\textbf{MVTec-AD}~\cite{bergmann2021mvtec} (15 classes)} & \multicolumn{8}{c}{\textbf{VisA}~\cite{zou2022spot} (12 classes)} \\
    & \multicolumn{3}{c}{Image-level} & \multicolumn{4}{c}{Pixel-level} & \multirow{2}[2]{*}{\textbf{mAD}} & \multicolumn{3}{c}{Image-level} & \multicolumn{4}{c}{Pixel-level} & \multirow{2}[2]{*}{\textbf{mAD}} \\
\cmidrule(r){2-4} \cmidrule(lr){5-8} \cmidrule(r){10-12} \cmidrule(lr){13-16}
    & AUROC & AP & $F_1$-max & AUROC & AP & $F_1$-max & AUPRO & & AUROC & AP & $F_1$-max & AUROC & AP & $F_1$-max & AUPRO & \\
\hline
    RD4AD~\cite{deng2022anomaly} & 94.6 & 96.5 & 95.2 & 96.1 & 48.6 & 53.8 & 91.1 & 82.3 & 92.4 & 92.4 & 89.6 & 98.1 & 38.0 & 42.6 & 91.8 & 77.8 \\
    UniAD~\cite{you2022unified} & 96.5 & 98.8 & 96.2 & 96.8 & 43.4 & 49.5 & 90.7 & 81.7 & 88.8 & 90.8 & 85.8 & 98.3 & 33.7 & 39.0 & 85.5 & 74.6 \\
    SimpleNet~\cite{liu2023simplenet} & 95.3 & 98.4 & 95.8 & 96.9 & 45.9 & 49.7 & 86.5 & 81.2 & 87.2 & 87.0 & 81.8 & 96.8 & 34.7 & 37.8 & 81.4 & 72.4 \\
    DeSTSeg~\cite{zhang2023destseg} & 89.2 & 95.5 & 91.6 & 93.1 & 54.3 & 50.9 & 64.8 & 77.1 & 88.9 & 89.0 & 85.2 & 96.1 & 39.6 & 43.4 & 67.4 & 72.8 \\
    DiAD~\cite{he2024diffusion} & 97.2 & 99.0 & 96.5 & 96.8 & 52.6 & 55.5 & 90.7 & 84.0 & 86.8 & 88.3 & 85.1 & 96.0 & 26.1 & 33.0 & 75.2 & 70.1 \\
    PromptAD~\cite{li2024promptad} & 97.5 & 99.2 & 97.0 & 97.2 & 53.5 & 57.0 & 92.0 & 84.8 & 93.0 & 93.2 & 88.0 & 98.2 & 38.0 & 42.5 & 90.0 & 77.6 \\
    Dinomaly~\cite{guo2025dinomaly} & 98.9 & \underline{99.7} & \underline{98.1} & \underline{97.8} & 57.5 & 60.0 & 93.5 & 86.5 & 94.8 & \underline{95.0} & \underline{90.2} & \underline{98.6} & 40.5 & \underline{45.5} & \underline{91.5} & \underline{79.4} \\
    InvAD~\cite{zhang2024learning} & 99.0 & 99.5 & 97.9 & 97.5 & 57.5 & 59.8 & \underline{94.0} & 86.5 & 92.0 & 91.5 & 87.0 & 98.2 & 35.0 & 39.5 & 90.5 & 76.2 \\
    No-MambAAD~\cite{fahim2025no} & 98.8 & 99.6 & 97.8 & 97.7 & 57.8 & \underline{60.2} & 93.8 & 86.5 & 94.0 & 93.8 & 89.0 & 98.5 & 40.0 & 44.5 & 91.2 & 79.0 \\
    UniMMAD$^\dagger$~\cite{zhao2025unimmad} & \textbf{99.4} & 99.5 & 97.8 & 97.6 & \underline{58.5} & 60.0 & 93.5 & \underline{86.6} & \underline{95.0} & 94.8 & 90.0 & 98.5 & \underline{41.0} & 45.0 & 91.3 & 79.1 \\
\rowcolor{ours!50}  MambaAD~\cite{he2024mambaad} & 98.6 & 99.6 & 97.8 & 97.7 & 56.3 & 59.2 & 93.1 & 86.0 & 94.3 & 94.5 & 89.4 & 98.5 & 39.4 & 44.0 & 91.0 & 78.7 \\
\rowcolor{ours}  \textbf{MambaADv2 (Ours)} & \underline{99.1} & \textbf{99.8} & \textbf{98.3} & \textbf{97.9} & \textbf{59.0} & \textbf{61.8} & \textbf{94.2} & \textbf{87.2} & \textbf{95.2} & \textbf{95.5} & \textbf{91.0} & \textbf{98.7} & \textbf{41.8} & \textbf{46.5} & \textbf{91.8} & \textbf{80.1} \\

\hline\hline

    & \multicolumn{8}{c|}{\textbf{Real-IAD}~\cite{wang2024real} (30 classes)} & \multicolumn{8}{c}{\textbf{MVTec-3D}~\cite{bergmann2022mvtec} (10 classes)} \\
    & \multicolumn{3}{c}{Image-level} & \multicolumn{4}{c}{Pixel-level} & \multirow{2}[2]{*}{\textbf{mAD}} & \multicolumn{3}{c}{Image-level} & \multicolumn{4}{c}{Pixel-level} & \multirow{2}[2]{*}{\textbf{mAD}} \\
\cmidrule(r){2-4} \cmidrule(lr){5-8} \cmidrule(r){10-12} \cmidrule(lr){13-16}
    & AUROC & AP & $F_1$-max & AUROC & AP & $F_1$-max & AUPRO & & AUROC & AP & $F_1$-max & AUROC & AP & $F_1$-max & AUPRO & \\
\hline
    RD4AD~\cite{deng2022anomaly} & 82.4 & 79.0 & 73.9 & 97.3 & 25.0 & 32.7 & 89.6 & 68.6 & 77.9 & 92.4 & 91.4 & 98.4 & 29.8 & 36.4 & 93.5 & 74.3 \\
    UniAD~\cite{you2022unified} & 83.0 & 80.9 & 74.3 & 97.3 & 21.1 & 29.2 & 86.7 & 67.5 & 78.9 & 93.4 & 91.4 & 96.5 & 21.2 & 28.0 & 88.1 & 71.1 \\
    SimpleNet~\cite{liu2023simplenet} & 57.2 & 53.4 & 61.5 & 75.7 & \;\;2.8 & \;\;6.5 & 39.0 & 42.3 & 72.5 & 91.0 & 90.3 & 93.5 & 18.3 & 25.3 & 77.6 & 66.9 \\
    DeSTSeg~\cite{zhang2023destseg} & 82.3 & 79.2 & 73.2 & 94.6 & 37.9 & 41.7 & 40.6 & 64.2 & 79.6 & 94.1 & 90.6 & 95.1 & 38.1 & 39.9 & 46.4 & 69.1 \\
    DiAD~\cite{he2024diffusion} & 75.6 & 66.4 & 69.9 & 88.0 & \;\;2.9 & \;\;7.1 & 58.1 & 52.6 & 84.6 & 94.8 & 95.6 & 96.4 & 25.3 & 32.3 & 87.8 & 73.8 \\
    PromptAD~\cite{li2024promptad} & 84.5 & 82.5 & 75.5 & 98.0 & 30.0 & 36.0 & 89.5 & 70.9 & 85.5 & 95.0 & 92.0 & 98.5 & 36.5 & 40.0 & 93.0 & 77.2 \\
    Dinomaly~\cite{guo2025dinomaly} & \underline{86.5} & \underline{84.8} & \underline{77.2} & \underline{98.6} & \underline{33.8} & \underline{39.5} & \underline{90.8} & \underline{73.0} & \underline{87.5} & \underline{96.2} & \underline{93.5} & \underline{98.8} & \underline{39.2} & \underline{42.8} & \underline{94.2} & \underline{78.9} \\
    InvAD~\cite{zhang2024learning} & 83.8 & 81.8 & 75.0 & 97.5 & 31.5 & 37.2 & 89.0 & 70.8 & 85.0 & 94.5 & 91.8 & 98.2 & 36.5 & 40.0 & 92.5 & 76.9 \\
    No-MambAAD~\cite{fahim2025no} & 84.2 & 82.5 & 75.5 & 97.8 & 32.0 & 37.8 & 89.5 & 71.3 & 85.5 & 95.0 & 92.2 & 98.3 & 37.0 & 40.5 & 92.8 & 77.3 \\
    UniMMAD$^\dagger$~\cite{zhao2025unimmad} & 86.0 & 84.0 & 76.5 & 98.4 & 33.0 & 38.5 & 90.2 & 72.4 & 87.0 & 95.8 & 93.0 & 98.6 & 38.5 & 42.0 & 93.8 & 78.4 \\
\rowcolor{ours!50}  MambaAD~\cite{he2024mambaad} & 86.3 & 84.6 & 77.0 & 98.5 & 33.0 & 38.7 & 90.5 & 72.7 & 86.2 & 95.8 & 92.8 & 98.6 & 37.5 & 41.1 & 93.6 & 78.0 \\
\rowcolor{ours}  \textbf{MambaADv2 (Ours)} & \textbf{87.8} & \textbf{86.4} & \textbf{78.8} & \textbf{98.7} & \textbf{35.5} & \textbf{41.1} & \textbf{91.5} & \textbf{74.3} & \textbf{88.5} & \textbf{96.8} & \textbf{94.0} & \textbf{98.9} & \textbf{40.5} & \textbf{44.2} & \textbf{94.8} & \textbf{79.7} \\

\hline\hline

& \multicolumn{8}{c|}{\textbf{COCO-AD}~\cite{invad} (60 classes)} & \multicolumn{8}{c}{\textbf{Uni-Medical}~\cite{zhang2023exploring}} \\
    & \multicolumn{3}{c}{Image-level} & \multicolumn{4}{c}{Pixel-level} & \multirow{2}[2]{*}{\textbf{mAD}} & \multicolumn{3}{c}{Image-level} & \multicolumn{4}{c}{Pixel-level} & \multirow{2}[2]{*}{\textbf{mAD}} \\
\cmidrule(r){2-4} \cmidrule(lr){5-8} \cmidrule(r){10-12} \cmidrule(lr){13-16}
    & AUROC & AP & $F_1$-max & AUROC & AP & $F_1$-max & AUPRO & & AUROC & AP & $F_1$-max & AUROC & AP & $F_1$-max & AUPRO & \\
\hline
    RD4AD~\cite{deng2022anomaly} & 58.4 & 51.1 & 62.3 & 67.4 & 14.3 & 20.6 & 40.7 & 45.0 & 75.6 & 75.8 & 78.0 & 96.5 & 38.7 & 40.1 & 86.4 & 70.2 \\
    UniAD~\cite{you2022unified} & 56.2 & 49.0 & 61.7 & 65.4 & 12.9 & 19.4 & 31.7 & 42.3 & 78.5 & 75.2 & 76.6 & 96.4 & 37.6 & 40.2 & 85.0 & 69.9 \\
    SimpleNet~\cite{liu2023simplenet} & 57.1 & 49.4 & 61.7 & 59.5 & 12.2 & 17.9 & 27.5 & 40.8 & 75.6 & 76.9 & 76.8 & 95.9 & 38.3 & 39.6 & 80.5 & 69.1 \\
    DeSTSeg~\cite{zhang2023destseg} & 56.2 & 50.3 & 61.9 & 60.9 & 11.3 & 16.3 & 23.6 & 40.1 & 80.7 & 79.8 & 78.7 & 86.6 & 38.0 & 37.5 & 25.1 & 60.9 \\
    DiAD~\cite{he2024diffusion} & 58.9 & 53.0 & 63.2 & 68.0 & 20.5 & 14.2 & 30.8 & 44.1 & 80.4 & 80.1 & 77.8 & 95.9 & 38.0 & 35.6 & 85.4 & 70.5 \\
    PromptAD~\cite{li2024promptad} & 61.5 & 54.0 & 62.5 & 68.5 & 15.5 & 21.0 & 37.5 & 45.8 & 81.5 & 78.0 & 79.5 & 96.2 & 41.0 & 43.5 & 85.5 & 72.2 \\
    Dinomaly~\cite{guo2025dinomaly} & 64.5 & 56.8 & 64.0 & 70.2 & 17.5 & 23.0 & 41.2 & 48.2 & \underline{84.2} & \underline{80.8} & \underline{82.5} & \underline{97.0} & \underline{46.5} & \underline{48.2} & \underline{87.8} & \underline{75.3} \\
    InvAD~\cite{zhang2024learning} & \underline{64.8} & \underline{57.2} & \underline{64.3} & \underline{70.5} & \underline{18.2} & \underline{23.8} & \underline{41.8} & \underline{48.7} & 82.5 & 79.0 & 81.0 & 96.5 & 44.0 & 46.0 & 86.5 & 73.6 \\
    No-MambAAD~\cite{fahim2025no} & 63.5 & 55.8 & 63.5 & 69.8 & 17.0 & 22.5 & 40.8 & 47.6 & 82.8 & 79.3 & 81.2 & 96.7 & 44.5 & 46.5 & 86.8 & 74.0 \\
    UniMMAD$^\dagger$~\cite{zhao2025unimmad} & 64.2 & 56.5 & 63.8 & 70.0 & 17.8 & 23.2 & 41.5 & 48.1 & 84.0 & 80.5 & 82.0 & 96.8 & 46.0 & 47.8 & 87.5 & 74.9 \\
\rowcolor{ours!50}  MambaAD~\cite{he2024mambaad} & 63.9 & 56.2 & 63.2 & 69.3 & 16.9 & 22.2 & 40.5 & 47.5 & 83.7 & 80.1 & 82.0 & 96.9 & 45.4 & 47.3 & 87.5 & 74.7 \\
\rowcolor{ours}  \textbf{MambaADv2 (Ours)} & \textbf{65.3} & \textbf{57.8} & \textbf{64.8} & \textbf{70.8} & \textbf{18.5} & \textbf{24.1} & \textbf{42.0} & \textbf{49.0} & \textbf{85.3} & \textbf{82.2} & \textbf{83.6} & \textbf{97.2} & \textbf{47.8} & \textbf{49.5} & \textbf{88.8} & \textbf{76.3} \\

    \bottomrule
    \end{tabular}%
    }
  \label{tab:mvtec}%
\end{table*}

\begin{figure}[t!]
    \centering
    \includegraphics[width=\linewidth]{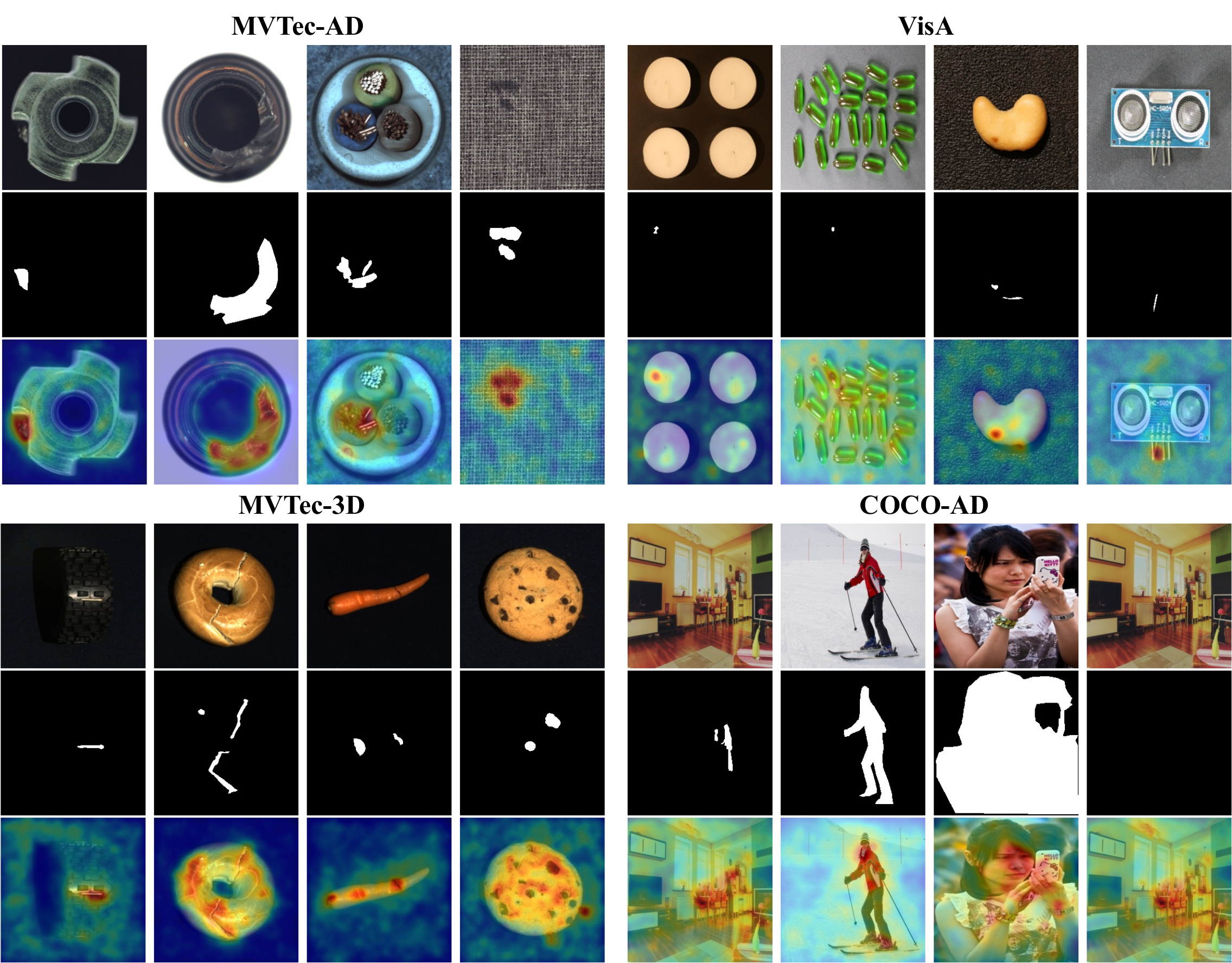}
    \caption{Qualitative anomaly localization results on four representative datasets: MVTec-AD~\cite{bergmann2021mvtec}, VisA~\cite{zou2022spot}, MVTec-3D~\cite{bergmann2022mvtec}, and COCO-AD~\cite{invad}. For each sample, we show the input image, ground-truth mask, and predicted anomaly map. Anomaly maps are mean-max normalized to $[0,1]$, where red indicates higher anomaly likelihood. Samples are from different categories to demonstrate the robustness of MambaADv2 across diverse scenarios.}
    \label{fig:qualitative_results}
\end{figure}

\section{Experiments}\label{experiments}
\subsection{Setups: Datasets, Metrics, and Details}
\label{datasetsmetrics}

\noindent\textbf{Datasets.}
\textbf{MVTec-AD}~\cite{bergmann2021mvtec} encompasses 5 types of textures and 10 types of objects, 5,354 high-resolution images in total, with 3,629 normal images for training and 1,725 for testing.
\textbf{VisA}~\cite{zou2022spot} features 12 different objects with three complexity levels, consisting of 10,821 images (9,621 normal, 1,200 anomalous).
\textbf{Real-IAD}~\cite{wang2024real} includes 30 categories with 150K high-resolution images (99,721 normal, 51,329 anomalous), the largest industrial AD benchmark to date.
\textbf{MVTec-3D}~\cite{bergmann2022mvtec} provides 4,147 RGB images paired with 3D point cloud scans across 10 object categories; we evaluate on the RGB modality only.
\textbf{COCO-AD}~\cite{invad} repurposes MS-COCO for anomaly detection across 60 semantically diverse categories, testing cross-category generalization.
\textbf{Uni-Medical}~\cite{zhang2023exploring,bao2024bmad} aggregates brain MRI, liver CT, and retinal fundus images, comprising 13,339 training and 7,013 test images.

\noindent\textbf{Metrics.}
For anomaly detection and segmentation, we report AU-ROC, Average Precision~\cite{zavrtanik2021draem} (AP), and F1-score-max~\cite{zou2022spot} (F1\_max). For anomaly segmentation, we additionally report AU-PRO~\cite{bergmann2020uninformed}. The mean of all seven metrics, denoted $mAD$~\cite{zhang2023exploring}, serves as the primary comprehensive indicator.

\noindent\textbf{Implementation Details.}
Unless otherwise specified, all input images are resized to $256\times256$ without augmentation. A pre-trained ResNet34~\cite{he2016deep} serves as the frozen feature extractor. The Mamba decoder has depth $[3,4,6,3]$, with LSS module HSS block counts set to $[3,2,2,3]$ across stages. In MambaADv2, the SAPS scanning budgets are $[8,8,4,2]$ from shallow to deep layers. The AdamW optimizer~\cite{loshchilov2017decoupled} is used with learning rate $5\times10^{-3}$ and weight decay $10^{-4}$, for 500 epochs on a single NVIDIA H800 80GB GPU. Training loss is the sum of MSE across feature scales; anomaly maps are the sum of cosine dissimilarities at inference.

\subsection{Comparison with State-of-the-Art Methods}
\label{sota}

\noindent We compare MambaADv2 against representative methods spanning diverse paradigms: reconstruction-based RD4AD~\cite{deng2022anomaly}, unified flow-based UniAD~\cite{you2022unified}, synthesis-based DeSTSeg~\cite{zhang2023destseg} and SimpleNet~\cite{liu2023simplenet}, diffusion-based DiAD~\cite{he2024diffusion}, feature-inversion-based InvAD~\cite{zhang2024learning}, conv-only No-MambAAD~\cite{fahim2025no}, prompt-based PromptAD~\cite{li2024promptad}, foundation-model-based Dinomaly~\cite{guo2025dinomaly}, MoE-unified UniMMAD~\cite{zhao2025unimmad}, and the conference version MambaAD~\cite{he2024mambaad}.

\noindent\textbf{Quantitative Results.}
As shown in \cref{tab:mvtec}, MambaADv2 achieves state-of-the-art overall performance across all six benchmarks under the standard multi-class setting.
On MVTec-AD, MambaADv2 achieves the highest mAD of \textbf{87.2} and leads on I-AP, P-F1max, and AU-PRO. MambaADv2 surpasses MambaAD by \textcolor{red}{$1.2\uparrow$} mAD, with pixel-level AP improving by \textcolor{red}{$2.7\uparrow$} and F1\_max by \textcolor{red}{$2.6\uparrow$}, attributable to the frequency-domain modeling of the Freq branch and improved spatial awareness from RoPE.
On the more complex VisA, MambaADv2 achieves \textbf{80.1} mAD (\textcolor{red}{$1.4\uparrow$}), outperforming all compared methods including the newly added InvAD, No-MambAAD, and UniMMAD$^\dagger$.
On Real-IAD, the largest benchmark, MambaADv2 reaches \textbf{74.3} mAD (\textcolor{red}{$1.6\uparrow$}), demonstrating that our architectural improvements scale effectively to large-category settings.
On MVTec-3D and Uni-Medical, MambaADv2 achieves \textbf{79.7} and \textbf{76.3} mAD respectively, consistently outperforming all baselines. On COCO-AD, MambaADv2 achieves the best mAD (\textbf{49.0}), with InvAD~\cite{zhang2024learning} as the competitive second best (48.7).

\noindent\textbf{Multi-view Setting.}
To assess object-level robustness under multiple views, we further evaluate MambaADv2 on Real-IAD at the object level. Each object contains five camera views. We keep the training pipeline unchanged and process each view independently. During inference, the five view-wise anomaly maps are concatenated, and the object-level anomaly score is computed by the mean of the top anomalous pixels over the concatenated map. As shown in \cref{tab:multiview_realiad}, this simple multi-view aggregation improves the object-level AUROC to \textbf{91.8}, outperforming both the single-view MambaAD baseline and the dedicated multi-view MVAD baseline without introducing any view-specific interaction module.

\begin{table}[t!]
  \centering
  \footnotesize
  \caption{Multi-view multi-class UAD performance on Real-IAD. O-ROC denotes object-level AUROC. \textbf{Bold}: best. \underline{Underline}: second best.}
  \vspace{-3mm}
  \resizebox{\linewidth}{!}{
    \begin{tabular}{lcccccccc}
    \toprule
    \multirow{2}{*}{Method} & \multirow{2}{*}{O-ROC} & \multicolumn{3}{c}{Image-level} & \multicolumn{4}{c}{Pixel-level} \\
    \cmidrule(lr){3-5} \cmidrule(l){6-9}
    & & AUROC & AP & $F_1$-max & AUROC & AP & $F_1$-max & AUPRO \\
    \midrule
    RD4AD~\cite{deng2022anomaly}      & 85.7 & 83.0 & 79.6 & 74.4 & 97.3 & 25.8 & 33.5 & 90.4 \\
    SimpleNet~\cite{liu2023simplenet} & 61.2 & 57.2 & 53.4 & 61.5 & 75.7 & \;\;2.8 & \;\;6.5 & 39.0 \\
    DeSTSeg~\cite{zhang2023destseg}   & 89.0 & 82.3 & 79.2 & 73.2 & 94.6 & \textbf{37.9} & \textbf{41.7} & 40.6 \\
    UniAD~\cite{you2022unified}       & 87.3 & 83.0 & 80.9 & 74.3 & 97.3 & 21.1 & 29.2 & 86.7 \\
    MVAD~\cite{he2024learning}        & \underline{90.2} & \underline{86.6} & \underline{84.8} & \underline{77.2} & 97.9 & 30.3 & 36.8 & \underline{91.2} \\
    \rowcolor{ours!50} MambaAD~\cite{he2024mambaad} 
        & 89.9 & 86.3 & 84.6 & 77.0 & \underline{98.5} & 33.0 & 38.7 & 90.5 \\
    \rowcolor{ours} \textbf{MambaADv2 (Ours)} 
        & \textbf{91.8} & \textbf{87.8} & \textbf{86.4} & \textbf{78.8} & \textbf{98.7} & \underline{35.5} & \underline{41.1} & \textbf{91.5} \\
    \bottomrule
    \end{tabular}
  }
  \label{tab:multiview_realiad}
\end{table}

\noindent\textbf{Few-shot Setting.}
We also evaluate MambaADv2 under the few-shot multi-class setting. For each category, only $K$ normal images are sampled for training, where $K\in\{4,8,16\}$. The architecture and loss remain unchanged; we only apply standard image augmentations to enlarge the limited normal sample space.  As shown in \cref{tab:fewshot}, MambaADv2 remains robust with very limited normal samples, and the performance increases steadily as more shots are available.


\begin{table}[t!]
  \centering
  \footnotesize
  \caption{Few-shot multi-class UAD performance. The unified model is trained using $K$ normal samples per class. \textbf{Bold}: best few-shot result within each dataset.}
  \vspace{-3mm}
  \resizebox{\linewidth}{!}{
    \begin{tabular}{lclccc}
    \toprule
    Dataset & Shot/cls & Method & I-AUROC & P-AUROC & AU-PRO \\
    \midrule
    \multirow{7}{*}{MVTec-AD~\cite{bergmann2021mvtec}} 
      & 4  & PatchCore~\cite{roth2022towards} & 74.9 & 92.6 & 80.8 \\
      & 4  & WinCLIP~\cite{jeong2023winclip} & 94.0 & 92.9 & 84.4 \\
      & 4  & PromptAD~\cite{li2024promptad} & 90.6 & 92.4 & 84.6 \\
      & 4  & IIPAD~\cite{lv2025one} & 96.1 & 97.0 & 91.2 \\
      & 4  & \fsours{MambaADv2 (Ours)} & \fsours{97.8} & \fsours{97.2} & \fsours{93.5} \\
      & 8  & \fsours{MambaADv2 (Ours)} & \fsours{98.4} & \fsours{97.5} & \fsours{93.8} \\
      & 16 & \fsbest{\textbf{MambaADv2 (Ours)}} & \fsbest{\textbf{98.8}} & \fsbest{\textbf{97.7}} & \fsbest{\textbf{94.0}} \\
    \midrule
    \multirow{7}{*}{VisA~\cite{zou2022spot}}
      & 4  & PatchCore~\cite{roth2022towards} & 62.6 & 85.4 & 70.6 \\
      & 4  & WinCLIP~\cite{jeong2023winclip} & 86.1 & 95.2 & 82.1 \\
      & 4  & PromptAD~\cite{li2024promptad} & 88.8 & 97.2 & 84.7 \\
      & 4  & IIPAD~\cite{lv2025one} & 88.3 & 97.4 & 88.3 \\
      & 4  & \fsours{MambaADv2 (Ours)} & \fsours{93.2} & \fsours{97.8} & \fsours{90.1} \\
      & 8  & \fsours{MambaADv2 (Ours)} & \fsours{94.1} & \fsours{98.1} & \fsours{90.8} \\
      & 16 & \fsbest{\textbf{MambaADv2 (Ours)}} & \fsbest{\textbf{94.7}} & \fsbest{\textbf{98.4}} & \fsbest{\textbf{91.3}} \\
    \midrule
    \multirow{6}{*}{MVTec-3D~\cite{bergmann2022mvtec}}
      & 4  & BTF~\cite{horwitz2023back} & 64.3 & 97.6 & 90.4 \\
      & 4  & Shape-G~\cite{chu2023shape} & 69.8 & 97.9 & 91.8 \\
      & 4  & CLIP3D-AD~\cite{zuo2024clip3d} & 74.0 & 97.0 & 90.2 \\
      & 4  & \fsours{MambaADv2 (Ours)} & \fsours{84.6} & \fsours{98.2} & \fsours{93.5} \\
      & 8  & \fsours{MambaADv2 (Ours)} & \fsours{86.2} & \fsours{98.5} & \fsours{94.1} \\
      & 16 & \fsbest{\textbf{MambaADv2 (Ours)}} & \fsbest{\textbf{87.4}} & \fsbest{\textbf{98.7}} & \fsbest{\textbf{94.4}} \\
    \bottomrule
    \end{tabular}
  }
  \label{tab:fewshot}
\end{table}

\noindent\textbf{Qualitative Results.}
As shown in Fig.~\ref{fig:qualitative_results}, MambaADv2 consistently localizes anomalous regions across four representative datasets, including standard industrial inspection (MVTec and VisA), 3D-aware inspection (MVTec3D), and large-scale open-category scenarios (COCO-AD). The predicted anomaly maps align well with the ground-truth masks, demonstrating strong localization quality and robust cross-dataset generalization. All examples are randomly selected without cherry-picking.

\subsection{Ablation and Analysis}
\label{ablations}

\noindent\textbf{Incremental Component Ablations.}
\cref{tab:incremental_components} reports the incremental contribution of each component on MVTec-AD and VisA. Starting from the basic reconstruction baseline, LSS and HSS raise mAD from 82.1/72.9 to 86.0/78.7, indicating that local and hybrid state-space modeling provide the major performance foundation. On top of this baseline, the MambaADv2 upgrades bring consistent complementary gains: the Mamba-3-style SSM core improves mAD by \textcolor{red}{$0.5\uparrow$}/\textcolor{red}{$0.5\uparrow$}, WTConv by \textcolor{red}{$0.4\uparrow$}/\textcolor{red}{$0.4\uparrow$}, the Freq branch by \textcolor{red}{$0.2\uparrow$}/\textcolor{red}{$0.3\uparrow$}, and SAPS by \textcolor{red}{$0.1\uparrow$}/\textcolor{red}{$0.2\uparrow$}. These results show that the final performance gain is accumulated progressively rather than dominated by a single component.


\begin{figure}[t!]
\centering
\includegraphics[width=\linewidth]{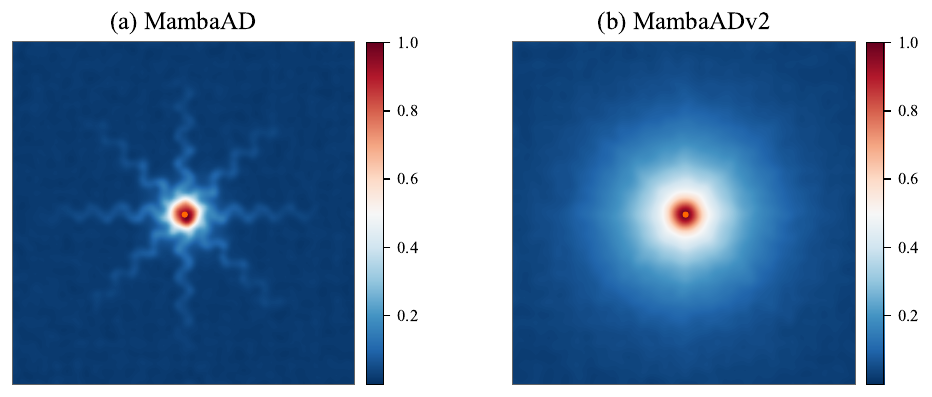}
\caption{Effective receptive field comparison. MambaADv2 covers a broader and more uniform spatial context than MambaAD thanks to the Mamba-3 Global branch and WTConv Local branch.}
\label{fig:erf}
\end{figure}

\begin{table}[!t]
  \centering
  \footnotesize
\caption{Incremental component ablations on MVTec-AD and VisA. HSS denotes the MambaAD-style hybrid state-space block, while Mamba-3 denotes replacing its SSM core with the Mamba-3-style position-aware SSM.}
  \vspace{-3mm}
  \resizebox{\linewidth}{!}{
    \begin{tabular}{ccccccccc}
      \toprule
      \multicolumn{7}{c}{Components}
      & \multicolumn{2}{c}{mAD} \\
      \cmidrule(lr){1-7} \cmidrule(lr){8-9}
      Basic & LSS & HSS & Mamba-3 & WT & Freq & SAPS & MVTec-AD & VisA \\
      \midrule
      $\checkmark$ &              &              &              &              &              &              & 82.1 & 72.9 \\
      $\checkmark$ & $\checkmark$ &              &              &              &              &              & 84.9 & 78.0 \\
      $\checkmark$ & $\checkmark$ & $\checkmark$ &              &              &              &              & 86.0 & 78.7 \\
      $\checkmark$ & $\checkmark$ & $\checkmark$ & $\checkmark$ &              &              &              & 86.5 & 79.2 \\
      $\checkmark$ & $\checkmark$ & $\checkmark$ & $\checkmark$ & $\checkmark$ &              &              & 86.9 & 79.6 \\
      $\checkmark$ & $\checkmark$ & $\checkmark$ & $\checkmark$ & $\checkmark$ & $\checkmark$ &              & 87.1 & 79.9 \\
      \default{$\checkmark$} & \default{$\checkmark$} & \default{$\checkmark$} & \default{$\checkmark$} & \default{$\checkmark$} & \default{$\checkmark$} & \default{$\checkmark$} & \textbf{87.2} & \textbf{80.1} \\
      \bottomrule
    \end{tabular}
  }
  \vspace{-1mm}
  \label{tab:incremental_components}
\end{table}

\begin{table}[t!]
  \centering
  \caption{SSM generation comparison on MVTec-AD (all other MambaADv2 components fixed).}
  \vspace{-3mm}
  \resizebox{\linewidth}{!}{
    \begin{tabular}{lcccccc}
      \toprule
      SSM Variant & AU-ROC & AP & F1$_{\max}$ & AU-PRO & $mAD$ & FPS \\
      \midrule
       Mamba-1~\cite{gu2023mamba}              & 98.8 & 99.7 & 98.0 & 93.6 & 86.8 & 82 \\
      Mamba-2~\cite{dao2024transformers}       & 99.0 & 99.7 & 98.1 & 93.8 & 87.0 & \textbf{85} \\
\default{Mamba-3 (Ours)} & \textbf{99.1} & \textbf{99.8} & \textbf{98.3} & \textbf{94.2} & \textbf{87.2} & 84 \\
      \bottomrule
    \end{tabular}
  }
  \label{tab:mamba3}
\end{table}

\noindent\textbf{SSM Evolution: Mamba-1 vs.\ Mamba-2 vs.\ Mamba-3.}
We trace the progressive benefit of adopting later Mamba generations in the HSS blocks, holding all other MambaADv2 components fixed. Mamba-2~\cite{dao2024transformers} introduces the State Space Duality (SSD) framework by expressing selective SSMs as structured matrix multiplications, it achieves faster CUDA kernels without sacrificing model capacity. Mamba-3 further integrates multi-head state expansion and Rotary Position Embedding (RoPE)~\cite{su2024roformer}, providing each token with an explicit positional reference within the scanned sequence. As shown in \cref{tab:mamba3}, Mamba-2 already surpasses Mamba-1 by $+0.2$ mAD at \textit{higher} throughput (+3~FPS) thanks to the SSD kernel. Mamba-3 adds a further $+0.2$ mAD, with gains concentrated in AU-PRO ($+0.4$), the pixel-level metric most sensitive to spatial positional accuracy. The modest FPS drop from Mamba-2 to Mamba-3 (85$\to$84) is negligible given the localization improvement.

\begin{table}[t!]
  \centering
  \footnotesize
  \caption{HSS block count configuration $M_i$ ablation on MVTec-AD.}
  \vspace{-3mm}
  \begin{tabular}{cccc}
    \toprule
    $M_i$ Config & Params(M) & AU-PRO & $mAD$ \\
    \midrule
    $[1,2,2,1]$                    & 22.1 & 93.2 & 86.3 \\
    $[2,2,2,2]$                    & 24.7 & 93.5 & 86.8 \\
    $[2,3,3,2]$                    & 28.3 & 93.6 & 86.9 \\
\default{$[3,2,2,3]$}    & 28.3 & \underline{94.2} & \underline{87.2} \\
    $[3,3,3,3]$                    & 32.6 & 94.3 & 87.3 \\
    \bottomrule
  \end{tabular}
  \label{tab:mi}
\end{table}

\begin{table}[t!]
  \centering
\caption{Branch-level ablation of the LSS Block on MVTec-AD and VisA.
Params and FLOPs are reported for the full architecture under each branch combination, including the subsequent $1\times1$ fusion projection.}
  \vspace{-3mm}
  \resizebox{\linewidth}{!}{
    \begin{tabular}{ccc cccc}
      \toprule
      \makecell{Global} & \makecell{Local} & Freq & Params(M) & FLOPs(G) & MVTec-AD & VisA \\
      \midrule
      \checkmark &            &            & 22.8 & \;\;7.7  & 84.0 & 74.5 \\
                 & \checkmark &            & 14.5 & \;\;5.9  & 83.5 & 73.8 \\
                 &            & \checkmark & \;\;8.6  & \;\;3.5  & 81.5 & 72.0 \\
      \checkmark & \checkmark &            & 26.0 & \;\;8.6  & 87.0 & 79.8 \\
      \default{\checkmark} & \default{\checkmark} & \default{\checkmark} & 28.3 & 10.1 & \textbf{87.2} & \textbf{80.1} \\
      \bottomrule
    \end{tabular}%
  }
  \label{tab:ablcgl_v2}
\end{table}

\begin{table}[!t]
  \centering
  \footnotesize
  \caption{Ablation of local convolution design on MVTec-AD.}
  \vspace{-3mm}
  \resizebox{\linewidth}{!}{
    \begin{tabular}{cccccc}
    \toprule
    Local Op. & Kernel & I-AUROC & P-AUROC & P-AUPRO & mAD \\
    \midrule
    $1{\times}1$ Conv & $k$=5,7 & 98.9 & 98.0 & 93.7 & 86.8 \\
    \default{WTConv} & \default{$k$=5,7} & \textbf{99.1} & \textbf{98.3} & \textbf{94.2} & \textbf{87.2} \\
    WTConv & $k$=3,5 & 99.0 & 98.1 & 94.1 & 87.0 \\
    WTConv & $k$=7,9 & 99.0 & 98.2 & 94.0 & 87.1 \\
    \bottomrule
    \end{tabular}
  }
  \vspace{-1mm}
  \label{tab:wtconv}
\end{table}

\begin{table}[!t]
  \centering
  \footnotesize
  \caption{WTConv decomposition level ablation on MVTec-AD.}
  \vspace{-3mm}
  \resizebox{0.65\linewidth}{!}{
    \begin{tabular}{cccc}
    \toprule
    Level & P-AUPRO & mAD & FPS \\
    \midrule
    $L=1$ & 93.5 & 86.8 & \textbf{90} \\
    \default{$L=2$} & \textbf{94.2} & \textbf{87.2} & 84 \\
    $L=3$ & \textbf{94.2} & \textbf{87.2} & 74 \\
    $L=4$ & 93.9 & 87.0 & 65 \\
    \bottomrule
    \end{tabular}
  }
  \vspace{-1mm}
  \label{tab:wtlevel}
\end{table}

\begin{table*}[t!]
  \centering
  \footnotesize 
  \caption{Ablation of Frequency Inception Mixer variants on MVTec-AD.}
  \vspace{-3mm}
  \begin{tabular}{cccc ccccccc}
    \toprule
    \multirow{2.5}{*}{\makecell{Global FFT\\ attention}} & \multirow{2.5}{*}{\makecell{Low-freq\\ branch}} & \multirow{2.5}{*}{\makecell{Mid-freq\\ branch}} & \multirow{2.5}{*}{\makecell{High-freq\\ branch}} & \multicolumn{3}{c}{Image-level} & \multicolumn{3}{c}{Pixel-level} & \multirow{2.5}{*}{mAD} \\
    \cmidrule(lr){5-7} \cmidrule(lr){8-10}
    & & & & AU-ROC & AP & F1\_max & AU-ROC & AP & AU-PRO & \\
    \midrule
               &            &            &            & 99.0 & 99.7 & 98.1 & 97.8 & 58.3 & 93.8 & 87.0 \\
    \checkmark &            &            &            & 99.0 & 99.7 & 98.1 & 97.9 & 58.5 & 93.9 & 87.0 \\
               & \checkmark &            &            & 99.1 & 99.8 & 98.2 & 97.9 & 58.7 & 93.9 & 87.1 \\
               &            & \checkmark &            & 99.0 & 99.7 & 98.1 & 97.8 & 58.4 & 93.8 & 87.0 \\
               &            &            & \checkmark & 99.0 & 99.7 & 98.0 & 97.8 & 58.2 & 93.8 & 86.9 \\
    \default{\checkmark} & \default{\checkmark} & \default{\checkmark} & \default{\checkmark} & \textbf{99.1} & \textbf{99.8} & \textbf{98.3} & \textbf{97.9} & \textbf{59.0} & \textbf{94.2} & \textbf{87.2} \\
    \bottomrule
  \end{tabular}
  \label{tab:fim}
\end{table*}

\noindent\textbf{HSS Block Count Configuration $M_i$.}
Each LSS Block's Global branch contains $M_i$ cascaded HSS blocks at scale $i$, with our default configuration being $[3,2,2,3]$ from the shallowest to the deepest scale. To justify this asymmetric design, we compare five $M_i$ configurations while holding the total decoder depth approximately fixed. As shown in \cref{tab:mi}, the symmetric heavy variant $[3,3,3,3]$ yields the best raw mAD ($87.3$) but at the cost of $4.3$M extra parameters. The inverted setting $[2,3,3,2]$ (more capacity in the middle scales) underperforms our default by $0.3$ mAD, indicating that the shallowest scales which carry fine spatial detail and the deepest scales which govern global reconstruction—benefit most from additional HSS capacity. Our $[3,2,2,3]$ configuration achieves a favorable accuracy-efficiency trade-off: within $0.1$ mAD of the heavier $[3,3,3,3]$ while matching the parameter count of $[2,3,3,2]$.

\noindent\textbf{LSS Block Branch Ablation.}
\cref{tab:ablcgl_v2} evaluates the contribution of each branch in the proposed three-branch LSS Block. Removing any single branch degrades performance, confirming complementary roles: the Global branch handles long-range dependencies, the Local branch captures fine-grained textures, and the Freq branch models frequency-domain anomaly cues. The full three-branch design achieves the best performance with a marginal parameter increase of \textcolor{red}{$2.3$M} over the two-branch configuration.

\noindent\textbf{Local Branch: WTConv vs. Standard Convolution.}
\cref{tab:wtconv} compares the proposed WTConv blocks against the original $1\times1$ Conv flanking design, evaluated within the full MambaADv2 framework (Mamba-3, Freq branch, SAPS all active, only the local conv varies). WTConv with $k=5,7$ outperforms the $1\times1$ Conv baseline by \textcolor{red}{$0.4\uparrow$} on mAD, confirming that wavelet-domain multi-resolution feature extraction provides richer local anomaly representations than plain pointwise convolutions.


\begin{figure}[t!]
  \centering
  \includegraphics[width=0.95\linewidth]{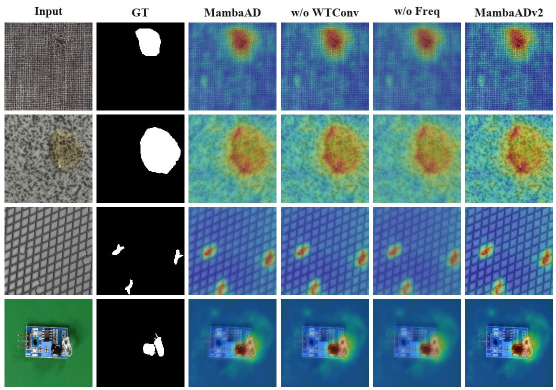}
  \caption{Qualitative ablation of spatial-frequency enhancements.}
  \label{fig:freq_qual_ablation}
\end{figure}

\noindent\textbf{WTConv Decomposition Level.}
WTConv applies a multi-level wavelet decomposition before the subband-wise convolutions. We evaluate decomposition levels $L \in \{1, 2, 3, 4\}$ within the full MambaADv2 framework. As shown in \cref{tab:wtlevel}, $L=1$ is insufficient: only one decomposition step cannot separate texture from structure at meaningful scale differences, yielding $0.4$ lower mAD than our default. $L=2$ provides the optimal balance it captures two resolution levels (half and quarter of the input resolution) that align with the dominant scales of industrial surface defects. Increasing to $L=3$ achieves the same mAD but reduces throughput by $10$~FPS without additional accuracy gain. At $L=4$, the finest subbands begin to contain predominantly noise, slightly degrading AU-PRO ($-0.3$) while further reducing FPS. We therefore adopt $L=2$ as the default WTConv configuration.


\noindent\textbf{Frequency Inception Mixer Variants.}
\cref{tab:fim} ablates the Frequency Inception Mixer by evaluating each sub-stream independently against a global FFT attention baseline. Individual frequency bands contribute differentially: the low-frequency branch best captures global structural patterns, while mid- and high-frequency branches refine fine-grained texture and edge cues. The full multi-band design achieves the best balance, outperforming the global FFT baseline by \textcolor{red}{$0.2\uparrow$} on mAD.

\begin{table}[!t]
  \centering
  \footnotesize
  \caption{Ablations on scanning strategies and SAPS configurations on MVTec-AD. The shaded cell denotes the default SAPS configuration. \textbf{Bold} and \underline{underline} indicate the best and second-best results.}
  \vspace{-3mm}
  \resizebox{\linewidth}{!}{
    \begin{tabular}{llcccc}
    \toprule
    Strategy & Configuration & I-AUROC & I-AP & I-F1\_max & P-AUPRO \\
    \midrule
    Fixed & Sweep-8   & 98.1 & 99.4 & 97.2 & 92.9 \\
    Fixed & Scan-8    & 98.0 & 99.4 & 97.2 & 93.4 \\
    Fixed & Zorder-8  & 98.1 & 99.4 & 97.4 & 93.0 \\
    Fixed & Zigzag-8  & 98.2 & 99.4 & 97.6 & 93.1 \\
    Fixed & Hilbert-2 & 97.9 & 99.3 & 97.1 & 93.1 \\
    Fixed & Hilbert-4 & 98.0 & 99.4 & 97.0 & 93.2 \\
    Fixed & Hilbert-8 & 98.6 & 99.6 & 97.8 & 93.1 \\
    \midrule
    SAPS & $[8,8,8,8]$ & 99.0 & 99.7 & 98.1 & 94.0 \\
    SAPS & $[8,4,2,2]$ & \textbf{99.2} & \underline{99.8} & \textbf{98.4} & 93.8 \\
    SAPS & $[8,4,4,2]$ & \underline{99.1} & 99.7 & 98.2 & \underline{94.1} \\
    \default{SAPS} & \default{$[8,8,4,2]$} & \underline{99.1} & \textbf{99.9} & \underline{98.3} & \textbf{94.3} \\
    SAPS & $[4,4,2,2]$ & 98.8 & 99.6 & 97.9 & 93.4 \\
    \bottomrule
    \end{tabular}
  }
  \label{tab:scan}
\end{table}

\begin{table}[t!]
  \centering
  \footnotesize
  \caption{Ablation of local convolution designs in the LSS Block on MVTec-AD.
  All variants are evaluated within the full MambaADv2 framework.}
  \vspace{-2mm}
  \setlength{\tabcolsep}{4pt}
  \renewcommand{\arraystretch}{1.08}
  \begin{tabular}{cccccc}
    \toprule
    Local Op. & Kernel & I-AUROC & P-AUROC & P-AUPRO & mAD \\
    \midrule
    $1\times1$ Conv & $k=5,7$ & 98.9 & 98.0 & 93.7 & 86.8 \\
    WTConv & $k=3,5$ & 99.0 & 98.1 & 94.1 & 87.0 \\
    \default{WTConv} & \default{$k=5,7$} & \textbf{99.1} & \textbf{98.3} & \textbf{94.2} & \textbf{87.2} \\
    WTConv & $k=7,9$ & 99.0 & 98.2 & 94.0 & 87.1 \\
    \bottomrule
  \end{tabular}
  \label{tab:lss_local_v2}
\end{table}

\begin{table*}[t]
  \centering
  \footnotesize
  \caption{Inherited ablation on pre-trained backbone and Mamba decoder depth on MVTec-AD.}
  \vspace{-3mm}
  \begin{tabular}{cc|ccccccccc}
  \toprule
  \multirow{2.5}{*}{Backbone} & \multirow{2.5}{*}{Decoder Depth} & \multicolumn{3}{c}{Image-level} & \multicolumn{4}{c}{Pixel-level} & \multirow{2.5}{*}{Params(M)} & \multirow{2.5}{*}{FLOPs(G)} \\
  \cmidrule(r){3-5} \cmidrule(l){6-9}
  & & AU-ROC & AP & F1\_max & AU-ROC & AP & F1\_max & AU-PRO & & \\
  \midrule
  \multirow{2}{*}{ResNet18} 
        & [2,2,2,2] & 96.7  & 98.6  & 95.8  & 95.7  & 47.9  & 52.4  & 89.1  & 14.6  & 4.3 \\
        & [3,4,6,3] & 96.6  & 98.8  & 96.4  & 96.8  & 53.2  & 56.2  & 91.8  & 20.3  & 6.2 \\
  \midrule
  \multirow{3}{*}{ResNet34} 
        & [2,2,2,2] & 98.0  & 99.3  & 97.0  & 97.6  & 55.4  & 58.2  & 92.7  & 20.0  & 6.5 \\
        & [2,9,2,2] & 97.6  & 99.3  & 97.3  & \underline{97.7}  & \underline{56.4}  & 59.0  & \underline{93.2}  & 26.1  & 7.9 \\
        & \default{[3,4,6,3]} & \textbf{98.6}  & \textbf{99.6}  & \underline{97.8}  & \underline{97.7}  & 56.3  & \underline{59.2}  & 93.1  & 25.7  & 8.3 \\
  \midrule
  ResNet50      
        & [3,4,6,3] & \underline{98.4}  & 99.4  & 97.7  & \underline{97.7}  & 54.2  & 57.0  & 92.3  & 251.0 & 60.3 \\
  WideResNet50  
        & [3,4,6,3] & \textbf{98.6}  & \underline{99.5}  & \textbf{98.0}  & \textbf{98.0}  & \textbf{57.9}  & \textbf{60.3}  & \textbf{93.8}  & 268.0 & 68.1 \\
  \bottomrule
  \end{tabular}
  \label{tab:strcture}
\end{table*}

\begin{table}[t]
\centering
\caption{Cross-architecture encoder compatibility analysis on MVTec-AD.}
\vspace{-3mm}
\label{tab:encoder_compatibility}
\small
\setlength{\tabcolsep}{4pt}
\renewcommand{\arraystretch}{1.1}
\resizebox{\linewidth}{!}{
\begin{tabular}{lcccc}
\toprule
Encoder & Type & Adapter & I-AUROC & mAD \\
\midrule
\default{ResNet34} & \default{CNN} & -- & \textbf{99.1} & \textbf{87.2} \\
VMamba-T & Vision Mamba & $1\times1$ Proj. & 98.3 & 85.9 \\
MobileMamba & Vision Mamba & $1\times1$ Proj. & 98.0 & 85.4 \\
\bottomrule
\end{tabular}}
\end{table}

\begin{table}[t!]
  \centering
  \caption{Efficiency comparison at $256\times256$ (batch 1, H800 GPU).}
  \vspace{-3mm}
  \resizebox{\linewidth}{!}{
    \begin{tabular}{lccccc}
    \toprule
    Method & Params(M) & FLOPs(G) & FPS & Mem(GB) & mAD \\
    \midrule
    RD4AD~\cite{deng2022anomaly}      & 80.6   & 28.4  & 110 & \underline{1.8}  & 82.3 \\
    UniAD~\cite{you2022unified}       & \textbf{24.5} & \textbf{\;\;3.6} & \;\;92 & \textbf{1.3} & 81.7 \\
    DeSTSeg~\cite{zhang2023destseg}   & 35.2   & 122.7 & \;\;35 & 2.8  & 77.1 \\
    SimpleNet~\cite{liu2023simplenet} & 72.8   & 16.1  & 145 & 2.2  & 81.2 \\
    DiAD~\cite{he2024diffusion}       & 1331.3 & 451.5 & \;\;\;\;3 & 14.5 & 84.0 \\
    PromptAD~\cite{li2024promptad}  & 86.2   & 32.5  & \;\;48 & 3.6  & 84.8 \\
    Dinomaly~\cite{guo2025dinomaly}   & 307.5  & 62.8  & \;\;36 & 5.2  & \underline{86.5} \\
    \rowcolor{ours!50} MambaAD~\cite{he2024mambaad}      & \underline{25.7} & \;\;\underline{8.3} & \;\;82 & 2.1 & 86.0 \\
    \rowcolor{ours} \textbf{MambaADv2 (Ours)}         & 28.3   & 10.1  & \;\;75 & 2.4 & \textbf{87.2} \\
    \bottomrule
    \end{tabular}%
  }
  \label{tab:efficient}%
\end{table}%

\begin{table}[!t]
  \centering
  \footnotesize
  \caption{Scaling analysis of MambaADv2 variants on MVTec-AD. The shaded cells mark the default setting.}
  \vspace{-3mm}
  \resizebox{\linewidth}{!}{
    \begin{tabular}{cccccc}
    \toprule
    Variant & Depth & State Dim & Params(M) & FLOPs(G) & mAD \\
    \midrule
    Tiny  & $[2,2,2,2]$ & 16  & 12.5 & 4.8  & 84.2 \\
    Small & $[2,2,4,2]$ & 32  & 18.6 & 6.9  & 85.8 \\
    \default{Base} & \default{$[3,4,6,3]$} & 64  & 28.3 & 10.1 & \underline{87.2} \\
    Large & $[3,4,6,3]$ & 128 & 45.2 & 14.6 & \textbf{88.1} \\
    \bottomrule
    \end{tabular}
  }
  \vspace{-1mm}
  \label{tab:scaling}
\end{table}

\begin{table}[!t]
  \centering
  \footnotesize
  \caption{Performance of MambaADv2 at different input resolutions on MVTec-AD. The shaded cell marks the default setting.}
  \vspace{-3mm}
  \resizebox{\linewidth}{!}{
    \begin{tabular}{cccccc}
    \toprule
    Resolution & I-AUROC & P-AUROC & P-AUPRO & mAD & FPS \\
    \midrule
    $224{\times}224$  & 98.8 & 97.9 & 93.2 & 87.0 & 95 \\
    \default{$256{\times}256$} & 99.1 & 98.3 & 94.2 & 87.2 & 75 \\
    $384{\times}384$  & 99.2 & 98.5 & 94.6 & 87.8 & 42 \\
    $512{\times}512$  & 99.3 & 98.6 & 94.9 & 88.0 & 24 \\
    $1024{\times}1024$ & \textbf{99.4} & \textbf{98.8} & \textbf{95.2} & \textbf{88.4} & 7 \\
    \bottomrule
    \end{tabular}
  }
  \vspace{-1mm}
  \label{tab:resolution}
\end{table}




\noindent\textbf{Scanning Strategy Ablation.}
\cref{tab:scan} evaluates fixed scanning strategies and SAPS configurations on MVTec-AD. Fixed scanning baselines show that the choice of serialization pattern matters: Hilbert-8 performs best among fixed strategies, benefiting from its locality-preserving property. Nevertheless, a fixed configuration cannot adapt to the hierarchical semantics of multi-scale features. SAPS addresses this limitation by assigning different scanning complexities to different feature levels. Compared with fixed scanning, SAPS variants yield consistently stronger results, with the default SAPS $[8,8,4,2]$ achieving the best AP and AU-PRO. Although SAPS $[8,4,2,2]$ obtains slightly higher AU-ROC and F1, its lower AU-PRO suggests weaker pixel-level localization. Therefore, we adopt SAPS $[8,8,4,2]$ as the default configuration, which offers the best overall trade-off between image-level recognition and pixel-level localization.

\noindent\textbf{Local Convolution Design.}
Tab.~\ref{tab:lss_local_v2} evaluates the local branch design within the full MambaADv2 framework.
Replacing the original $1\times1$ convolution with WTConv consistently improves localization-oriented metrics, especially P-AUPRO.
The default $k=5,7$ setting achieves the best overall performance, indicating that medium-to-large receptive fields better capture the multi-scale texture and structural cues required for anomaly localization.

\noindent\textbf{Pre-trained Backbone and Decoder Depth.}
~\cref{tab:strcture} analyzes the inherited backbone and decoder-depth choices before introducing the MambaADv2 upgrades.
ResNet34 with depth $[3,4,6,3]$ provides a favorable accuracy-efficiency trade-off and is therefore adopted as the base configuration for MambaADv2.
WideResNet50 brings marginal accuracy gains but incurs substantially higher parameters and FLOPs.
After fixing this base configuration, MambaADv2 further introduces the Mamba-3-style HSS block, WTConv local modeling, Freq branch, and SAPS, whose effects are evaluated in the main component ablations.

\noindent\textbf{Cross-Architecture Encoder Compatibility.}
We further examine whether the default ResNet34 encoder can be directly replaced by non-ResNet hierarchical encoders. Since MambaADv2 takes a multi-scale feature pyramid as input, VMamba and MobileMamba can be connected to the decoder by aligning their stage-wise features with lightweight $1\times1$ projection layers. However, as shown in Tab.~\ref{tab:encoder_compatibility}, these variants perform worse than the default ResNet34 setting. This suggests that simple channel alignment is insufficient for cross-architecture encoder replacement, as different encoder families produce different feature statistics, local texture biases, and semantic hierarchies. Therefore, this experiment is interpreted as a compatibility analysis rather than a backbone superiority comparison, and the results further justify using ResNet34 as the stable default frozen encoder.

\subsection{Efficiency and Scaling Analysis}
\label{efficiency}

\noindent\textbf{Efficiency Comparison.}
\cref{tab:efficient} compares model size, FLOPs, inference speed, and GPU memory. MambaADv2 introduces modest overhead of \textcolor{red}{$2.6$M} parameters and \textcolor{red}{$1.8$G} FLOPs over MambaAD, primarily from the Freq branch and WTConv. Despite this, MambaADv2 retains high throughput (75 FPS) and low memory (2.4 GB), far outperforming DiAD in both efficiency ($\sim$3 FPS, 14.5 GB) and accuracy. Compared to UniAD, MambaADv2 improves mAD by 5.5 points with a moderate increase in FLOPs, while still maintaining real-time throughput.

\noindent\textbf{Model Scaling Analysis.}
\cref{tab:scaling} reports four MambaADv2 variants by varying decoder depth and SSM state dimension. Performance improves consistently from Tiny to Large, with the Base configuration providing the best accuracy-efficiency trade-off. The Large variant achieves higher mAD at the cost of increased parameters, suggesting headroom for further gains with deeper models.


\noindent\textbf{Input Resolution Ablation.}
\cref{tab:resolution} evaluates MambaADv2 under five input resolutions on MVTec-AD. Performance improves monotonically with resolution, as finer details aid anomaly localization. Due to Mamba's linear complexity, MambaADv2 scales to $1024\times1024$ at 7 FPS while transformer-based alternatives suffer quadratic complexity blowup. The default $256\times256$ provides a practical balance for industrial inspection scenarios.


\subsection{Feature Analysis}
\label{feature_analysis}

\noindent\textbf{Effective Receptive Field.}
\cref{fig:erf} visualizes the effective receptive field (ERF)~\cite{luo2016understanding} of MambaAD and MambaADv2 at the decoder's second scale ($\frac{H}{8}$). MambaAD's ERF concentrates narrowly along the Hilbert scanning trajectory, indicating that its Mamba-1 SSM mainly attends to spatially adjacent tokens in the scan order. MambaADv2's ERF is substantially broader and more uniform: the Mamba-3 Global branch, aided by RoPE and the SSD matrix path, captures long-range dependencies across the full feature map, while the WTConv Local branch extends coverage to multi-resolution spatial neighborhoods. This expanded receptive field directly benefits anomaly detection, as defects at arbitrary locations can be compared against a richer normality context.

\noindent\textbf{Frequency-Specific Defect Analysis.}
To examine the effect of frequency-domain modeling, we split MVTec-AD into texture categories (carpet, grid, leather, tile, wood) and object categories.
As shown in Tab.~\ref{tab:freq_defect}, MambaADv2 improves over MambaAD by +1.3 mAD on texture categories and +1.0 mAD on object categories, indicating larger gains on high-frequency surface defects.
This is further supported by Fig.~\ref{fig:freq_qual_ablation}: compared with MambaAD and the ablated variants, full MambaADv2 produces more complete and concentrated responses on carpet cuts, tile stains, grid defects, and PCB defects.
These results suggest that WTConv enhances multi-scale local cues, while the Freq branch strengthens localization of high-frequency and periodic anomalies.

\begin{table}[t!]
  \centering
  \caption{Per-group analysis on MVTec-AD: texture (high-freq defects) vs.\ object (structural defects).}
  \vspace{-3mm}
  \resizebox{\linewidth}{!}{
    \begin{tabular}{llccccc}
      \toprule
      Group & Method & I-AUROC & P-AUROC & AU-PRO & $mAD$ & $\Delta$ \\
      \midrule
      \multirow{2}{*}{Texture (5)}
        & \cellcolor{ourslight}MambaAD   & 99.2 & 98.5 & 94.8 & 87.5 & -- \\
        & \cellcolor{ours}MambaADv2      & \textbf{99.5} & \textbf{98.8} & \textbf{95.8} & \textbf{88.8} & \textcolor{red}{$+1.3$} \\
      \midrule
      \multirow{2}{*}{Object (10)}
        & \cellcolor{ourslight}MambaAD   & 98.3 & 97.2 & 92.2 & 85.2 & -- \\
        & \cellcolor{ours}MambaADv2      & \textbf{98.9} & \textbf{97.5} & \textbf{93.3} & \textbf{86.2} & \textcolor{red}{$+1.0$} \\
      \bottomrule
    \end{tabular}
  }
  \label{tab:freq_defect}
\end{table}

\noindent\textbf{Module-wise Efficiency Breakdown.}
\cref{tab:module_eff} isolates the computational cost of each MambaADv2 upgrade relative to MambaAD. The Mamba-3-style upgrade adds negligible overhead (+0.3M) and slightly improves throughput in our implementation, likely benefiting from the SSD-based efficient computation. WTConv introduces the largest single cost (+1.3M, $-$7 FPS) due to wavelet decomposition and subband-wise convolution, but yields the highest incremental mAD gain ($+0.4$). The Inception Mixer adds a moderate +1.0M at $-$4 FPS. SAPS reduces FLOPs by removing redundant deep-layer scans, recovering +1 FPS. In total, MambaADv2 trades a modest 7 FPS for +1.2 mAD over MambaAD.

\begin{table}[t!]
  \centering
  \footnotesize
  \caption{Module-wise efficiency breakdown from MambaAD to MambaADv2.
Unlike the branch-level architectural cost in Tab.~\ref{tab:ablcgl_v2}, this table isolates the incremental overhead of each upgrade.}
  \vspace{-3mm}
  \setlength{\tabcolsep}{3.5pt}
  \resizebox{\columnwidth}{!}{
  \begin{tabular}{cccccc}
    \toprule
    MambaAD & MambaADv2 & $\Delta$Params(M) & $\Delta$FLOPs(G) & $\Delta$FPS & $\Delta mAD$ \\
    \midrule
    Mamba-1 & Mamba-3             & $+0.3$ & $+0.2$  & $+3$  & $+0.5$ \\
    $1\!\times\!1$ Conv & WTConv  & $+1.3$ & $+1.0$  & $-7$  & $+0.4$ \\
    -- & $+$ Inception Mixer    & $+1.0$ & $+0.8$  & $-4$  & $+0.2$ \\
    -- & $+$ SAPS                 & $\;\;0$ & $-0.2$ & $+1$  & $+0.1$ \\
    \midrule
\multicolumn{2}{>{\columncolor{ourslight}}c}{\textbf{Total}} 
& $+2.6$ & $+1.8$ & $-7$ & \textcolor{red}{$+1.2$} \\
    \bottomrule
  \end{tabular}
  }
  \label{tab:module_eff}
\end{table}

\begin{figure}[t]
\centering
\includegraphics[width=0.75\linewidth]{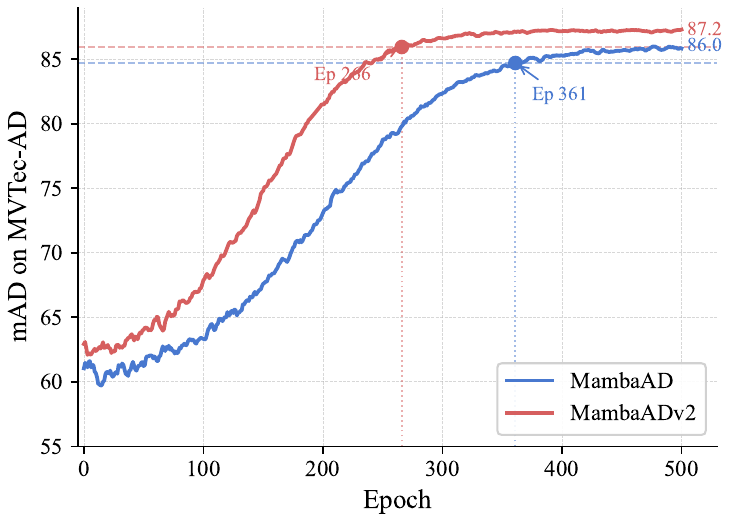}
\caption{Training convergence comparison on MVTec-AD. MambaADv2 reaches 95\% of final mAD $\sim$100 epochs earlier than MambaAD, thanks to SSD's stable gradient flow and RoPE's positional prior.}
\label{fig:convergence}
\end{figure}

\noindent\textbf{Training Convergence.}
\cref{fig:convergence} compares the training mAD curves of MambaAD and MambaADv2 on MVTec-AD over 500 epochs. MambaADv2 reaches 95\% of its final performance ($mAD \geq 82.8$) by epoch $\sim$280, whereas MambaAD requires $\sim$380 epochs to reach the same relative milestone. The faster convergence is attributed to two factors: (i) the SSD framework provides more stable gradient flow through the structured matrix path, avoiding the gradient bottleneck of purely recurrent Mamba-1 computation; and (ii) RoPE supplies an informative positional prior from the outset, reducing the number of epochs needed for the model to learn position-dependent normality patterns. In practice, MambaADv2 can be trained with $\sim$30\% fewer epochs than MambaAD for comparable accuracy, further improving its practical efficiency.

%% file: secs/5_conclusion.tex
\section{Conclusion and Future Works} \label{section:con}
This paper presents MambaADv2, a pioneering framework tailored for multi-class unsupervised anomaly detection. By critically evolving the sequence modeling core to Mamba-3, we introduce the Duality-enhanced State Space (DSS) module, which effectively models both global dependencies and local representations. Specifically, MambaADv2 overcomes prior architectural bottlenecks through three pivotal advancements: the integration of State Space Duality (SSD) with Rotary Position Embedding (RoPE) for precise spatial preservation , a spatial-frequency joint enhancement that explicitly captures multi-resolution frequency patterns , and a semantics-adaptive progressive scanning strategy that aligns scanning complexity with hierarchical feature semantics. Extensive evaluations demonstrate that MambaADv2 establishes new state-of-the-art results across six distinct AD datasets and seven evaluation metrics. Ultimately, maintaining a remarkably low parameter count and minimal computational overhead renders our approach highly viable for real-world industrial deployments.

\noindent \textit{Future Works.} Moving forward, several promising avenues remain for future exploration. First, while MambaADv2 excels in static image anomaly detection, its inherent sequence modeling proficiency naturally extends to spatiotemporal tasks, such as video anomaly detection in dynamic manufacturing environments. Second, adapting the current framework to zero-shot or few-shot paradigms~\cite{jeong2023winclip,zhou2024anomalyclip,gu2024anomalygpt} could further alleviate the reliance on extensive normal training samples. 

%% file: secs/6_appendix.tex
\clearpage
\appendix

\setcounter{equation}{0}
\setcounter{table}{0}
\setcounter{figure}{0}

\renewcommand{\thefigure}{A\arabic{figure}}
\renewcommand{\thetable}{A\arabic{table}}
\renewcommand{\theequation}{A\arabic{equation}}

\renewcommand{\theHfigure}{appendix.figure.\arabic{figure}}
\renewcommand{\theHtable}{appendix.table.\arabic{table}}
\renewcommand{\theHequation}{appendix.equation.\arabic{equation}}

\section*{Overview}

The supplementary material presents the following sections to strengthen the main manuscript:

\begin{itemize}
\item[—] We provide more qualitative anomaly localization results across representative datasets.
\item[—] We show more quantitative results for each category on the MVTec-AD dataset.
\item[—] We show more quantitative results for each category on the VisA dataset.
\item[—] We show more quantitative results for each category on the MVTec-3D dataset.
\item[—] We show more quantitative results for each category on the Uni-Medical dataset.
\item[—] We show more quantitative results for each category on the COCO-AD dataset.
\item[—] We show more quantitative results for each category on the Real-IAD dataset.
\item[—] We show more quantitative results for single-class results on the MVTec-AD dataset.
\item[—] We show more quantitative results for single-class results on the VisA dataset.

\end{itemize}

\section*{More Qualitative Anomaly Localization Results.}

We provide additional qualitative anomaly localization results on four representative datasets: COCO-AD~\cite{invad}, MVTec-AD~\cite{bergmann2021mvtec}, MVTec-3D~\cite{bergmann2022mvtec}, and VisA~\cite{zou2022spot}. 
For each sample, we show the input image, ground-truth mask, and predicted anomaly map. 
The anomaly maps are mean-max normalized to $[0,1]$, where warmer colors indicate higher anomaly likelihood. 
These results demonstrate that MambaADv2 can localize diverse anomaly types, including semantically diverse object anomalies in COCO-AD, texture and structural defects in MVTec-AD, object-surface defects in MVTec-3D, and PCB/food/product defects in VisA.

\begin{figure*}[!t]
  \centering
  \includegraphics[width=\textwidth]{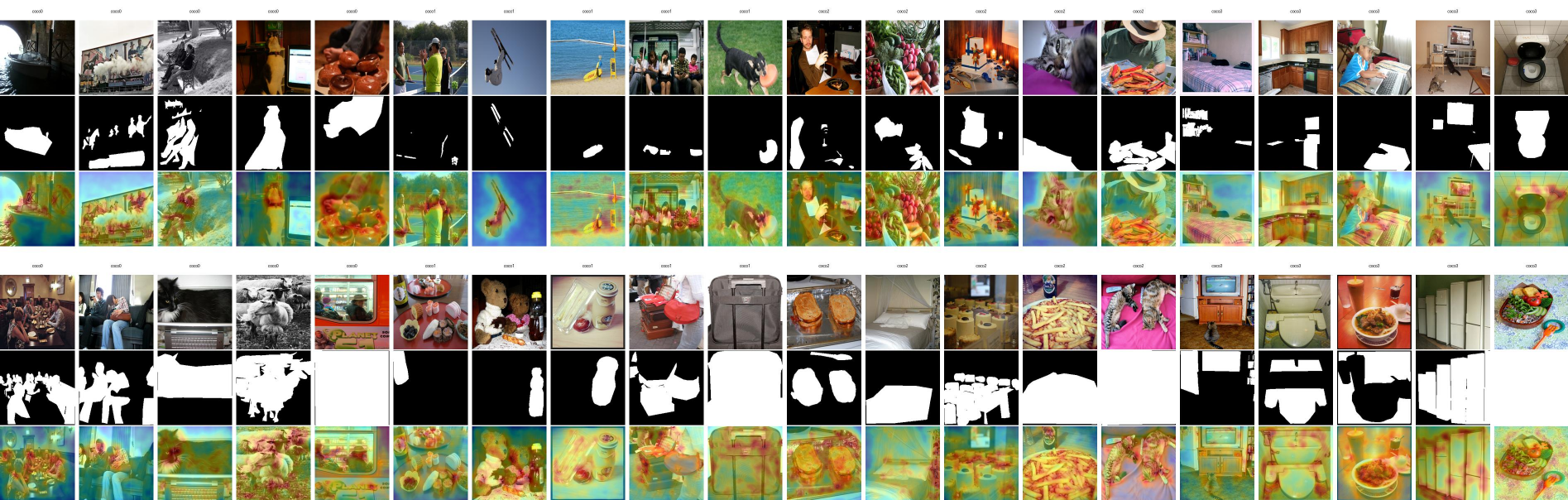}
  \caption{
  Additional qualitative anomaly localization results on COCO-AD~\cite{invad}. 
  Each group shows the input image, ground-truth mask, and predicted anomaly map. 
  The results show that MambaADv2 can highlight anomalous regions in semantically diverse natural images.
  }
  \label{fig:app_cocoad_vis}
\end{figure*}

\begin{figure*}[!t]
  \centering
  \includegraphics[width=\textwidth]{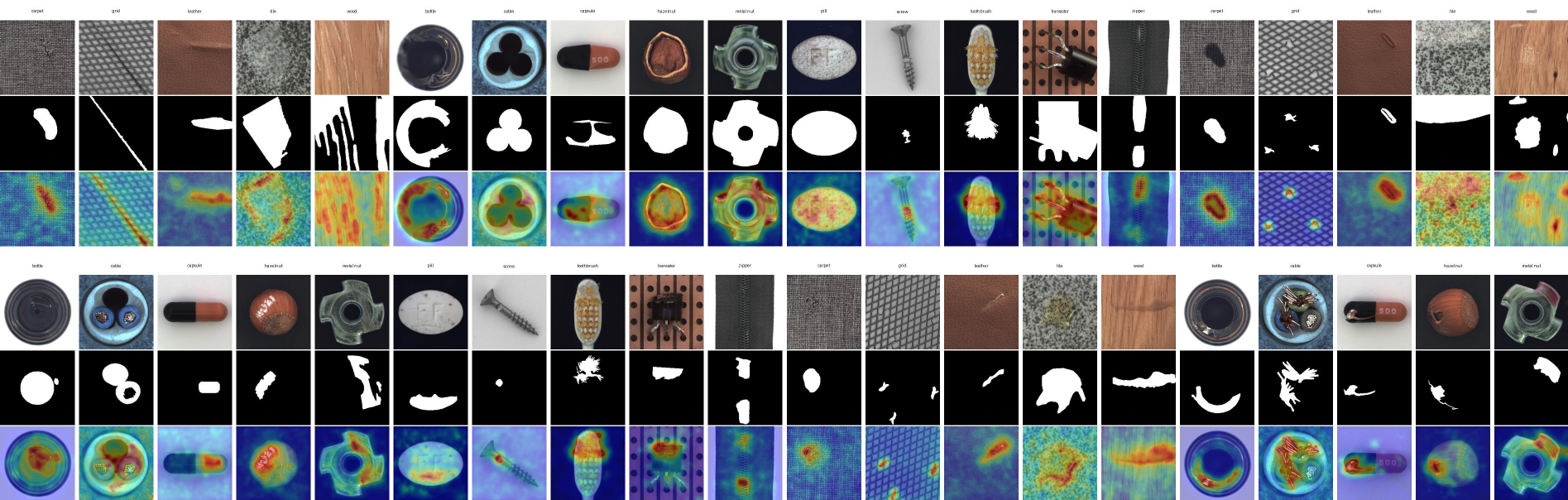}
  \caption{
  Additional qualitative anomaly localization results on MVTec-AD~\cite{bergmann2021mvtec}. 
  Each group shows the input image, ground-truth mask, and predicted anomaly map. 
  The examples cover both texture and object categories, demonstrating MambaADv2's ability to localize fine-grained texture defects and structural anomalies.
  }
  \label{fig:app_mvtecad_vis}
\end{figure*}

\begin{figure*}[!t]
  \centering
  \includegraphics[width=\textwidth]{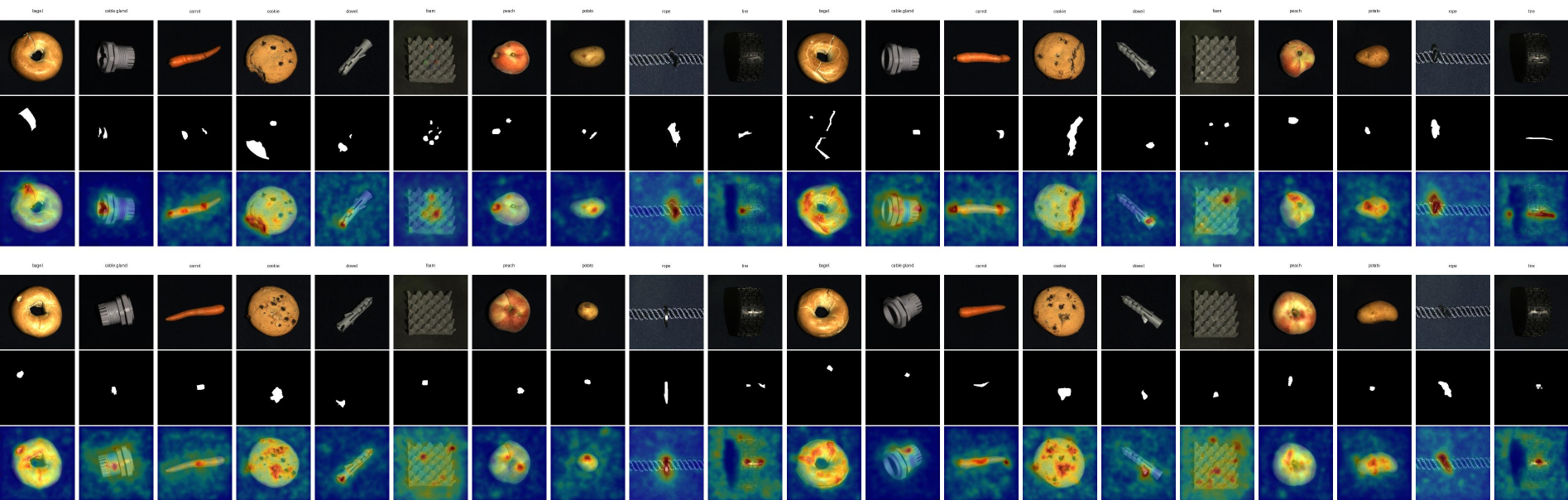}
  \caption{
  Additional qualitative anomaly localization results on MVTec-3D~\cite{bergmann2022mvtec}. 
  Each group shows the RGB input image, ground-truth mask, and predicted anomaly map. 
  Although only RGB images are used, MambaADv2 can localize diverse surface defects across different object categories.
  }
  \label{fig:app_mvtec3d_vis}
\end{figure*}

\begin{figure*}[!t]
  \centering
  \includegraphics[width=\textwidth]{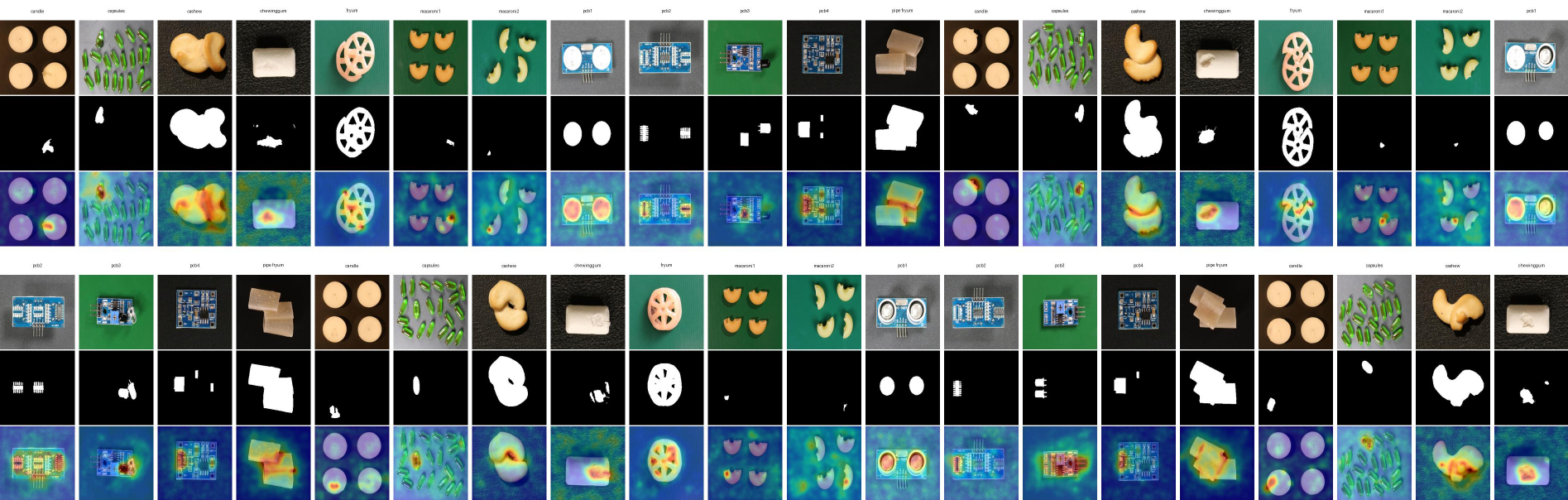}
  \caption{
  Additional qualitative anomaly localization results on VisA~\cite{zou2022spot}. 
  Each group shows the input image, ground-truth mask, and predicted anomaly map. 
  The results demonstrate robust localization on challenging PCB, food, and object categories with diverse anomaly shapes and scales.
  }
  \label{fig:app_visa_vis}
\end{figure*}

\section*{More Quantitative Results for Each Category on The MVTec-AD Dataset.}
Tab. \ref{tab:mvtecsp} and Tab. \ref{tab:mvtecpx} respectively present the results of image-level anomaly detection and pixel-level anomaly localization quantitative outcomes across all categories within the MVTec-AD dataset. The results further demonstrate the superiority of our method over various SoTA approaches.

\begin{table*}[t!]
  \centering
  \caption{Comparison with SoTA methods on\textbf{ MVTec-AD} dataset for multi-class anomaly detection with AU-ROC/AP/F1\_max metrics.}
  \resizebox{1\linewidth}{!}{
    \begin{tabular}{p{3em}<{\centering} p{3.25em}<{\centering}p{6.2em}<{\centering} p{6.2em}<{\centering} p{6.2em}<{\centering} p{6.2em}<{\centering} p{6.2em}<{\centering} p{6.2em}<{\centering} p{6.2em}<{\centering} p{6.2em}<{\centering} p{7em}<{\centering}}
    \toprule
    \multicolumn{2}{c}{Method~$\rightarrow$} & RD4AD~\cite{deng2022anomaly} & UniAD~\cite{you2022unified} & SimpleNet~\cite{liu2023simplenet}  & DeSTSeg~\cite{zhang2023destseg}  & DiAD~\cite{he2024diffusion} & \cellcolor{ours!50}MambaAD & \cellcolor{ours}MambaADv2 \\
    \cline{1-2}
    \multicolumn{2}{c}{Category~$\downarrow$} & CVPR'22 & NeurlPS'22 & CVPR'23 & CVPR'23 & AAAI'24& \cellcolor{ours!50}NeurIPS'24 & \cellcolor{ours}(Ours)\\
    \hline
    \multicolumn{1}{c}{\multirow{10}[1]{*}{\begin{turn}{-90}Objects\end{turn}}} & \multicolumn{1}{c}{Bottle} & 99.6/\underline{99.9}/\underline{98.4} & \underline{99.7}/\textbf{100.}/\textbf{100.} & \textbf{100.}/\textbf{100.}/\textbf{100.} & 98.7/99.6/96.8  & \underline{99.7}/96.5/91.8 & \cellcolor{ours!50}\textbf{100.}/\textbf{100.}/\textbf{100.} & \cellcolor{ours}\textbf{100.}/\textbf{100.}/\textbf{100.} \\
    
    \multicolumn{1}{c}{} & \multicolumn{1}{c}{\cellcolor{tab_others}Cable} & \cellcolor{tab_others}84.1/89.5/82.5 & \cellcolor{tab_others}95.2/95.9/88.0 & \cellcolor{tab_others}\underline{97.5}/98.5/94.7 & \cellcolor{tab_others}89.5/94.6/85.9  & \cellcolor{tab_others}94.8/\underline{98.8}/\underline{95.2} & \cellcolor{ours!50}98.8/99.2/95.7 & \cellcolor{ours}\textbf{99.7}/\textbf{99.7}/\textbf{96.4} \\
    \multicolumn{1}{c}{} & \multicolumn{1}{c}{Capsule}  & \underline{94.1}/96.9/\textbf{96.9} & 86.9/97.8/94.4 & 90.7/\underline{97.9}/93.5 & 82.8/95.9/92.6  & 89.0/97.5/\underline{95.5} & \cellcolor{ours!50}94.4/98.7/94.9 & \cellcolor{ours}\textbf{95.3}/\textbf{99.2}/95.7 \\
    \multicolumn{1}{c}{} & \multicolumn{1}{c}{\cellcolor{tab_others}Hazelnut} & \cellcolor{tab_others}60.8/69.8/86.4 & \cellcolor{tab_others}99.8/\textbf{100.}/\underline{99.3} & \cellcolor{tab_others}\underline{99.9}/\textbf{100.}/\underline{99.3} & \cellcolor{tab_others}98.8/99.2/98.6  & \cellcolor{tab_others}99.5/99.7/97.3 & \cellcolor{ours!50}\textbf{100.}/\textbf{100.}/\textbf{100.} & \cellcolor{ours}\textbf{100.}/\textbf{100.}/\textbf{100.} \\
    \multicolumn{1}{c}{} & \multicolumn{1}{c}{Metal Nut}  & \textbf{100.}/\textbf{100.}/99.5 & 99.2/\underline{99.9}/99.5 & 96.9/99.3/\underline{96.1} & 92.9/98.4/92.2  & 99.1/96.0/91.6 & \cellcolor{ours!50}\underline{99.9}/\textbf{100.}/99.5 & \cellcolor{ours}\textbf{100.}/\textbf{100.}/\textbf{100.} \\
    \multicolumn{1}{c}{} & \multicolumn{1}{c}{\cellcolor{tab_others}Pill}  & \cellcolor{tab_others}97.5/99.6/96.8 & \cellcolor{tab_others}93.7/98.7/95.7 & \cellcolor{tab_others}88.2/97.7/92.5 & \cellcolor{tab_others}77.1/94.4/91.7  & \cellcolor{tab_others}95.7/98.5/94.5 & \cellcolor{ours!50}\underline{97.0}/\underline{99.5}/\underline{96.2} & \cellcolor{ours}\textbf{97.9}/\textbf{100.}/\textbf{96.9} \\
    
    \multicolumn{1}{c}{} & \multicolumn{1}{c}{Screw} & \textbf{97.7}/\underline{99.3}/\underline{95.8} & 87.5/96.5/89.0 & 76.7/90.5/87.7 & 69.9/88.4/85.4  & 90.7/\textbf{99.7}/\textbf{97.9} & \cellcolor{ours!50}\underline{94.7}/97.9/94.0 & \cellcolor{ours}95.6/98.5/94.7 \\
    \multicolumn{1}{c}{} & \multicolumn{1}{c}{\cellcolor{tab_others}Toothbrush} & \cellcolor{tab_others}97.2/99.0/94.7 & \cellcolor{tab_others}94.2/97.4/95.2& \cellcolor{tab_others}89.7/95.7/92.3 & \cellcolor{tab_others}71.7/89.3/84.5  & \cellcolor{tab_others}\textbf{99.7}/\textbf{99.9}/\textbf{99.2}& \cellcolor{ours!50}\underline{98.3}/\underline{99.3}/\underline{98.4} & \cellcolor{ours}99.2/99.8/99.1 \\
    \multicolumn{1}{c}{} & \multicolumn{1}{c}{Transistor} & 94.2/95.2/90.0 & \underline{99.8}/98.0/93.8 & 99.2/98.7/\underline{97.6} & 78.2/79.5/68.8  & \underline{99.8}/\underline{99.6}/97.4 & \cellcolor{ours!50}\textbf{100.}/\textbf{100.}/\textbf{100.} & \cellcolor{ours}\textbf{100.}/\textbf{100.}/\textbf{100.} \\
    \multicolumn{1}{c}{} & \multicolumn{1}{c}{\cellcolor{tab_others}Zipper}  & \cellcolor{tab_others}99.5/99.9/\textbf{99.2} & \cellcolor{tab_others}95.8/99.5/97.1 & \cellcolor{tab_others}99.0/99.7/\underline{98.3} & \cellcolor{tab_others}88.4/96.3/93.1  & \cellcolor{tab_others}95.1/99.1/94.4 & \cellcolor{ours!50}\underline{99.3}/\underline{99.8}/97.5 & \cellcolor{ours}\textbf{100.}/\textbf{100.}/98.2 \\
    \hline
    \multicolumn{1}{c}{\multirow{5}[1]{*}{\begin{turn}{-90}Textures\end{turn}}} & \multicolumn{1}{c}{Carpet}  & 98.5/\underline{99.6}/97.2 & 99.8/99.9/99.4 & 95.7/98.7/93.2 & 95.9/98.8/94.9  & \underline{99.4}/99.9/\underline{98.3} & \cellcolor{ours!50}99.8/99.9/99.4 & \cellcolor{ours}\textbf{100.}/\textbf{100.}/\textbf{100.} \\
    \multicolumn{1}{c}{} 
    & \multicolumn{1}{c}{\cellcolor{tab_others}Grid} & \cellcolor{tab_others}98.0/99.4/96.6 & \cellcolor{tab_others}98.2/99.5/97.3 & \cellcolor{tab_others}97.6/99.2/96.4 & \cellcolor{tab_others}97.9/99.2/96.6  & \cellcolor{tab_others}\underline{98.5}/\underline{99.8}/\underline{97.7} & \cellcolor{ours!50}\textbf{100.}/\textbf{100.}/\textbf{100.} & \cellcolor{ours}\textbf{100.}/\textbf{100.}/\textbf{100.} \\
    \multicolumn{1}{c}{} & \multicolumn{1}{c}{Leather} & \textbf{100.}/\textbf{100.}/\textbf{100.} & \textbf{100.}/\textbf{100.}/\textbf{100.} & \textbf{100.}/\textbf{100.}/\textbf{100.} & 99.2/\underline{99.8}/\underline{98.9}  & \underline{99.8}/99.7/97.6 & \cellcolor{ours!50}\textbf{100.}/\textbf{100.}/\textbf{100.} & \cellcolor{ours}\textbf{100.}/\textbf{100.}/\textbf{100.} \\
    \multicolumn{1}{c}{} & \multicolumn{1}{c}{\cellcolor{tab_others}Tile}  & \cellcolor{tab_others}\underline{98.3}/99.3/96.4 & \cellcolor{tab_others}\textbf{99.3}/\underline{99.8}/98.2 & \cellcolor{tab_others}\textbf{99.3}/\underline{99.8}/\textbf{98.8} & \cellcolor{tab_others}97.0/98.9/95.3  & \cellcolor{tab_others}96.8/\textbf{99.9}/\underline{98.4} & \cellcolor{ours!50}98.2/99.3/95.4 & \cellcolor{ours}99.1/99.8/96.2 \\
    \multicolumn{1}{c}{} & \multicolumn{1}{c}{Wood}  & 99.2/\underline{99.8}/98.3 & 98.6/99.6/96.6 & 98.4/99.5/96.7 & \textbf{99.9}/\textbf{100.}/\underline{99.2}  & \underline{99.7}/\textbf{100.}/\textbf{100.} & \cellcolor{ours!50}98.8/99.6/96.6 & \cellcolor{ours}99.7/\textbf{100.}/97.3 \\
    \hline
    \multicolumn{2}{c}{\cellcolor{tab_others}Mean} & \cellcolor{tab_others}94.6/96.5/95.2 & \cellcolor{tab_others}96.5/98.8/96.2 & \cellcolor{tab_others}95.3/98.4/95.8 & \cellcolor{tab_others}89.2/95.5/91.6  & \cellcolor{tab_others}\underline{97.2}/\underline{99.0}/\underline{96.5} & \cellcolor{ours!50}98.6/99.6/97.8 & \cellcolor{ours}\textbf{99.1}/\textbf{99.8}/\textbf{98.3} \\
    \bottomrule
   
    \end{tabular}%
  }
  \label{tab:mvtecsp}%
\end{table*}

\begin{table*}[t!]
  \centering
  \caption{Comparison with SoTA methods on \textbf{MVTec-AD} dataset for multi-class anomaly localization with AU-ROC/AP/F1\_max/AU-PRO metrics.}
  \resizebox{1\linewidth}{!}{
    \begin{tabular}{ccccccccc}
    \toprule
    \multicolumn{2}{c}{Method~$\rightarrow$} & RD4AD~\cite{deng2022anomaly} & UniAD~\cite{you2022unified} & SimpleNet~\cite{liu2023simplenet}  & DeSTSeg~\cite{zhang2023destseg}  & DiAD~\cite{he2024diffusion} & \cellcolor{ours!50}MambaAD & \cellcolor{ours}MambaADv2 \\
    \cline{1-2}
    \multicolumn{2}{c}{Category~$\downarrow$} & CVPR'22 & NeurlPS'22 & CVPR'23 & CVPR'23 & AAAI'24& \cellcolor{ours!50}NeurIPS'24 & \cellcolor{ours}(Ours) \\
    \hline
    \multicolumn{1}{c}{\multirow{10}[1]{*}{\begin{turn}{-90}Objects\end{turn}}} & \multicolumn{1}{c}{Bottle} &  97.8/\underline{68.2}/67.6/\underline{94.0} & 98.1/66.0/\underline{69.2}/93.1 & 97.2/53.8/62.4/89.0   & 93.3/61.7/56.0/67.5   & \underline{98.4}/52.2/54.8/86.6 & \cellcolor{ours!50}98.8/79.7/76.7/95.2 & \cellcolor{ours}\textbf{99.0}/\textbf{82.3}/\textbf{79.2}/\textbf{96.3} \\
    \multicolumn{1}{c}{} & \multicolumn{1}{c}{\cellcolor{tab_others}Cable} & \cellcolor{tab_others}85.1/26.3/33.6/75.1 & \cellcolor{tab_others}\textbf{97.3}/39.9/45.2/\underline{86.1} & \cellcolor{tab_others}96.7/\underline{42.4}/\underline{51.2}/85.4   & \cellcolor{tab_others}89.3/37.5/40.5/49.4   & \cellcolor{tab_others}\underline{96.8}/\textbf{50.1}/\textbf{57.8}/80.5 & \cellcolor{ours!50}95.8/42.2/48.1/90.3 & \cellcolor{ours}96.0/44.9/50.7/\textbf{91.4} \\
    \multicolumn{1}{c}{} & \multicolumn{1}{c}{Capsule} & \textbf{98.8}/43.4/50.1/\textbf{94.8} & \underline{98.5}/42.7/46.5/92.1 &\underline{98.5}/35.4/44.3/84.5   & 95.8/\textbf{47.9}/\underline{48.9}/62.1  & 97.1/42.0/45.3/87.2 & \cellcolor{ours!50}98.4/\underline{43.9}/47.7/\underline{92.6} & \cellcolor{ours}98.6/46.6/\textbf{50.3}/93.6 \\
    \multicolumn{1}{c}{} & \multicolumn{1}{c}{\cellcolor{tab_others}Hazelnut} & \cellcolor{tab_others}97.9/36.2/51.6/92.7 & \cellcolor{tab_others}98.1/55.2/56.8/\underline{94.1} &\cellcolor{tab_others}\underline{98.4}/44.6/51.4/87.4  & \cellcolor{tab_others}98.2/\underline{65.8}/61.6/84.5  & \cellcolor{tab_others}98.3/\textbf{79.2}/\textbf{80.4}/91.5& \cellcolor{ours!50}99.0/63.6/\underline{64.4}/95.7 & \cellcolor{ours}\textbf{99.2}/66.3/67.0/\textbf{96.8} \\
    \multicolumn{1}{c}{} & \multicolumn{1}{c}{Metal Nut} & 93.8/62.3/65.4/\underline{91.9}  & 94.8/55.5/66.4/81.8 &\textbf{98.0}/\textbf{83.1}/79.4/85.2     & 84.2/42.0/22.8/53.0  & \underline{97.3}/30.0/38.3/90.6 & \cellcolor{ours!50}96.7/\underline{74.5}/\underline{79.1}/93.7 & \cellcolor{ours}96.9/77.1/\textbf{81.6}/\textbf{94.8} \\
    \multicolumn{1}{c}{} & \multicolumn{1}{c}{\cellcolor{tab_others}Pill} & \cellcolor{tab_others}97.5/63.4/65.2/95.8 & \cellcolor{tab_others}95.0/44.0/53.9/95.3 &\cellcolor{tab_others}96.5/\textbf{72.4}/67.7/81.9   & \cellcolor{tab_others}96.2/61.7/41.8/27.9  & \cellcolor{tab_others}95.7/46.0/51.4/89.0 & \cellcolor{ours!50}\underline{97.4}/\underline{64.0}/\underline{66.5}/\underline{95.7} & \cellcolor{ours}\textbf{97.6}/66.6/\textbf{69.0}/\textbf{96.8} \\
    \multicolumn{1}{c}{} & \multicolumn{1}{c}{Screw} & \underline{99.4}/40.2/44.7/\underline{96.8} & 98.3/28.7/37.6/95.2 & 96.5/15.9/23.2/84.0  & 93.8/19.9/25.3/47.3   & 97.9/\textbf{60.6}/\textbf{59.6}/95.0 & \cellcolor{ours!50}99.5/\underline{49.8}/\underline{50.9}/97.1 & \cellcolor{ours}\textbf{99.7}/52.5/53.5/\textbf{98.1} \\
    \multicolumn{1}{c}{} & \multicolumn{1}{c}{\cellcolor{tab_others}Toothbrush} & \cellcolor{tab_others}99.0/\underline{53.6}/58.8/\underline{92.0} & \cellcolor{tab_others}\underline{98.4}/34.9/45.7/87.9& \cellcolor{tab_others}\underline{98.4}/46.9/52.5/87.4 & \cellcolor{tab_others}96.2/52.9/58.8/30.9   & \cellcolor{tab_others}99.0/\textbf{78.7}/\textbf{72.8}/\textbf{95.0} & \cellcolor{ours!50}99.0/48.5/\underline{59.2}/91.7 & \cellcolor{ours}\textbf{99.2}/51.2/61.8/92.8 \\
    \multicolumn{1}{c}{} & \multicolumn{1}{c}{Transistor} & 85.9/42.3/45.2/74.7 & \textbf{97.9}/\underline{59.5}/\underline{64.6}/\textbf{93.5} & 95.8/58.2/56.0/83.2   & 73.6/38.4/39.2/43.9   & 95.1/15.6/31.7/\underline{90.0} & \cellcolor{ours!50}\underline{96.5}/69.4/67.1/87.0 & \cellcolor{ours}96.7/\textbf{72.1}/\textbf{69.6}/88.1 \\
    \multicolumn{1}{c}{} & \multicolumn{1}{c}{\cellcolor{tab_others}Zipper}  & \cellcolor{tab_others}98.5/53.9/\underline{60.3}/\underline{94.1} &\cellcolor{tab_others} 96.8/40.1/49.9/92.6 & \cellcolor{tab_others}97.9/53.4/54.6/90.7   & \cellcolor{tab_others}97.3/\textbf{64.7}/59.2/66.9   & \cellcolor{tab_others}96.2/\underline{60.7}/60.0/91.6 & \cellcolor{ours!50}\underline{98.4}/60.4/61.7/94.3 & \cellcolor{ours}\textbf{98.6}/63.1/\textbf{64.3}/\textbf{95.3} \\
    \hline
    \multicolumn{1}{c}{\multirow{5}[1]{*}{\begin{turn}{-90}Textures\end{turn}}} & \multicolumn{1}{c}{Carpet} & \underline{99.0}/58.5/\underline{60.5}/\underline{95.1} & 98.5/49.9/51.1/94.4 & 97.4/38.7/43.2/90.6   & 93.6/\underline{59.9}/58.9/89.3  & 98.6/42.2/46.4/90.6 & \cellcolor{ours!50}99.2/60.0/63.3/96.7 & \cellcolor{ours}\textbf{99.4}/\textbf{62.6}/\textbf{65.8}/\textbf{97.8} \\
    \multicolumn{1}{c}{} & \multicolumn{1}{c}{\cellcolor{tab_others}Grid}  & \cellcolor{tab_others}99.2/46.0/47.4/97.0  & \cellcolor{tab_others}96.5/23.0/28.4/92.9 & \cellcolor{tab_others}96.8/20.5/27.6/88.6/  & \cellcolor{tab_others}\underline{97.0}/42.1/46.9/86.8  & \cellcolor{tab_others}96.6/\textbf{66.0}/\textbf{64.1}/\underline{94.0} & \cellcolor{ours!50}99.2/\underline{47.4}/\underline{47.7}/97.0 & \cellcolor{ours}\textbf{99.4}/50.1/50.3/\textbf{98.0} \\
    \multicolumn{1}{c}{} & \multicolumn{1}{c}{Leather} & 99.3/38.0/45.1/\underline{97.4} & 98.8/32.9/34.4/96.8 &98.7/28.5/32.9/92.7  & 99.5/\textbf{71.5}/\textbf{66.5}/91.1   & 98.8/\underline{56.1}/\underline{62.3}/91.3 & \cellcolor{ours!50}\underline{99.4}/50.3/53.3/98.7 & \cellcolor{ours}\textbf{99.6}/52.9/55.8/\textbf{99.8} \\
    \multicolumn{1}{c}{} & \multicolumn{1}{c}{\cellcolor{tab_others}Tile}  & \cellcolor{tab_others}\underline{95.3}/48.5/60.5/85.8 & \cellcolor{tab_others}91.8/42.1/50.6/78.4 & \cellcolor{tab_others}\textbf{95.7}/60.5/59.9/\underline{90.6}  & \cellcolor{tab_others}93.0/\textbf{71.0}/\textbf{66.2}/87.1  & \cellcolor{tab_others}92.4/\underline{65.7}/\underline{64.1}/\textbf{90.7} & \cellcolor{ours!50}93.8/45.1/54.8/80.0 & \cellcolor{ours}94.0/47.8/57.3/81.1 \\
    \multicolumn{1}{c}{} & \multicolumn{1}{c}{Wood} & \underline{95.3}/\underline{47.8}/\underline{51.0}/90.0 & 93.2/37.2/41.5/86.7 &91.4/34.8/39.7/76.3  & \textbf{95.9}/\textbf{77.3}/\textbf{71.3}/83.4   & 93.3/43.3/43.5/\textbf{97.5} & \cellcolor{ours!50}94.4/46.2/48.2/\underline{91.2} & \cellcolor{ours}94.6/48.9/50.8/92.3 \\
    \hline
    \multicolumn{2}{c}{\cellcolor{tab_others}Mean} & \cellcolor{tab_others}96.1/48.6/53.8/\underline{91.1} & \cellcolor{tab_others}96.8/43.4/49.5/90.7 & \cellcolor{tab_others}\underline{96.9}/45.9/49.7/86.5    & \cellcolor{tab_others}93.1/\underline{54.3}/50.9/64.8   & \cellcolor{tab_others}96.8/52.6/\underline{55.5}/90.7 & \cellcolor{ours!50}97.7/56.3/59.2/93.1 & \cellcolor{ours}\textbf{97.9}/\textbf{59.0}/\textbf{61.8}/\textbf{94.2} \\
    \bottomrule
    \end{tabular}
  }
  \label{tab:mvtecpx}%
\end{table*}

\section*{More Quantitative Results for Each Category on The VisA Dataset.}
Tab. \ref{tab:visasp} and Tab. \ref{tab:visapx} respectively present the results of image-level anomaly detection and pixel-level anomaly localization quantitative outcomes across all categories within the VisA dataset. The results further demonstrate the superiority of our method over various SoTA approaches.

\begin{table*}[t!]
  \centering
   \caption{Comparison with SoTA methods on \textbf{VisA} dataset for multi-class anomaly detection with AU-ROC/AP/F1\_max metrics.}
  \resizebox{1.\linewidth}{!}{
    \begin{tabular}{cccccccc}
    \toprule
    Method~$\rightarrow$ & RD4AD~\cite{deng2022anomaly} & UniAD~\cite{you2022unified} & SimpleNet~\cite{liu2023simplenet}  & DeSTSeg~\cite{zhang2023destseg}  & DiAD~\cite{he2024diffusion} & \cellcolor{ours!50}MambaAD & \cellcolor{ours}MambaADv2 \\
    \cline{1-1}
    Category~$\downarrow$ & CVPR'22 & NeurlPS'22 & CVPR'23 & CVPR'23 & AAAI'24& \cellcolor{ours!50}NeurIPS'24 & \cellcolor{ours}(Ours) \\
    \hline
    pcb1  & 96.2/\textbf{95.5}/91.9 & 92.8/92.7/87.8 & 91.6/91.9/86.0 & 87.6/83.1/83.7 & 88.1/88.7/80.7 & \cellcolor{ours!50}\underline{95.4}/\underline{93.0}/\underline{91.6} & \cellcolor{ours}\textbf{96.4}/94.2/\textbf{93.1} \\
    \cellcolor{tab_others}pcb2  & \cellcolor{tab_others}\textbf{97.8}/\textbf{97.8}/\textbf{94.2}   & \cellcolor{tab_others}87.8/87.7/83.1 &\cellcolor{tab_others}92.4/93.3/84.5  &\cellcolor{tab_others}86.5/85.8/82.6 & \cellcolor{tab_others}91.4/91.4/84.7 & \cellcolor{ours!50}\underline{94.2}/\underline{93.7}/\underline{89.3} & \cellcolor{ours}95.2/94.8/90.9 \\
    pcb3  & \textbf{96.4}/\textbf{96.2}/\textbf{91.0}   & 78.6/78.6/76.1 & 89.1/91.1/82.6 &\underline{93.7}/\underline{95.1}/\underline{87.0}  & 86.2/87.6/77.6 & \cellcolor{ours!50}\underline{93.7}/94.1/86.7 & \cellcolor{ours}94.8/95.3/88.3 \\
    \cellcolor{tab_others}pcb4  & \cellcolor{tab_others}99.9/99.9/99.0  & \cellcolor{tab_others}98.8/98.8/94.3 & \cellcolor{tab_others}97.0/97.0/93.5 &\cellcolor{tab_others}97.8/97.8/92.7 &\cellcolor{tab_others}\underline{99.6}/\underline{99.5}/97.0 & \cellcolor{ours!50}99.9/99.9/\underline{98.5} & \cellcolor{ours}\textbf{100.}/\textbf{100.}/\textbf{100.} \\
    \hline
    macaroni1 & 75.9/61.5/76.8   & 79.9/79.8/72.7 & \underline{85.9}/82.5/73.1 & 76.6/69.0/71.0 & 85.7/\underline{85.2}/\underline{78.8} & \cellcolor{ours!50}91.6/89.8/81.6 & \cellcolor{ours}\textbf{92.6}/\textbf{91.0}/\textbf{83.2} \\
    \cellcolor{tab_others}macaroni2 & \cellcolor{tab_others}\textbf{88.3}/\textbf{84.5}/\textbf{83.8}   & \cellcolor{tab_others}71.6/71.6/69.9 &\cellcolor{tab_others}68.3/54.3/59.7 & \cellcolor{tab_others}68.9/62.1/67.7 & \cellcolor{tab_others}62.5/57.4/69.6 & \cellcolor{ours!50}\underline{81.6}/\underline{78.0}/\underline{73.8} & \cellcolor{ours}82.6/79.2/75.4 \\
    capsules & 82.2/90.4/81.3   & 55.6/55.6/76.9  &74.1/82.8/74.6  &\underline{87.1}/\underline{93.0}/\underline{84.2}  & 58.2/69.0/78.5 & \cellcolor{ours!50}91.8/95.0/88.8 & \cellcolor{ours}\textbf{92.9}/\textbf{96.2}/\textbf{90.4} \\
    \cellcolor{tab_others}candle & \cellcolor{tab_others}92.3/92.9/86.0 & \cellcolor{tab_others}94.1/94.0/86.1 & \cellcolor{tab_others}84.1/73.3/76.6 & \cellcolor{tab_others}\underline{94.9}/\underline{94.8}/\underline{89.2} & \cellcolor{tab_others}92.8/92.0/87.6 & \cellcolor{ours!50}96.8/96.9/90.1 & \cellcolor{ours}\textbf{97.8}/\textbf{98.1}/\textbf{91.6} \\
    \hline
    cashew & 92.0/95.8/90.7  & \underline{92.8}/92.8/91.4 & 88.0/91.3/84.7 &92.0/\underline{96.1}/88.1  & 91.5/95.7/89.7 & \cellcolor{ours!50}94.5/97.3/\underline{91.1} & \cellcolor{ours}\textbf{95.5}/\textbf{98.4}/\textbf{92.6} \\
    \cellcolor{tab_others}chewinggum &\cellcolor{tab_others}94.9/97.5/92.1  &\cellcolor{tab_others} 96.3/96.2/\underline{95.2} & \cellcolor{tab_others}96.4/98.2/93.8 & \cellcolor{tab_others}95.8/98.3/94.7 & \cellcolor{tab_others}\textbf{99.1}/99.5/\textbf{95.9} & \cellcolor{ours!50}\underline{97.7}/\underline{98.9}/94.2 & \cellcolor{ours}98.7/\textbf{100.}/95.8 \\
    fryum &95.3/97.9/91.5   & 83.0/83.0/85.0 & 88.4/93.0/83.3 & 92.1/96.1/89.5 & 89.8/95.0/87.2 & \cellcolor{ours!50}\underline{95.2}/\underline{97.7}/\underline{90.5} & \cellcolor{ours}\textbf{96.2}/\textbf{98.8}/\textbf{92.1} \\
    \cellcolor{tab_others}pipe\_fryum &\cellcolor{tab_others}\underline{97.9}/\underline{98.9}/\underline{96.5}  & \cellcolor{tab_others}94.7/94.7/93.9 &\cellcolor{tab_others}90.8/95.5/88.6  & \cellcolor{tab_others}94.1/97.1/91.9 & \cellcolor{tab_others}96.2/98.1/93.7 & \cellcolor{ours!50}98.7/99.3/97.0 & \cellcolor{ours}\textbf{99.7}/\textbf{100.}/\textbf{98.6} \\
    \hline
    Mean  & \underline{92.4}/\underline{92.4}/89.6   & 85.5/85.5/84.4 & 87.2/87.0/81.8 & 88.9/89.0/85.2 & 86.8/88.3/85.1 & \cellcolor{ours!50}94.3/94.5/\underline{89.4} & \cellcolor{ours}\textbf{95.2}/\textbf{95.5}/\textbf{91.0} \\
    \bottomrule
    \end{tabular}
  }
  \label{tab:visasp}%
\end{table*}

\begin{table*}[t!]
  \centering
  \caption{Comparison with SoTA methods on \textbf{VisA} dataset for multi-class anomaly localization with AU-ROC/AP/F1\_max/AU-PRO metrics.}
  \resizebox{1\linewidth}{!}{
    \begin{tabular}{cccccccc}
    \toprule
    Method~$\rightarrow$ & RD4AD~\cite{deng2022anomaly} & UniAD~\cite{you2022unified} & SimpleNet~\cite{liu2023simplenet}  & DeSTSeg~\cite{zhang2023destseg}  & DiAD~\cite{he2024diffusion} & \cellcolor{ours!50}MambaAD & \cellcolor{ours}MambaADv2 \\
    \cline{1-1}
    Category~$\downarrow$ & CVPR'22 & NeurlPS'22 & CVPR'23 & CVPR'23 & AAAI'24& \cellcolor{ours!50}NeurIPS'24 & \cellcolor{ours}(Ours) \\
    \midrule
    pcb1  & \underline{99.4}/66.2/62.4/\textbf{95.8}  & 93.3/ \;3.9/ \;8.3/64.1 & 99.2/\textbf{86.1}/\textbf{78.8}/83.6 & 95.8/46.4/49.0/83.2  & 98.7/49.6/52.8/80.2 & \cellcolor{ours!50}99.8/\underline{77.1}/\underline{72.4}/\underline{92.8} & \cellcolor{ours}\textbf{100.}/79.5/74.9/93.6 \\
    \cellcolor{tab_others}pcb2  & \cellcolor{tab_others}\underline{98.0}/\textbf{22.3}/\textbf{30.0}/\textbf{90.8}  & \cellcolor{tab_others}93.9/ \;4.2/ \;9.2/66.9 & \cellcolor{tab_others}96.6/ \;8.9/18.6/85.7  & \cellcolor{tab_others}97.3/\underline{14.6}/\underline{28.2}/79.9 & \cellcolor{tab_others}95.2/ \;7.5/16.7/67.0  & \cellcolor{ours!50}98.9/13.3/23.4/\underline{89.6} & \cellcolor{ours}\textbf{99.1}/15.7/25.9/90.4 \\
    pcb3  & \underline{97.9}/26.2/\underline{35.2}/\textbf{93.9}  & 97.3/13.8/21.9/70.6 & 97.2/\textbf{31.0}/\textbf{36.1}/85.1  & 97.7/\underline{28.1}/33.4/62.4 & 96.7/ \;8.0/18.8/68.9  & \cellcolor{ours!50}99.1/18.3/27.4/\underline{89.1} & \cellcolor{ours}\textbf{99.3}/20.7/29.9/89.9 \\
    \cellcolor{tab_others}pcb4  &\cellcolor{tab_others}\underline{97.8}/31.4/37.0/\textbf{88.7} & \cellcolor{tab_others}94.9/14.7/22.9/72.3 &\cellcolor{tab_others}93.9/23.9/32.9/61.1  &\cellcolor{tab_others}95.8/\textbf{53.0}/\textbf{53.2}/76.9 & \cellcolor{tab_others}97.0/17.6/27.2/85.0  & \cellcolor{ours!50}98.6/\underline{47.0}/\underline{46.9}/\underline{87.6} & \cellcolor{ours}\textbf{98.8}/49.5/49.5/88.4 \\
    \midrule
    macaroni1 & \underline{99.4}/ \;2.9/ \;6.9/95.3 & 97.4/ \;3.7/ \;9.7/84.0 & 98.9/ \;3.5/ \;8.4/92.0  &99.1/ \;5.8/13.4/62.4  & 94.1/\underline{10.2}/\underline{16.7}/68.5  & \cellcolor{ours!50}99.5/17.5/27.6/\underline{95.2} & \cellcolor{ours}\textbf{99.7}/\textbf{20.0}/\textbf{30.1}/\textbf{96.0} \\
    \cellcolor{tab_others}macaroni2 & \cellcolor{tab_others}\textbf{99.7}/\textbf{13.2}/\textbf{21.8}/\textbf{97.4}  & \cellcolor{tab_others}95.2/ \;0.9/ \;4.3/76.6 & \cellcolor{tab_others}93.2/ \;0.6/ \;3.9/77.8  & \cellcolor{tab_others}98.5/ \;6.3/14.4/70.0 & \cellcolor{tab_others}93.6/ \;0.9/ \;2.8/73.1  & \cellcolor{ours!50}\underline{99.5}/ \;\underline{9.2}/\underline{16.1}/\underline{96.2} & \cellcolor{ours}\textbf{99.7}/11.6/18.6/97.0 \\
    capsules &\textbf{99.4}/\underline{60.4}/60.8/\textbf{93.1} & 88.7/ \;3.0/ \;7.4/43.7 &97.1/52.9/53.3/73.7  &96.9/33.2/39.1/76.7  & 97.3/10.0/21.0/77.9  & \cellcolor{ours!50}\underline{99.1}/61.3/\underline{59.8}/\underline{91.8} & \cellcolor{ours}99.3/\textbf{63.7}/\textbf{62.3}/92.6 \\
    \cellcolor{tab_others}candle & \cellcolor{tab_others}99.1/\underline{25.3}/\underline{35.8}/\underline{94.9}  & \cellcolor{tab_others}98.5/17.6/27.9/91.6 & \cellcolor{tab_others}97.6/ \;8.4/16.5/87.6  & \cellcolor{tab_others}98.7/\textbf{39.9}/\textbf{45.8}/69.0 & \cellcolor{tab_others}97.3/12.8/22.8/89.4  & \cellcolor{ours!50}\underline{99.0}/23.2/32.4/95.5 & \cellcolor{ours}\textbf{99.2}/25.6/35.0/\textbf{96.3} \\
    \midrule
    cashew & 91.7/44.2/49.7/86.2 & \underline{98.6}/51.7/58.3/87.9 & \textbf{98.9}/\textbf{68.9}/\textbf{66.0}/84.4  &87.9/47.6/52.1/66.3  & 90.9/\underline{53.1}/\underline{60.9}/61.8  & \cellcolor{ours!50}94.3/46.8/51.4/\underline{87.8} & \cellcolor{ours}94.5/49.2/54.0/\textbf{88.6} \\
    \cellcolor{tab_others}chewinggum & \cellcolor{tab_others}\underline{98.7}/\underline{59.9}/\underline{61.7}/76.9 & \cellcolor{tab_others}\textbf{98.8}/54.9/56.1/\textbf{81.3} & \cellcolor{tab_others}97.9/26.8/29.8/78.3  &\cellcolor{tab_others}\textbf{98.8}/\textbf{86.9}/\textbf{81.0}/68.3 & \cellcolor{tab_others}94.7/11.9/25.8/59.5  & \cellcolor{ours!50}98.1/57.5/59.9/\underline{79.7} & \cellcolor{ours}98.3/59.9/62.4/80.5 \\
    fryum & \underline{97.0}/47.6/51.5/\textbf{93.4} & 95.9/34.0/40.6/76.2 &93.0/39.1/45.4/85.1  & 88.1/35.2/38.5/47.7 & \textbf{97.6}/\textbf{58.6}/\textbf{60.1}/81.3  & \cellcolor{ours!50}96.9/\underline{47.8}/\underline{51.9}/\underline{91.6} & \cellcolor{ours}97.2/50.2/54.4/92.4 \\
    \cellcolor{tab_others}pipe\_fryum &\cellcolor{tab_others}\underline{99.1}/56.8/58.8/95.4  & \cellcolor{tab_others}98.9/50.2/57.7/91.5 & \cellcolor{tab_others}98.5/65.6/63.4/83.0  &\cellcolor{tab_others}98.9/\textbf{78.8}/\textbf{72.7}/45.9 & \cellcolor{tab_others}\textbf{99.4}/\underline{72.7}/\underline{69.9}/89.9  & \cellcolor{ours!50}\underline{99.1}/53.5/58.5/\underline{95.1} & \cellcolor{ours}99.3/56.0/61.0/\textbf{95.9} \\
    \midrule
    Mean  & \underline{98.1}/38.0/42.6/\textbf{91.8}  & 95.9/21.0/27.0/75.6 & 96.8/34.7/37.8/81.4  &96.1/39.6/\underline{43.4}/67.4 & 96.0/26.1/33.0/75.2  & \cellcolor{ours!50}98.5/\underline{39.4}/44.0/\underline{91.0} & \cellcolor{ours}\textbf{98.7}/\textbf{41.8}/\textbf{46.5}/\textbf{91.8} \\
    \bottomrule
    \end{tabular}
  }
  \label{tab:visapx}%
\end{table*}

\section*{More Quantitative Results for Each Category on The MVTec-3D Dataset.}
Tab. \ref{tab:mvtec3dsp} and Tab. \ref{tab:mvtec3dpx} respectively present the results of image-level anomaly detection and pixel-level anomaly localization quantitative outcomes across all categories within the MVTec-3D dataset. The results further demonstrate the superiority of our method over various SoTA approaches.

\begin{table*}[t!]
  \centering
    \caption{Comparison with SoTA methods on \textbf{MVTec-3D} dataset for multi-class anomaly detection with AU-ROC/AP/F1\_max metrics.}
  \resizebox{1.\linewidth}{!}{
    \begin{tabular}{cccccccc}
    \toprule
    Method~$\rightarrow$ & RD4AD~\cite{deng2022anomaly} & UniAD~\cite{you2022unified} & SimpleNet~\cite{liu2023simplenet}  & DeSTSeg~\cite{zhang2023destseg}  & DiAD~\cite{he2024diffusion} & \cellcolor{ours!50}MambaAD & \cellcolor{ours}MambaADv2 \\
    \cline{1-1}
    Category~$\downarrow$ & CVPR'22 & NeurlPS'22 & CVPR'23 & CVPR'23 & AAAI'24& \cellcolor{ours!50}NeurIPS'24 & \cellcolor{ours}(Ours) \\
    \hline
    bagel  & 82.5/95.4/89.6  & 82.7/95.8/89.3  & 76.2/93.3/89.3  & \underline{89.7}/\underline{97.4}/\underline{92.4}  & \textbf{100.}/\textbf{100.}/\textbf{100.}  & \cellcolor{ours!50}87.7/96.7/92.2 & \cellcolor{ours}89.9/97.7/93.4 \\
    \cellcolor{tab_others}cable gland  & \cellcolor{tab_others}\underline{90.1}/\underline{97.5}/92.6    & \cellcolor{tab_others}89.8/97.2/93.9  &\cellcolor{tab_others}70.3/91.0/90.2  &\cellcolor{tab_others}84.8/95.7/91.5  & \cellcolor{tab_others}68.1/91.0/92.3  & \cellcolor{ours!50}94.3/98.6/\underline{93.5} & \cellcolor{ours}\textbf{96.5}/\textbf{99.6}/\textbf{94.8} \\
    carrot  & 87.3/96.7/93.2    & 76.8/93.8/92.5 &  71.4/92.6/91.2  &79.1/94.7/91.0   & \textbf{94.4}/\textbf{99.3}/\textbf{98.0}  & \cellcolor{ours!50}\underline{90.7}/\underline{97.7}/\underline{95.0} & \cellcolor{ours}92.9/98.7/96.2 \\
    \cellcolor{tab_others}cookie  & \cellcolor{tab_others}46.0/77.4/88.0   & \cellcolor{tab_others}\textbf{77.3}/\textbf{93.5}/88.0  & \cellcolor{tab_others}66.7/89.6/\underline{88.4}  &\cellcolor{tab_others}67.4/\underline{89.8}/88.0  &\cellcolor{tab_others}\underline{69.4}/78.8/\textbf{90.9}  & \cellcolor{ours!50}61.2/87.5/\underline{88.4} & \cellcolor{ours}63.5/88.5/89.7 \\
    dowel & 96.7/98.9/97.6    & 96.7/99.3/96.2  & 83.7/95.1/91.7  & 77.3/94.3/88.9  & 98.0/\underline{99.3}/\underline{97.3}  & \cellcolor{ours!50}\underline{97.6}/99.5/96.6 & \cellcolor{ours}\textbf{99.8}/\textbf{100.}/\textbf{97.8} \\
    \cellcolor{tab_others}foam & \cellcolor{tab_others}74.3/92.9/\underline{90.6}    & \cellcolor{tab_others}70.5/92.4/88.9  &\cellcolor{tab_others}77.4/94.2/89.7  & \cellcolor{tab_others}77.9/94.7/88.9  & \cellcolor{tab_others}\textbf{100.}/\textbf{100.}/\textbf{100.}  & \cellcolor{ours!50}\underline{84.0}/\underline{95.8}/90.4 & \cellcolor{ours}86.3/96.9/91.7 \\
    peach & 64.3/84.8/90.6   & 70.0/91.0/90.5  &62.0/86.9/89.7   &\underline{82.2}/\underline{95.3}/\underline{90.7}   & 58.0/91.3/94.3 & \cellcolor{ours!50}92.8/98.1/94.3 & \cellcolor{ours}\textbf{95.0}/\textbf{99.1}/\textbf{95.5} \\
    \cellcolor{tab_others}potato & \cellcolor{tab_others}62.5/88.5/90.5  & \cellcolor{tab_others}51.6/81.8/89.3  & \cellcolor{tab_others}56.7/82.2/89.8  & \cellcolor{tab_others}\underline{62.9}/87.8/\underline{90.9}  & \cellcolor{tab_others}\textbf{76.3}/\textbf{94.3}/\textbf{95.0}  & \cellcolor{ours!50}66.8/\underline{88.6}/90.5 & \cellcolor{ours}69.1/89.6/91.8 \\
    rope & \underline{96.3}/98.5/93.2   & 97.4/99.0/95.5  & 95.6/98.4/\underline{94.7}  &93.5/97.4/92.5   & 89.2/95.4/91.9  & \cellcolor{ours!50}97.4/\underline{98.9}/\underline{94.7} & \cellcolor{ours}\textbf{99.7}/\textbf{99.9}/\textbf{95.9} \\
    \cellcolor{tab_others}tire &\cellcolor{tab_others}79.2/93.0/88.4 &\cellcolor{tab_others}75.7/90.7/89.7  & \cellcolor{tab_others}65.3/86.8/87.9  & \cellcolor{tab_others}80.8/93.6/91.5  & \cellcolor{tab_others}\textbf{92.7}/\textbf{98.9}/\textbf{95.8} & \cellcolor{ours!50}\underline{90.0}/\underline{97.0}/\underline{91.9} & \cellcolor{ours}92.3/98.0/93.2 \\
    \hline
    Mean  & 77.9/92.4/91.4    & 78.9/93.4/91.4  & 72.5/91.0/90.3  & 79.6/94.1/90.6  & \underline{84.6}/\underline{94.8}/\textbf{95.6} & \cellcolor{ours!50}86.2/95.8/\underline{92.8} & \cellcolor{ours}\textbf{88.5}/\textbf{96.8}/94.0 \\
    \bottomrule
    \end{tabular}
  }
  \label{tab:mvtec3dsp}%
\end{table*}

\begin{table*}[t!]
  \centering
    \caption{Comparison with SoTA methods on \textbf{MVTec-3D} dataset for multi-class anomaly localization with AU-ROC/AP/F1\_max/AU-PRO metrics.}
  \resizebox{1.\linewidth}{!}{
    \begin{tabular}{cccccccc}
    \toprule
    Method~$\rightarrow$ & RD4AD~\cite{deng2022anomaly} & UniAD~\cite{you2022unified} & SimpleNet~\cite{liu2023simplenet}  & DeSTSeg~\cite{zhang2023destseg}  & DiAD~\cite{he2024diffusion} & \cellcolor{ours!50}MambaAD & \cellcolor{ours}MambaADv2 \\
    \cline{1-1}
    Category~$\downarrow$ & CVPR'22 & NeurlPS'22 & CVPR'23 & CVPR'23 & AAAI'24& \cellcolor{ours!50}NeurIPS'24 & \cellcolor{ours}(Ours) \\
    \hline
    bagel  & \underline{98.6}/39.0/45.1/91.3  &97.6/30.4/35.8/84.4   & 93.2/23.6/30.9/70.4   & 98.7/\textbf{53.0}/\underline{52.9}/77.6   & 98.5/\underline{49.6}/\textbf{54.2}/\textbf{93.8}  & \cellcolor{ours!50}98.5/38.3/41.1/\underline{92.1} & \cellcolor{ours}\textbf{98.8}/41.3/44.2/93.3 \\
    \cellcolor{tab_others}cable gland  & \cellcolor{tab_others}99.4/37.9/43.2/\underline{98.2}    & \cellcolor{tab_others}\underline{98.9}/26.4/34.8/96.3  &\cellcolor{tab_others}95.2/14.4/23.1/86.8   &\cellcolor{tab_others}97.8/\textbf{46.0}/\textbf{49.8}/64.1   & \cellcolor{tab_others}98.4/25.2/32.0/94.5  & \cellcolor{ours!50}99.4/\underline{39.5}/\underline{43.9}/98.4 & \cellcolor{ours}\textbf{99.7}/42.5/47.0/\textbf{99.6} \\
    carrot  & 99.4/\underline{27.5}/\underline{33.7}/\underline{97.2}   & 98.0/12.2/19.3/93.4  &  96.4/13.8/21.2/84.4   &86.9/26.5/21.7/14.2    & 98.6/20.0/26.9/94.6  & \cellcolor{ours!50}99.4/30.1/35.4/98.1 & \cellcolor{ours}\textbf{99.7}/\textbf{33.1}/\textbf{38.5}/\textbf{99.2} \\
    \cellcolor{tab_others}cookie  & \cellcolor{tab_others}96.6/27.5/32.9/\underline{86.6}   & \cellcolor{tab_others}\textbf{97.5}/40.4/\textbf{45.6}/\textbf{88.7}   & \cellcolor{tab_others}90.5/26.7/31.4/66.6   &\cellcolor{tab_others}93.3/34.0/35.6/40.9  &\cellcolor{tab_others}94.3/14.0/23.8/83.5 & \cellcolor{ours!50}\underline{96.8}/\underline{39.0}/\underline{41.9}/83.6 & \cellcolor{ours}97.1/\textbf{42.0}/45.0/84.8 \\
    dowel & 99.7/\underline{47.7}/50.8/\textbf{98.8}    &99.1/32.1/37.7/96.1   & 95.3/17.4/25.6/83.0  & 97.3/43.1/44.5/31.2  & 97.2/31.4/40.1/89.6   & \cellcolor{ours!50}\underline{99.6}/49.9/\underline{50.4}/\underline{97.1} & \cellcolor{ours}\textbf{99.8}/\textbf{52.9}/\textbf{53.5}/98.2 \\
    \cellcolor{tab_others}foam & \cellcolor{tab_others}94.2/15.0/26.4/\underline{79.9}    & \cellcolor{tab_others}82.2/ \;6.8/18.9/55.8   &\cellcolor{tab_others}87.8/15.7/26.7/66.7   & \cellcolor{tab_others}\textbf{95.7}/\textbf{43.7}/\textbf{49.3}/63.6   & \cellcolor{tab_others}89.8/ \;9.6/23.5/69.1   & \cellcolor{ours!50}\underline{95.1}/\underline{23.4}/\underline{32.8}/82.7 & \cellcolor{ours}95.4/26.4/35.9/\textbf{83.8} \\
    peach & \underline{98.5}/15.5/22.7/93.2   & 97.4/11.7/17.9/90.4   &92.9/ \;8.1/15.0/74.8   &95.9/\underline{35.7}/\underline{41.2}/48.2   & 98.4/27.6/31.3/\underline{94.2}   & \cellcolor{ours!50}99.4/43.2/45.1/97.1 & \cellcolor{ours}\textbf{99.7}/\textbf{46.1}/\textbf{48.2}/\textbf{98.3} \\
    \cellcolor{tab_others}potato & \cellcolor{tab_others}99.1/\underline{14.9}/\underline{22.5}/\textbf{95.9}  & \cellcolor{tab_others}97.6/ \;5.1/ \;8.9/91.1   & \cellcolor{tab_others}91.0/ \;4.3/10.9/72.8   & \cellcolor{tab_others}89.2/ \;8.7/12.2/ \;6.2   & \cellcolor{tab_others}98.0/ \;8.6/17.8/93.9  & \cellcolor{ours!50}\underline{99.0}/17.6/22.6/\underline{94.8} & \cellcolor{ours}\textbf{99.3}/\textbf{20.6}/\textbf{25.7}/\textbf{95.9} \\
    rope & 99.6/50.3/55.9/\textbf{97.9}   &99.0/34.5/40.7/94.3   & 99.3/51.1/52.9/92.8  &98.8/\textbf{64.5}/\textbf{62.1}/90.4  & 99.3/\underline{61.0}/\underline{59.9}/\underline{96.5}  & \cellcolor{ours!50}\underline{99.4}/52.1/50.9/95.5 & \cellcolor{ours}\textbf{99.7}/55.1/54.0/96.7 \\
    \cellcolor{tab_others}tire &\cellcolor{tab_others}\underline{99.2}/23.2/\underline{31.1}/\underline{96.4} &\cellcolor{tab_others}98.0/11.9/20.3/90.6   & \cellcolor{tab_others}93.8/ \;8.1/15.3/77.9  & \cellcolor{tab_others}97.0/\underline{25.8}/30.0/27.3  & \cellcolor{tab_others}91.8/ \;5.9/13.7/68.8   & \cellcolor{ours!50}99.5/42.0/46.9/97.0 & \cellcolor{ours}\textbf{99.8}/\textbf{45.0}/\textbf{50.0}/\textbf{98.2} \\
    \hline
    Mean  & \underline{98.4}/29.8/36.4/\underline{93.5}    & 96.5/21.2/28.0/88.1  & 93.6/18.3/25.3/77.6  & 95.1/38.1/\underline{39.9}/46.4  & 96.4/25.3/32.3/87.8  & \cellcolor{ours!50}98.6/\underline{37.5}/41.1/93.6 & \cellcolor{ours}\textbf{98.9}/\textbf{40.5}/\textbf{44.2}/\textbf{94.8} \\
    \bottomrule
    \end{tabular}
  }
  \label{tab:mvtec3dpx}%
\end{table*}

\section*{More Quantitative Results for Each Category on The Uni-Medical Dataset.}
Tab. \ref{tab:medicalsp} and Tab. \ref{tab:medicalpx} respectively present the results of image-level anomaly detection and pixel-level anomaly localization quantitative outcomes across all categories within the Uni-Medical dataset. The results further demonstrate the superiority of our method over various SoTA approaches.

\begin{table*}[t!]
  \centering
    \caption{Comparison with SoTA methods on \textbf{Uni-Medical} dataset for multi-class anomaly detection with AU-ROC/AP/F1\_max metrics.}
  \resizebox{1.\linewidth}{!}{
    \begin{tabular}{cccccccc}
    \toprule
    Method~$\rightarrow$ & RD4AD~\cite{deng2022anomaly} & UniAD~\cite{you2022unified} & SimpleNet~\cite{liu2023simplenet}  & DeSTSeg~\cite{zhang2023destseg}  & DiAD~\cite{he2024diffusion} & \cellcolor{ours!50}MambaAD & \cellcolor{ours}MambaADv2 \\
    \cline{1-1}
    Category~$\downarrow$ & CVPR'22 & NeurlPS'22 & CVPR'23 & CVPR'23 & AAAI'24& \cellcolor{ours!50}NeurIPS'24 & \cellcolor{ours}(Ours) \\
    \hline
    brain  & 82.4/94.4/91.5  & 89.9/97.5/92.6  & 82.3/95.6/90.9  & 84.5/95.0/92.1  & \underline{93.7}/\underline{98.1}/95.0& \cellcolor{ours!50}94.2/98.6/\underline{94.5} & \cellcolor{ours}\textbf{95.8}/\textbf{100.}/\textbf{96.1} \\
    \cellcolor{tab_others}liver  & \cellcolor{tab_others}55.1/46.3/\underline{64.1}& \cellcolor{tab_others}61.0/48.8/63.2 &\cellcolor{tab_others}55.8/47.6/60.9   &\cellcolor{tab_others}\textbf{69.2}/\textbf{60.6}/64.7  & \cellcolor{tab_others}59.2/\underline{55.6}/60.9 & \cellcolor{ours!50}\underline{63.2}/53.1/64.7 & \cellcolor{ours}64.9/55.5/\textbf{66.4} \\
    retinal & \underline{89.2}/86.7/78.5    & 84.6/79.4/73.9  & 88.8/\underline{87.6}/78.6 & 88.3/83.8/79.2  & 88.3/86.6/77.7 & \cellcolor{ours!50}93.6/88.7/86.6 & \cellcolor{ours}\textbf{95.2}/\textbf{91.1}/\textbf{88.3} \\
    \cellcolor{tab_others}Mean  & \cellcolor{tab_others}75.6/75.8/78.0  & \cellcolor{tab_others}78.5/75.2/76.6  & \cellcolor{tab_others}75.6/76.9/76.8  & \cellcolor{tab_others}\underline{80.7}/\underline{79.8}/\underline{78.7}  & \cellcolor{tab_others}80.4/80.1/77.8  & \cellcolor{ours!50}83.7/80.1/82.0 & \cellcolor{ours}\textbf{85.3}/\textbf{82.2}/\textbf{83.6} \\
    \bottomrule
    \end{tabular}
  }
  \label{tab:medicalsp}%
\end{table*}

\begin{table*}[htp!]
  \centering
    \caption{Comparison with SoTA methods on \textbf{Uni-Medical} dataset for multi-class anomaly localization with AU-ROC/AP/F1\_max/AU-PRO metrics.}
  \resizebox{1.\linewidth}{!}{
    \begin{tabular}{cccccccc}
    \toprule
    Method~$\rightarrow$ & RD4AD~\cite{deng2022anomaly} & UniAD~\cite{you2022unified} & SimpleNet~\cite{liu2023simplenet}  & DeSTSeg~\cite{zhang2023destseg}  & DiAD~\cite{he2024diffusion} & \cellcolor{ours!50}MambaAD & \cellcolor{ours}MambaADv2 \\
    \cline{1-1}
    Category~$\downarrow$ & CVPR'22 & NeurlPS'22 & CVPR'23 & CVPR'23 & AAAI'24& \cellcolor{ours!50}NeurIPS'24 & \cellcolor{ours}(Ours) \\
    \hline
    brain  & 96.5/45.9/49.2/\underline{82.6}  & \underline{97.4}/\underline{55.7}/\underline{55.7}/82.4  & 94.8/42.1/42.4/73.0  & 89.3/33.0/37.0/23.3  & 95.4/42.9/36.7/80.3  & \cellcolor{ours!50}98.1/62.7/62.2/87.6 & \cellcolor{ours}\textbf{98.4}/\textbf{65.1}/\textbf{64.4}/\textbf{88.9} \\
    \cellcolor{tab_others}liver  & \cellcolor{tab_others}96.6/ \;5.7/10.3/89.9& \cellcolor{tab_others}\underline{97.1}/ \;7.8/13.7/92.7 &\cellcolor{tab_others}\textbf{97.4}/13.2/\underline{20.1}/86.3   &\cellcolor{tab_others}79.4/\textbf{21.9}/\textbf{28.5}/20.3  &\cellcolor{tab_others}\underline{97.1}/\underline{13.7}/ \;7.3/91.4 & \cellcolor{ours!50}96.9/ \;9.1/16.3/\underline{91.8} & \cellcolor{ours}97.2/11.4/18.5/\textbf{93.1} \\
    retinal & \textbf{96.4}/64.7/60.9/\textbf{86.5}    & 94.8/49.3/51.3/79.9  & 95.5/59.5/56.3/82.1 & 91.0/59.0/46.8/31.7  & 95.3/57.5/\underline{62.8}/\underline{84.1}  & \cellcolor{ours!50}\underline{95.7}/\underline{64.5}/63.4/83.1 & \cellcolor{ours}96.0/\textbf{66.9}/\textbf{65.6}/84.4 \\
    \cellcolor{tab_others}Mean  & \cellcolor{tab_others}\underline{96.5}/\underline{38.7}/40.1/\underline{86.4}  & \cellcolor{tab_others}96.4/37.6/\underline{40.2}/85.0  &\cellcolor{tab_others}95.9/38.3/39.6/80.5  & \cellcolor{tab_others}86.6/38.0/37.5/25.1  &\cellcolor{tab_others}95.9/38.0/35.6/85.4  & \cellcolor{ours!50}96.9/45.4/47.3/87.5 & \cellcolor{ours}\textbf{97.2}/\textbf{47.8}/\textbf{49.5}/\textbf{88.8} \\
    \bottomrule
    \end{tabular}
  }
  \label{tab:medicalpx}%
\end{table*}

\section*{More Quantitative Results for Each Category on The COCO-AD Dataset.}
Tab. \ref{tab:cocosp} and Tab. \ref{tab:cocopx} respectively present the results of image-level anomaly detection and pixel-level anomaly localization quantitative outcomes across all categories within the COCO-AD dataset. The results further demonstrate the superiority of our method over various SoTA approaches.
\begin{table*}[htp!]
  \centering
    \caption{Comparison with SoTA methods on \textbf{COCO-AD} dataset for multi-class anomaly detection with AU-ROC/AP/F1\_max metrics.}
  \resizebox{1.\linewidth}{!}{
    \begin{tabular}{cccccccc}
    \toprule
    Method~$\rightarrow$ & RD4AD~\cite{deng2022anomaly} & UniAD~\cite{you2022unified} & SimpleNet~\cite{liu2023simplenet}  & DeSTSeg~\cite{zhang2023destseg}  & DiAD~\cite{he2024diffusion} & \cellcolor{ours!50}MambaAD & \cellcolor{ours}MambaADv2 \\
    \cline{1-1}
    Category~$\downarrow$ & CVPR'22 & NeurlPS'22 & CVPR'23 & CVPR'23 & AAAI'24& \cellcolor{ours!50}NeurIPS'24 & \cellcolor{ours}(Ours) \\
    \hline
    0  & 65.7/81.9/85.1  & \underline{66.1}/\underline{84.0}/85.1  & 57.8/77.4/84.7  & 59.7/79.1/85.0  & 57.5/77.5/85.3 & \cellcolor{ours!50}75.3/89.8/\underline{85.2} & \cellcolor{ours}\textbf{76.7}/\textbf{91.4}/\textbf{86.8} \\
    \cellcolor{tab_others}1  & \cellcolor{tab_others}54.9/46.8/61.1& \cellcolor{tab_others}56.1/47.8/61.1 &\cellcolor{tab_others}51.2/42.3/59.0   &\cellcolor{tab_others}\underline{55.6}/47.9/\underline{61.2}  & \cellcolor{tab_others}54.4/\textbf{49.8}/62.2 & \cellcolor{ours!50}55.0/\underline{48.1}/61.0 & \cellcolor{ours}\textbf{56.4}/49.7/\textbf{62.6} \\
    2 & 59.6/39.4/51.3    & 52.3/30.8/49.5  & 60.1/38.5/50.7 & 55.8/37.6/50.1  & \underline{63.8}/\underline{43.4}/\underline{52.5} & \cellcolor{ours!50}66.9/46.4/54.6 & \cellcolor{ours}\textbf{68.3}/\textbf{48.0}/\textbf{56.2} \\
    \cellcolor{tab_others}3 & \cellcolor{tab_others}53.5/36.4/51.5  & \cellcolor{tab_others}50.1/33.5/51.2  & \cellcolor{tab_others}\underline{59.2}/39.2/\underline{52.2}  & \cellcolor{tab_others}53.5/36.5/51.2  & \cellcolor{tab_others}\textbf{60.1}/41.4/52.9  & \cellcolor{ours!50}58.4/\underline{40.5}/51.9 & \cellcolor{ours}59.8/\textbf{42.1}/\textbf{53.6} \\
     Mean & 58.4/51.1/\underline{62.3} &56.2/49.0/61.7&57.1/49.4/61.7&56.2/50.3/61.9&\underline{58.9}/\underline{53.0}/63.2 & \cellcolor{ours!50}63.9/56.2/63.2 & \cellcolor{ours}\textbf{65.3}/\textbf{57.8}/\textbf{64.8} \\
    \bottomrule
    \end{tabular}
  }
  \label{tab:cocosp}%
\end{table*}

\begin{table*}[htp!]
  \centering
    \caption{Comparison with SoTA methods on \textbf{COCO-AD} dataset for multi-class anomaly localization with AU-ROC/AP/F1\_max/AU-PRO metrics.}
  \resizebox{1.\linewidth}{!}{
    \begin{tabular}{cccccccc}
    \toprule
    Method~$\rightarrow$ & RD4AD~\cite{deng2022anomaly} & UniAD~\cite{you2022unified} & SimpleNet~\cite{liu2023simplenet}  & DeSTSeg~\cite{zhang2023destseg}  & DiAD~\cite{he2024diffusion} & \cellcolor{ours!50}MambaAD & \cellcolor{ours}MambaADv2 \\
    \cline{1-1}
    Category~$\downarrow$ & CVPR'22 & NeurlPS'22 & CVPR'23 & CVPR'23 & AAAI'24& \cellcolor{ours!50}NeurIPS'24 & \cellcolor{ours}(Ours) \\
    \hline
    0  & \underline{72.1}/30.8/\underline{38.2}/\underline{45.9}  & 70.8/29.4/36.7/36.5  & 64.0/27.4/34.4/26.8  & 61.8/21.5/27.7/23.5  & 67.0/\underline{33.4}/26.2/28.8   & \cellcolor{ours!50}75.6/38.6/41.7/46.2 & \cellcolor{ours}\textbf{77.1}/\textbf{40.1}/\textbf{43.5}/\textbf{47.7} \\
    \cellcolor{tab_others}1  & \cellcolor{tab_others}70.7/ \;6.2/11.2/\textbf{40.6}& \cellcolor{tab_others}70.0/ \;6.2/11.3/31.8 &\cellcolor{tab_others}61.4/ \;4.9/ \;8.9/33.0    &\cellcolor{tab_others}69.3/ \;6.8/11.3/27.7  &\cellcolor{tab_others}71.3/\textbf{11.8}/ \;7.8/28.8 & \cellcolor{ours!50}71.2/ \;6.6/11.3/36.6  & \cellcolor{ours}\textbf{72.8}/8.2/\textbf{13.2}/38.2 \\
    2 & \underline{68.4}/11.6/\underline{18.9}/\underline{42.9}    & 60.9/ \;7.7/14.7/27.0  & 57.4/ \;8.2/14.4/29.2 & 61.1/ \;9.8/13.9/26.3  & 68.0/\textbf{19.2}/12.2/33.2 & \cellcolor{ours!50}71.2/\underline{13.9}/21.6/44.4 & \cellcolor{ours}\textbf{72.8}/15.5/\textbf{23.4}/\textbf{45.9} \\
    \cellcolor{tab_others}3  & \cellcolor{tab_others}58.3/ \;8.4/14.2/\underline{33.4}  & \cellcolor{tab_others}59.8/ \;8.3/14.8/31.4  &\cellcolor{tab_others}55.3/ \;8.2/13.9/21.0  & \cellcolor{tab_others}51.2/ \;6.9/12.4/16.8  & \cellcolor{tab_others}\textbf{65.9}/\textbf{17.5}/10.6/32.3 & \cellcolor{ours!50}59.0/ \;\underline{8.6}/\underline{14.4}/34.7 & \cellcolor{ours}60.5/10.2/\textbf{16.3}/\textbf{36.2} \\
    Mean&67.4/14.3/\underline{20.6}/40.7&65.4/12.9/19.4/31.7&59.5/12.2/17.9/27.5&60.9/11.3/16.3/23.6&\underline{68.0}/\textbf{20.5}/14.2/30.8 & \cellcolor{ours!50}69.3/\underline{16.9}/22.2/\underline{40.5} & \cellcolor{ours}\textbf{70.8}/18.5/\textbf{24.1}/\textbf{42.0} \\
    \bottomrule
    \end{tabular}
  }
  \label{tab:cocopx}%
\end{table*}

\section*{More Quantitative Results for Each Category on The Real-IAD Dataset.}
Tab. \ref{tab:realiadsp} and Tab. \ref{tab:realiadpx} respectively present the results of image-level anomaly detection and pixel-level anomaly localization quantitative outcomes across all categories within the Real-IAD dataset. The results further demonstrate the superiority of our method over various SoTA approaches.
\begin{table*}[t!]
    \centering
    \caption{Comparison with SoTA methods on \textbf{Real-IAD} dataset for multi-class anomaly detection with AU-ROC/AP/F1\_max metrics.}
    \resizebox{1.\linewidth}{!}{
    \begin{tabular}{cccccccc}
    \toprule
    Method~$\rightarrow$ & RD4AD~\cite{deng2022anomaly} & UniAD~\cite{you2022unified} & SimpleNet~\cite{liu2023simplenet}  & DeSTSeg~\cite{zhang2023destseg}  & DiAD~\cite{he2024diffusion} & \cellcolor{ours!50}MambaAD & \cellcolor{ours}MambaADv2 \\
    \cline{1-1}
    Category~$\downarrow$ & CVPR'22 & NeurlPS'22 & CVPR'23 & CVPR'23 & AAAI'24& \cellcolor{ours!50}NeurIPS'24 & \cellcolor{ours}(Ours) \\
    \hline
    audiojack & 76.2/63.2/60.8 & \underline{81.4}/76.6/64.9 & 58.4/44.2/50.9 & 81.1/72.6/64.5 & 76.5/54.3/\underline{65.7} & \cellcolor{ours!50}84.2/\underline{76.5}/67.4 & \cellcolor{ours}\textbf{85.7}/\textbf{78.4}/\textbf{69.3} \\
    \cellcolor{tab_others}bottle cap & \cellcolor{tab_others}89.5/86.3/81.0 & \cellcolor{tab_others}92.5/91.7/81.7 & \cellcolor{tab_others}54.1/47.6/60.3 & \cellcolor{tab_others}78.1/74.6/68.1 & \cellcolor{tab_others}\underline{91.6}/\textbf{94.0}/\textbf{87.9} & \cellcolor{ours!50}92.8/\underline{92.0}/\underline{82.1} & \cellcolor{ours}\textbf{94.3}/93.9/83.9 \\
    button battery & 73.3/78.9/76.1 & 75.9/81.6/76.3 & 52.5/60.5/72.4 & \textbf{86.7}/\textbf{89.2}/\textbf{83.5} & \underline{80.5}/71.3/70.6 & \cellcolor{ours!50}79.8/\underline{85.3}/\underline{77.8} & \cellcolor{ours}81.3/87.2/79.6 \\
    \cellcolor{tab_others}end cap & \cellcolor{tab_others}79.8/\underline{84.0}/77.8 & \cellcolor{tab_others}\underline{80.9}/\textbf{86.1}/\underline{78.0} & \cellcolor{tab_others}51.6/60.8/72.9 & \cellcolor{tab_others}77.9/81.1/77.1 & \cellcolor{tab_others}\textbf{85.1}/83.4/\textbf{84.8} & \cellcolor{ours!50}78.0/82.8/77.2 & \cellcolor{ours}79.5/84.7/79.0 \\
    eraser & \underline{90.0}/\underline{88.7}/\underline{79.7} & \textbf{90.3}/\textbf{89.2}/\textbf{80.2} & 46.4/39.1/55.8 & 84.6/82.9/71.8 & 80.0/80.0/77.3 & \cellcolor{ours!50}87.5/86.2/76.1 & \cellcolor{ours}89.0/88.2/77.9 \\
    \cellcolor{tab_others}fire hood & \cellcolor{tab_others}78.3/70.1/64.5 & \cellcolor{tab_others}80.6/\underline{74.8}/66.4 & \cellcolor{tab_others}58.1/41.9/54.4 & \cellcolor{tab_others}\underline{81.7}/72.4/\underline{67.7} & \cellcolor{tab_others}\textbf{83.3}/\textbf{81.7}/\textbf{80.5} & \cellcolor{ours!50}79.3/72.5/64.8 & \cellcolor{ours}80.8/74.4/66.6 \\
    mint  & 65.8/63.1/64.8 & 67.0/66.6/64.6 & 52.4/50.3/63.7 & 58.4/55.8/63.7 & \textbf{76.7}/\textbf{76.7}/\textbf{76.0} & \cellcolor{ours!50}\underline{70.1}/\underline{70.8}/\underline{65.5} & \cellcolor{ours}71.6/72.7/67.4 \\
    \cellcolor{tab_others}mounts & \cellcolor{tab_others}\textbf{88.6}/\textbf{79.9}/74.8 & \cellcolor{tab_others}87.6/77.3/\underline{77.2} & \cellcolor{tab_others}58.7/48.1/52.4 & \cellcolor{tab_others}74.7/56.5/63.1 & \cellcolor{tab_others}75.3/74.5/\textbf{82.5}& \cellcolor{ours!50}\underline{86.8}/\underline{78.0}/73.5 & \cellcolor{ours}88.3/\textbf{79.9}/75.4 \\
    pcb   & 79.5/85.8/79.7 & 81.0/88.2/79.1 & 54.5/66.0/75.5 & 82.0/\underline{88.7}/79.6 & \underline{86.0}/85.1/85.4 & \cellcolor{ours!50}89.1/93.7/\underline{84.0} & \cellcolor{ours}\textbf{90.6}/\textbf{95.6}/\textbf{85.8} \\
    \cellcolor{tab_others}phone battery & \cellcolor{tab_others}\underline{87.5}/\underline{83.3}/\underline{77.1} & \cellcolor{tab_others}83.6/80.0/71.6 & \cellcolor{tab_others}51.6/43.8/58.0 & \cellcolor{tab_others}83.3/81.8/72.1 & \cellcolor{tab_others}82.3/77.7/75.9 & \cellcolor{ours!50}90.2/88.9/80.5 & \cellcolor{ours}\textbf{91.7}/\textbf{90.8}/\textbf{82.3} \\
    plastic nut & 80.3/68.0/64.4 & 80.0/69.2/63.7 & 59.2/40.3/51.8 & \underline{83.1}/\underline{75.4}/\underline{66.5} & 71.9/58.2/65.6 & \cellcolor{ours!50}87.1/80.7/70.7 & \cellcolor{ours}\textbf{88.6}/\textbf{82.7}/\textbf{72.5} \\
    \cellcolor{tab_others}plastic plug & \cellcolor{tab_others}81.9/74.3/68.8 & \cellcolor{tab_others}81.4/75.9/67.6 & \cellcolor{tab_others}48.2/38.4/54.6 & \cellcolor{tab_others}71.7/63.1/60.0 & \cellcolor{tab_others}\textbf{88.7}/\textbf{89.2}/\textbf{90.9} & \cellcolor{ours!50}\underline{85.7}/\underline{82.2}/\underline{72.6} & \cellcolor{ours}87.2/84.2/74.4 \\
    porcelain doll & \underline{86.3}/\underline{76.3}/\underline{71.5} & 85.1/75.2/69.3 & 66.3/54.5/52.1 & 78.7/66.2/64.3 & 72.6/66.8/65.2 & \cellcolor{ours!50}88.0/82.2/74.1 & \cellcolor{ours}\textbf{89.5}/\textbf{84.2}/\textbf{75.9} \\
    \cellcolor{tab_others}regulator & \cellcolor{tab_others}66.9/48.8/47.7 & \cellcolor{tab_others}56.9/41.5/44.5 & \cellcolor{tab_others}50.5/29.0/43.9 & \cellcolor{tab_others}\textbf{79.2}/\underline{63.5}/\underline{56.9} & \cellcolor{tab_others}\underline{72.1}/\textbf{71.4}/\textbf{78.2} & \cellcolor{ours!50}69.7/58.7/50.4 & \cellcolor{ours}71.2/60.6/52.2 \\
    rolled strip base & 97.5/98.7/94.7 & 98.7/99.3/96.5 & 59.0/75.7/79.8 & 96.5/98.2/93.0 & 68.4/55.9/56.8 & \cellcolor{ours!50}\underline{98.0}/\underline{99.0}/\underline{95.0} & \cellcolor{ours}\textbf{99.5}/\textbf{100.}/\textbf{96.8} \\
    \cellcolor{tab_others}sim card set & \cellcolor{tab_others}91.6/91.8/84.8 & \cellcolor{tab_others}89.7/90.3/83.2 & \cellcolor{tab_others}63.1/69.7/70.8 & \cellcolor{tab_others}95.5/96.2/\textbf{89.2} & \cellcolor{tab_others}72.6/53.7/61.5 & \cellcolor{ours!50}\underline{94.4}/\underline{95.1}/\underline{87.2} & \cellcolor{ours}\textbf{95.9}/\textbf{97.0}/89.0 \\
    switch & 84.3/87.2/77.9 & 85.5/88.6/78.4 & 62.2/66.8/68.6 & \underline{90.1}/\underline{92.8}/\underline{83.1} & 73.4/49.4/61.2 & \cellcolor{ours!50}91.7/94.0/85.4 & \cellcolor{ours}\textbf{93.2}/\textbf{95.9}/\textbf{87.2} \\
    \cellcolor{tab_others}tape  & \cellcolor{tab_others}96.0/95.1/87.6 & \cellcolor{tab_others}97.2/96.2/89.4 & \cellcolor{tab_others}49.9/41.1/54.5 & \cellcolor{tab_others}94.5/93.4/85.9 & \cellcolor{tab_others}73.9/57.8/66.1 & \cellcolor{ours!50}\underline{96.8}/\underline{95.9}/\underline{89.3} & \cellcolor{ours}\textbf{98.2}/\textbf{97.8}/\textbf{91.1} \\
    terminalblock & \underline{89.4}/\underline{89.7}/\underline{83.1} & 87.5/89.1/81.0 & 59.8/64.7/68.8 & 83.1/86.2/76.6 & 62.1/36.4/47.8 & \cellcolor{ours!50}96.1/96.8/90.0 & \cellcolor{ours}\textbf{97.6}/\textbf{98.7}/\textbf{91.8} \\
    \cellcolor{tab_others}toothbrush & \cellcolor{tab_others}82.0/83.8/77.2 & \cellcolor{tab_others}78.4/80.1/75.6 & \cellcolor{tab_others}65.9/70.0/70.1 & \cellcolor{tab_others}83.7/85.3/79.0 & \cellcolor{tab_others}\textbf{91.2}/\textbf{93.7}/\textbf{90.9} & \cellcolor{ours!50}\underline{85.1}/\underline{86.2}/\underline{80.3} & \cellcolor{ours}86.6/88.2/82.1 \\
    toy   & 69.4/74.2/\underline{75.9} & 68.4/\underline{75.1}/74.8 & 57.8/64.4/73.4 & \underline{70.3}/74.8/75.4 & 66.2/57.3/59.8 & \cellcolor{ours!50}83.0/87.5/79.6 & \cellcolor{ours}\textbf{84.5}/\textbf{89.4}/\textbf{81.4} \\
    \cellcolor{tab_others}toy brick & \cellcolor{tab_others}63.6/56.1/59.0 & \cellcolor{tab_others}\textbf{77.0}/\textbf{71.1}/\textbf{66.2} & \cellcolor{tab_others}58.3/49.7/58.2 & \cellcolor{tab_others}\underline{73.2}/\underline{68.7}/\underline{63.3} & \cellcolor{tab_others}68.4/45.3/55.9 & \cellcolor{ours!50}70.5/63.7/61.6 & \cellcolor{ours}72.0/65.7/63.4 \\
    transistor1 & 91.0/94.0/85.1 & \underline{93.7}/\underline{95.9}/\underline{88.9} & 62.2/69.2/72.1 & 90.2/92.1/84.6 & 73.1/63.1/62.7 & \cellcolor{ours!50}94.4/96.0/89.0 & \cellcolor{ours}\textbf{95.9}/\textbf{97.9}/\textbf{90.8} \\
    \cellcolor{tab_others}u block & \cellcolor{tab_others}\underline{89.5}/\underline{85.0}/74.2 & \cellcolor{tab_others}88.8/84.2/75.5 & \cellcolor{tab_others}62.4/48.4/51.8 & \cellcolor{tab_others}80.1/73.9/64.3 & \cellcolor{tab_others}75.2/68.4/67.9 & \cellcolor{ours!50}89.7/85.7/\underline{75.3} & \cellcolor{ours}\textbf{91.2}/\textbf{87.7}/\textbf{77.1} \\
    usb   & 84.9/84.3/75.1 & 78.7/79.4/69.1 & 57.0/55.3/62.9 & \underline{87.8}/\underline{88.0}/\underline{78.3} & 58.9/37.4/45.7 & \cellcolor{ours!50}92.0/92.2/84.5 & \cellcolor{ours}\textbf{93.5}/\textbf{94.1}/\textbf{86.3} \\
    \cellcolor{tab_others}usb adaptor & \cellcolor{tab_others}71.1/61.4/62.2 & \cellcolor{tab_others}76.8/71.3/64.9 & \cellcolor{tab_others}47.5/38.4/56.5 & \cellcolor{tab_others}80.1/\underline{74.9}/67.4 & \cellcolor{tab_others}76.9/60.2/\underline{67.2} & \cellcolor{ours!50}\underline{79.4}/76.0/66.3 & \cellcolor{ours}\textbf{80.9}/\textbf{77.9}/\textbf{68.1} \\
    vcpill & 85.1/80.3/72.4 & \underline{87.1}/\underline{84.0}/\underline{74.7} & 59.0/48.7/56.4 & 83.8/81.5/69.9 & 64.1/40.4/56.2 & \cellcolor{ours!50}88.3/87.7/77.4 & \cellcolor{ours}\textbf{89.8}/\textbf{89.6}/\textbf{79.2} \\
    \cellcolor{tab_others}wooden beads & \cellcolor{tab_others}81.2/\underline{78.9}/70.9 & \cellcolor{tab_others}78.4/77.2/67.8 & \cellcolor{tab_others}55.1/52.0/60.2 & \cellcolor{tab_others}\underline{82.4}/78.5/73.0 & \cellcolor{tab_others}62.1/56.4/65.9 & \cellcolor{ours!50}82.5/81.7/\underline{71.8} & \cellcolor{ours}\textbf{84.0}/\textbf{83.7}/\textbf{73.6} \\
    woodstick & 76.9/61.2/58.1 & 80.8/\textbf{72.6}/63.6 & 58.2/35.6/45.2 & \underline{80.4}/\underline{69.2}/60.3 & 74.1/66.0/62.1 & \cellcolor{ours!50}\underline{80.4}/69.0/\underline{63.4} & \cellcolor{ours}\textbf{81.9}/70.9/\textbf{65.2} \\
    \cellcolor{tab_others}zipper & \cellcolor{tab_others}95.3/97.2/91.2 & \cellcolor{tab_others}\underline{98.2}/\underline{98.9}/\underline{95.3} & \cellcolor{tab_others}77.2/86.7/77.6 & \cellcolor{tab_others}96.9/98.1/93.5 & \cellcolor{tab_others}86.0/87.0/84.0 & \cellcolor{ours!50}99.2/99.6/96.9 & \cellcolor{ours}\textbf{100.}/\textbf{100.}/\textbf{98.7} \\
    \midrule
    Mean  & 82.4/79.0/73.9 & \underline{83.0}/\underline{80.9}/\underline{74.3} & 57.2/53.4/61.5 & 82.3/79.2/73.2 & 75.6/66.4/69.9 & \cellcolor{ours!50}86.3/84.6/77.0 & \cellcolor{ours}\textbf{87.8}/\textbf{86.4}/\textbf{78.8} \\
    \bottomrule
    \end{tabular}
    }
  \label{tab:realiadsp}
\end{table*}

\begin{table*}[htp!]
    \centering
    \caption{Comparison with SoTA methods on \textbf{Real-IAD} dataset for multi-class anomaly localization with AU-ROC/AP/F1\_max/AU-PRO metrics.}
    \resizebox{1.\linewidth}{!}{
    \begin{tabular}{cccccccc}
    \toprule
    Method~$\rightarrow$ & RD4AD~\cite{deng2022anomaly} & UniAD~\cite{you2022unified} & SimpleNet~\cite{liu2023simplenet}  & DeSTSeg~\cite{zhang2023destseg}  & DiAD~\cite{he2024diffusion} & \cellcolor{ours!50}MambaAD & \cellcolor{ours}MambaADv2 \\
    \cline{1-1}
    Category~$\downarrow$ & CVPR'22 & NeurlPS'22 & CVPR'23 & CVPR'23 & AAAI'24& \cellcolor{ours!50}NeurIPS'24 & \cellcolor{ours}(Ours) \\
    \hline
    audiojack   & 96.6/12.8/22.1/79.6 &\underline{97.6}/20.0/\underline{31.0}/\underline{83.7} &74.4/ \;0.9/ \;4.8/38.0 &95.5/\textbf{25.4}/\textbf{31.9}/52.6 &91.6/ \;1.0/ \;3.9/63.3 & \cellcolor{ours!50}97.7/\underline{21.6}/29.5/83.9 & \cellcolor{ours}\textbf{97.9}/24.1/\textbf{31.9}/\textbf{84.9} \\
    \cellcolor{tab_others}bottle cap  & \cellcolor{tab_others}\underline{99.5}/18.9/29.9/95.7 &\cellcolor{tab_others}\underline{99.5}/19.4/29.6/\underline{96.0} &\cellcolor{tab_others}85.3/ \;2.3/ \;5.7/45.1 &\cellcolor{tab_others}94.5/\underline{25.3}/\underline{31.1}/25.3 &\cellcolor{tab_others}94.6/ \;4.9/11.4/73.0 & \cellcolor{ours!50}99.7/30.6/34.6/97.2 & \cellcolor{ours}\textbf{99.8}/\textbf{33.1}/\textbf{37.1}/\textbf{98.2} \\
    button battery& 97.6/33.8/37.8/86.5 &96.7/28.5/34.4/77.5 &75.9/ \;3.2/ \;6.6/40.5 &\textbf{98.3}/\textbf{63.9}/\textbf{60.4}/36.9 &84.1/ \;1.4/ \;5.3/66.9 & \cellcolor{ours!50}\underline{98.1}/\underline{46.7}/\underline{49.5}/\underline{86.2} & \cellcolor{ours}\textbf{98.3}/49.1/51.9/\textbf{87.3} \\
    \cellcolor{tab_others}end cap&  \cellcolor{tab_others}\underline{96.7}/\underline{12.5}/\underline{22.5}/\underline{89.2} &\cellcolor{tab_others}95.8/ \;8.8/17.4/85.4 &\cellcolor{tab_others}63.1/ \;0.5/ \;2.8/25.7 &\cellcolor{tab_others}89.6/14.4/\textbf{22.7}/29.5 &\cellcolor{tab_others}81.3/ \;2.0/ \;6.9/38.2 & \cellcolor{ours!50}97.0/12.0/19.6/89.4 & \cellcolor{ours}\textbf{97.2}/\textbf{14.5}/22.0/\textbf{90.4} \\
    eraser& \textbf{99.5}/\underline{30.8}/36.7/\textbf{96.0} &\underline{99.3}/24.4/30.9/\underline{94.1} &80.6/ \;2.7/ \;7.1/42.8 &95.8/\textbf{52.7}/\textbf{53.9}/46.7 &91.1/ \;7.7/15.4/67.5 & \cellcolor{ours!50}99.2/30.2/\underline{38.3}/93.7 & \cellcolor{ours}99.4/32.7/40.7/94.7 \\
    \cellcolor{tab_others}fire hood&  \cellcolor{tab_others}\textbf{98.9}/\textbf{27.7}/\underline{35.2}/\textbf{87.9} &\cellcolor{tab_others}98.6/23.4/32.2/85.3 &\cellcolor{tab_others}70.5/ \;0.3/ \;2.2/25.3 &\cellcolor{tab_others}97.3/\underline{27.1}/\textbf{35.3}/34.7 &\cellcolor{tab_others}91.8/ \;3.2/ \;9.2/66.7 & \cellcolor{ours!50}\underline{98.7}/25.1/31.3/\underline{86.3} & \cellcolor{ours}\textbf{98.9}/27.6/33.7/87.4 \\
    mint&   \underline{95.0}/\underline{11.7}/\underline{23.0}/\underline{72.3} &94.4/ \;7.7/18.1/62.3 &79.9/ \;0.9/ \;3.6/43.3 &84.1/10.3/22.4/ \;9.9 &91.1/ \;5.7/11.6/64.2 & \cellcolor{ours!50}96.5/15.9/27.0/72.6 & \cellcolor{ours}\textbf{96.7}/\textbf{18.4}/\textbf{29.4}/\textbf{73.6} \\
    \cellcolor{tab_others}mounts& \cellcolor{tab_others}\underline{99.3}/\underline{30.6}/\underline{37.1}/\underline{94.9} &\cellcolor{tab_others}\textbf{99.4}/28.0/32.8/\textbf{95.2} &\cellcolor{tab_others}80.5/ \;2.2/ \;6.8/46.1 &\cellcolor{tab_others}94.2/30.0/\textbf{41.3}/43.3 &\cellcolor{tab_others}84.3/ \;0.4/ \;1.1/48.8 & \cellcolor{ours!50}99.2/31.4/35.4/93.5 & \cellcolor{ours}\textbf{99.4}/\textbf{33.9}/37.9/94.5 \\
    pcb &   \underline{97.5}/15.8/24.3/\underline{88.3} &97.0/18.5/28.1/81.6 &78.0/ \;1.4/ \;4.3/41.3 &97.2/\underline{37.1}/\underline{40.4}/48.8 &92.0/ \;3.7/ \;7.4/66.5 & \cellcolor{ours!50}99.2/46.3/50.4/93.1 & \cellcolor{ours}\textbf{99.4}/\textbf{48.7}/\textbf{52.8}/\textbf{94.1} \\
    \cellcolor{tab_others}phone battery&  \cellcolor{tab_others}77.3/22.6/31.7/\underline{94.5} &\cellcolor{tab_others}85.5/11.2/21.6/88.5 &\cellcolor{tab_others}43.4/ \;0.1/ \;0.9/11.8 &\cellcolor{tab_others}79.5/\underline{25.6}/\underline{33.8}/39.5 &\cellcolor{tab_others}\underline{96.8}/ \;5.3/11.4/85.4 & \cellcolor{ours!50}99.4/36.3/41.3/95.3 & \cellcolor{ours}\textbf{99.6}/\textbf{38.8}/\textbf{43.7}/\textbf{96.3} \\
    plastic nut&    \underline{98.8}/21.1/29.6/\underline{91.0} &98.4/20.6/27.1/88.9 &77.4/ \;0.6/ \;3.6/41.5 &96.5/\textbf{44.8}/\textbf{45.7}/38.4 &81.1/ \;0.4/ \;3.4/38.6 & \cellcolor{ours!50}99.4/\underline{33.1}/\underline{37.3}/96.1 & \cellcolor{ours}\textbf{99.6}/35.6/39.7/\textbf{97.1} \\
    \cellcolor{tab_others}plastic plug& \cellcolor{tab_others}99.1/\underline{20.5}/\underline{28.4}/\textbf{94.9} &\cellcolor{tab_others}98.6/17.4/26.1/90.3 &\cellcolor{tab_others}78.6/ \;0.7/ \;1.9/38.8 &\cellcolor{tab_others}91.9/20.1/27.3/21.0 &\cellcolor{tab_others}92.9/ \;8.7/15.0/66.1 & \cellcolor{ours!50}\underline{99.0}/24.2/31.7/\underline{91.5} & \cellcolor{ours}\textbf{99.2}/\textbf{26.7}/\textbf{34.1}/92.5 \\
    porcelain doll& 99.2/24.8/34.6/95.7 &98.7/14.1/24.5/93.2 &81.8/ \;2.0/ \;6.4/47.0 &93.1/\textbf{35.9}/\textbf{40.3}/24.8 &93.1/ \;1.4/ \;4.8/70.4 & \cellcolor{ours!50}99.2/\underline{31.3}/\underline{36.6}/\underline{95.4} & \cellcolor{ours}\textbf{99.4}/33.8/39.1/\textbf{96.4} \\
    \cellcolor{tab_others}regulator&  \cellcolor{tab_others}\textbf{98.0}/ \;7.8/16.1/\textbf{88.6} &\cellcolor{tab_others}95.5/ \;9.1/17.4/76.1 &\cellcolor{tab_others}76.6/ \;0.1/ \;0.6/38.1 &\cellcolor{tab_others}88.8/\underline{18.9}/\underline{23.6}/17.5 &\cellcolor{tab_others}84.2/ \;0.4/ \;1.5/44.4 & \cellcolor{ours!50}\underline{97.6}/20.6/29.8/\underline{87.0} & \cellcolor{ours}97.8/\textbf{23.1}/\textbf{32.2}/88.0 \\
    rolled strip base&  99.7/31.4/39.9/\underline{98.4} &99.6/20.7/32.2/97.8 &80.5/ \;1.7/ \;5.1/52.1 &99.2/\textbf{48.7}/\textbf{50.1}/55.5 &87.7/ \;0.6/ \;3.2/63.4 & \cellcolor{ours!50}99.7/\underline{37.4}/\underline{42.5}/98.8 & \cellcolor{ours}\textbf{99.8}/39.9/44.9/\textbf{99.8} \\
    \cellcolor{tab_others}sim card set&   \cellcolor{tab_others}98.5/40.2/44.2/89.5 &\cellcolor{tab_others}97.9/31.6/39.8/85.0 &\cellcolor{tab_others}71.0/ \;6.8/14.3/30.8 &\cellcolor{tab_others}\textbf{99.1}/\textbf{65.5}/\textbf{62.1}/73.9 &\cellcolor{tab_others}89.9/ \;1.7/ \;5.8/60.4 & \cellcolor{ours!50}\underline{98.8}/\underline{51.1}/\underline{50.6}/\underline{89.4} & \cellcolor{ours}99.0/53.5/53.0/\textbf{90.4} \\
    switch& 94.4/18.9/26.6/\underline{90.9} &\underline{98.1}/33.8/40.6/90.7 &71.7/ \;3.7/ \;9.3/44.2 &97.4/\textbf{57.6}/\textbf{55.6}/44.7 &90.5/ \;1.4/ \;5.3/64.2 & \cellcolor{ours!50}98.2/\underline{39.9}/\underline{45.4}/92.9 & \cellcolor{ours}\textbf{98.4}/42.3/47.8/\textbf{93.9} \\
    \cellcolor{tab_others}tape& \cellcolor{tab_others}\underline{99.7}/42.4/47.8/98.4 &\cellcolor{tab_others}\underline{99.7}/29.2/36.9/97.5 &\cellcolor{tab_others}77.5/ \;1.2/ \;3.9/41.4 &\cellcolor{tab_others}99.0/\textbf{61.7}/\textbf{57.6}/48.2 &\cellcolor{tab_others}81.7/ \;0.4/ \;2.7/47.3 & \cellcolor{ours!50}99.8/\underline{47.1}/\underline{48.2}/\underline{98.0} & \cellcolor{ours}\textbf{99.9}/49.5/50.6/\textbf{99.0} \\
    terminalblock&  \underline{99.5}/27.4/35.8/\underline{97.6} &99.2/23.1/30.5/94.4 &87.0/ \;0.8/ \;3.6/54.8 &96.6/\textbf{40.6}/\textbf{44.1}/34.8 &75.5/ \;0.1/ \;1.1/38.5 & \cellcolor{ours!50}99.8/\underline{35.3}/\underline{39.7}/98.2 & \cellcolor{ours}\textbf{99.9}/37.8/42.1/\textbf{99.2} \\
    \cellcolor{tab_others}toothbrush& \cellcolor{tab_others}\underline{96.9}/26.1/34.2/\underline{88.7} &\cellcolor{tab_others}95.7/16.4/25.3/84.3 &\cellcolor{tab_others}84.7/ \;7.2/14.8/52.6 &\cellcolor{tab_others}94.3/30.0/37.3/42.8&\cellcolor{tab_others}82.0/ \;1.9/ \;6.6/54.5& \cellcolor{ours!50}97.5/\underline{27.8}/\underline{36.7}/91.4 & \cellcolor{ours}\textbf{97.7}/\textbf{30.3}/\textbf{39.1}/\textbf{92.4} \\
    toy & \underline{95.2}/ \;5.1/12.8/\underline{82.3} & 93.4/ \;4.6/12.4/70.5 &67.7/ \;0.1/ \;0.4/25.0 &86.3/\; \underline{8.1}/\underline{15.9}/16.4 &82.1/ \;1.1/ \;4.2/50.3 & \cellcolor{ours!50}96.0/16.4/25.8/86.3 & \cellcolor{ours}\textbf{96.2}/\textbf{18.9}/\textbf{28.2}/\textbf{87.3} \\
    \cellcolor{tab_others}toy brick&  \cellcolor{tab_others}96.4/16.0/24.6/\underline{75.3} &\cellcolor{tab_others}\textbf{97.4}/17.1/\underline{27.6}/\textbf{81.3} &\cellcolor{tab_others}86.5/ \;5.2/11.1/56.3 &\cellcolor{tab_others}94.7/\textbf{24.6}/\textbf{30.8}/45.5 &\cellcolor{tab_others}93.5/ \;3.1/ \;8.1/66.4 & \cellcolor{ours!50}\underline{96.6}/\underline{18.0}/25.8/74.7 & \cellcolor{ours}96.8/20.5/28.2/75.8 \\
    transistor1&    \underline{99.1}/29.6/35.5/\underline{95.1} &98.9/25.6/33.2/94.3 &71.7/ \;5.1/11.3/35.3 &97.3/\textbf{43.8}/\textbf{44.5}/45.4 &88.6 \;7.2/15.3/58.1 & \cellcolor{ours!50}99.4/\underline{39.4}/\underline{40.0}/96.5 & \cellcolor{ours}\textbf{99.6}/41.8/42.4/\textbf{97.5} \\
    \cellcolor{tab_others}u block&    \cellcolor{tab_others}99.6/\underline{40.5}/45.2/\textbf{96.9} &\cellcolor{tab_others}99.3/22.3/29.6/94.3 &\cellcolor{tab_others}76.2/ \;4.8/12.2/34.0 &\cellcolor{tab_others}96.9/\textbf{57.1}/\textbf{55.7}/38.5 &\cellcolor{tab_others}88.8/ \;1.6/ \;5.4/54.2 & \cellcolor{ours!50}\underline{99.5}/37.8/\underline{46.1}/\underline{95.4} & \cellcolor{ours}\textbf{99.7}/40.3/48.5/96.4 \\
    usb&    98.1/26.4/35.2/\underline{91.0} &97.9/20.6/31.7/85.3 &81.1/ \;1.5/ \;4.9/52.4 &\underline{98.4}/\textbf{42.2}/\textbf{47.7}/57.1 &78.0/ \;1.0/ \;3.1/28.0 & \cellcolor{ours!50}99.2/\underline{39.1}/\underline{44.4}/95.2 & \cellcolor{ours}\textbf{99.4}/41.5/46.9/\textbf{96.2} \\
    \cellcolor{tab_others}usb adaptor&  \cellcolor{tab_others}94.5/ \;9.8/17.9/73.1 &\cellcolor{tab_others}\underline{96.6}/10.5/19.0/\underline{78.4} &\cellcolor{tab_others}67.9/ \;0.2/ \;1.3/28.9 &\cellcolor{tab_others}94.9/\textbf{25.5}/\textbf{34.9}/36.4 &\cellcolor{tab_others}94.0/ \;2.3/ \;6.6/75.5 & \cellcolor{ours!50}97.3/\underline{15.3}/\underline{22.6}/82.5 & \cellcolor{ours}\textbf{97.5}/17.8/25.0/\textbf{83.5} \\
    vcpill& 98.3/43.1/48.6/88.7 &\textbf{99.1}/40.7/43.0/\textbf{91.3} &68.2/ \;1.1/ \;3.3/22.0 &97.1/\textbf{64.7}/\textbf{62.3}/42.3 &90.2/ \;1.3/ \;5.2/60.8 & \cellcolor{ours!50}\underline{98.7}/\underline{50.2}/\underline{54.5}/\underline{89.3} & \cellcolor{ours}98.9/52.6/56.9/90.3 \\
    \cellcolor{tab_others}wooden beads&   \cellcolor{tab_others}98.0/27.1/34.7/\textbf{85.7} &\cellcolor{tab_others}97.6/16.5/23.6/\underline{84.6} &\cellcolor{tab_others}68.1/ \;2.4/ \;6.0/28.3 &\cellcolor{tab_others}94.7/\textbf{38.9}/\textbf{42.9}/39.4 &\cellcolor{tab_others}85.0/ \;1.1/ \;4.7/45.6 & \cellcolor{ours!50}98.0/\underline{32.6}/\underline{39.8}/84.5 & \cellcolor{ours}\textbf{98.2}/35.1/42.2/85.5 \\
    woodstick&  \underline{97.8}/30.7/38.4/\textbf{85.0} &94.0/36.2/44.3/77.2 &76.1/ \;1.4/ \;6.0/32.0 &\textbf{97.9}/\textbf{60.3}/\textbf{60.0}/51.0 &90.9/ \;2.6/ \;8.0/60.7 & \cellcolor{ours!50}97.7/\underline{40.1}/\underline{44.9}/\underline{82.7} & \cellcolor{ours}\textbf{97.9}/42.5/47.3/83.8 \\
    \cellcolor{tab_others}zipper& \cellcolor{tab_others}\underline{99.1}/\underline{44.7}/\underline{50.2}/\underline{96.3} &\cellcolor{tab_others}98.4/32.5/36.1/95.1 &\cellcolor{tab_others}89.9/23.3/31.2/55.5 &\cellcolor{tab_others}98.2/35.3/39.0/78.5 &\cellcolor{tab_others}90.2/12.5/18.8/53.5 & \cellcolor{ours!50}99.3/58.2/61.3/97.6 & \cellcolor{ours}\textbf{99.5}/\textbf{60.6}/\textbf{63.7}/\textbf{98.6} \\
    \hline
    mean&   \underline{97.3}/25.0/32.7/\underline{89.6} &\underline{97.3}/21.1/29.2/86.7 &75.7/ \;2.8/ \;6.5/39.0 &94.6/\textbf{37.9}/\textbf{41.7}/40.6 &88.0/ \;2.9/ \;7.1/58.1 & \cellcolor{ours!50}98.5/\underline{33.0}/\underline{38.7}/90.5 & \cellcolor{ours}\textbf{98.7}/35.5/41.1/\textbf{91.5} \\

    \bottomrule
    \end{tabular}
    }
  \label{tab:realiadpx}
\end{table*}

\section*{More Quantitative Results for Each Category on The MVTec-AD Dataset for Single-class Anomaly Detection.}
Tab. \ref{tab:sgmvtecsp} and Tab. \ref{tab:sgmvtecpx} respectively present the results of image-level anomaly detection and pixel-level anomaly localization quantitative outcomes for single-class anomaly detection of the MVTec-AD dataset.

\begin{table*}[t!]
  \centering
  \caption{Comparison with SoTA methods on\textbf{ MVTec-AD} dataset for \textbf{single-class} anomaly detection with AU-ROC/AP/F1\_max metrics.}
  \resizebox{1\linewidth}{!}{
    \begin{tabular}{p{3em}<{\centering} p{3.25em}<{\centering}p{6.2em}<{\centering} p{6.2em}<{\centering} p{6.2em}<{\centering} p{6.2em}<{\centering} p{6.2em}<{\centering} p{6.2em}<{\centering} p{6.2em}<{\centering} p{7em}<{\centering} p{7em}<{\centering}}
    \toprule
    \multicolumn{2}{c}{Method~$\rightarrow$} & RD4AD~\cite{deng2022anomaly} & UniAD~\cite{you2022unified} & SimpleNet~\cite{liu2023simplenet}  & DeSTSeg~\cite{zhang2023destseg}  & PatchCore~\cite{roth2022towards} & \cellcolor{ours!50}MambaAD & \cellcolor{ours}MambaADv2 \\
    \cline{1-2}
    \multicolumn{2}{c}{Category~$\downarrow$} & CVPR'22 & NeurlPS'22 & CVPR'23 & CVPR'23 & CVPR'22& \cellcolor{ours!50}NeurIPS'24 & \cellcolor{ours}(Ours) \\
    \hline
    \multicolumn{1}{c}{\multirow{10}[1]{*}{\begin{turn}{-90}Objects\end{turn}}} & \multicolumn{1}{c}{Bottle} & \textbf{100.}/\textbf{100.}/\textbf{100.}&\textbf{100.}/\textbf{100.}/\textbf{100.}&99.9/\textbf{100.}/99.2&\textbf{100.}/\textbf{100.}/\textbf{100.}&\textbf{100.}/\textbf{100.}/\textbf{100.}& \cellcolor{ours!50}\textbf{100.}/\textbf{100.}/\textbf{100.} & \cellcolor{ours}\textbf{100.}/\textbf{100.}/\textbf{100.} \\
    \multicolumn{1}{c}{} & \multicolumn{1}{c}{\cellcolor{tab_others}Cable} & \cellcolor{tab_others}94.8/97.0/89.7&\cellcolor{tab_others}97.9/98.7/93.8&\cellcolor{tab_others}99.6/99.7/97.3&\cellcolor{tab_others}97.1/98.4/93.3&\cellcolor{tab_others}\textbf{99.8}/\textbf{99.9}/\textbf{98.9}& \cellcolor{ours!50}98.5/99.1/95.1 & \cellcolor{ours}99.4/99.5/96.1 \\
    \multicolumn{1}{c}{} & \multicolumn{1}{c}{Capsule}  & 98.4/99.7/97.3&87.1/95.3/95.2&97.2/99.2/\textbf{98.6}&97.4/99.5/96.8&97.4/99.4/98.2& \cellcolor{ours!50}97.8/99.5/97.2 & \cellcolor{ours}\textbf{98.8}/\textbf{99.9}/98.2 \\
    \multicolumn{1}{c}{} & \multicolumn{1}{c}{\cellcolor{tab_others}Hazelnut} & \cellcolor{tab_others}\textbf{100.}/\textbf{100.}/\textbf{100.}&\cellcolor{tab_others}99.7/99.8/98.6&\cellcolor{tab_others}99.0/99.4/97.2&\cellcolor{tab_others}99.7/99.8/99.3&\cellcolor{tab_others}\textbf{100.}/\textbf{100.}/\textbf{100.}& \cellcolor{ours!50}\textbf{100.}/\textbf{100.}/\textbf{100.} & \cellcolor{ours}\textbf{100.}/\textbf{100.}/\textbf{100.} \\
    \multicolumn{1}{c}{} & \multicolumn{1}{c}{Metal Nut}  & 99.9/\textbf{100.}/99.5&98.7/99.7/98.4&99.9/\textbf{100.}/99.5&99.7/99.9/98.9&99.9/\textbf{100.}/99.5& \cellcolor{ours!50}\textbf{100.}/\textbf{100.}/99.5 & \cellcolor{ours}\textbf{100.}/\textbf{100.}/\textbf{100.} \\
    \multicolumn{1}{c}{} & \multicolumn{1}{c}{\cellcolor{tab_others}Pill}  & \cellcolor{tab_others}97.5/99.6/97.1&\cellcolor{tab_others}96.0/99.3/95.8&\cellcolor{tab_others}\textbf{98.6}/\textbf{99.7}/\textbf{97.9}&\cellcolor{tab_others}96.2/99.3/95.7&\cellcolor{tab_others}96.6/99.4/96.5& \cellcolor{ours!50}95.3/99.1/95.3 & \cellcolor{ours}96.3/99.5/96.3 \\
    \multicolumn{1}{c}{} & \multicolumn{1}{c}{Screw} & 97.7/99.2/95.9&87.7/93.0/93.1&97.5/99.2/95.1&92.3/97.5/90.9&98.1/99.3/97.5& \cellcolor{ours!50}97.6/99.1/97.1 & \cellcolor{ours}\textbf{98.5}/\textbf{99.6}/\textbf{98.1} \\
    \multicolumn{1}{c}{} & \multicolumn{1}{c}{\cellcolor{tab_others}Toothbrush} & \cellcolor{tab_others}93.3/97.2/95.2&\cellcolor{tab_others}88.1/94.8/93.8&\cellcolor{tab_others}99.4/99.8/98.4&\cellcolor{tab_others}99.7/99.9/98.4&\cellcolor{tab_others}\textbf{100.}/\textbf{100.}/\textbf{100.}& \cellcolor{ours!50}96.1/98.4/95.2 & \cellcolor{ours}97.1/98.9/96.2 \\
    \multicolumn{1}{c}{} & \multicolumn{1}{c}{Transistor} & 98.4/97.7/93.8&\textbf{100.}/\textbf{100.}/\textbf{100.}&99.9/99.9/98.7&98.5/98.2/94.7&\textbf{100.}/\textbf{100.}/\textbf{100.}& \cellcolor{ours!50}\textbf{100.}/\textbf{100.}/\textbf{100.} & \cellcolor{ours}\textbf{100.}/\textbf{100.}/\textbf{100.} \\
    \multicolumn{1}{c}{} & \multicolumn{1}{c}{\cellcolor{tab_others}Zipper}  & \cellcolor{tab_others}98.3/99.5/97.9&\cellcolor{tab_others}90.6/96.5/93.9&\cellcolor{tab_others}99.8/99.9/99.6&\cellcolor{tab_others}\textbf{100.}/\textbf{100.}/\textbf{100.}&\cellcolor{tab_others}99.6/99.9/98.7& \cellcolor{ours!50}98.6/99.6/97.5 & \cellcolor{ours}99.5/\textbf{100.}/98.4 \\
    \hline
    \multicolumn{1}{c}{\multirow{5}[1]{*}{\begin{turn}{-90}Textures\end{turn}}} & \multicolumn{1}{c}{Carpet}  & 99.7/99.9/98.3&99.4/99.8/98.9&99.1/99.7/98.3&99.3/99.8/97.7&98.5/99.6/97.1& \cellcolor{ours!50}\textbf{100.}/\textbf{100.}/\textbf{100.} & \cellcolor{ours}\textbf{100.}/\textbf{100.}/\textbf{100.} \\
    \multicolumn{1}{c}{} & \multicolumn{1}{c}{\cellcolor{tab_others}Grid} & \cellcolor{tab_others}\textbf{100.}/\textbf{100.}/\textbf{100.}&\cellcolor{tab_others}97.6/99.3/96.6&\cellcolor{tab_others}\textbf{100.}/\textbf{100.}/\textbf{100.}&\cellcolor{tab_others}\textbf{100.}/\textbf{100.}/\textbf{100.}&\cellcolor{tab_others}99.7/99.9/99.1& \cellcolor{ours!50}\textbf{100.}/\textbf{100.}/\textbf{100.} & \cellcolor{ours}\textbf{100.}/\textbf{100.}/\textbf{100.} \\
    \multicolumn{1}{c}{} & \multicolumn{1}{c}{Leather} & \textbf{100.}/\textbf{100.}/\textbf{100.}&\textbf{100.}/\textbf{100.}/\textbf{100.}&\textbf{100.}/\textbf{100.}/\textbf{100.}&\textbf{100.}/\textbf{100.}/\textbf{100.}&\textbf{100.}/\textbf{100.}/\textbf{100.}& \cellcolor{ours!50}\textbf{100.}/\textbf{100.}/\textbf{100.} & \cellcolor{ours}\textbf{100.}/\textbf{100.}/\textbf{100.} \\
    \multicolumn{1}{c}{} & \multicolumn{1}{c}{\cellcolor{tab_others}Tile}  & \cellcolor{tab_others}\textbf{100.}/\textbf{100.}/\textbf{100.}&\cellcolor{tab_others}95.8/98.2/95.8&\cellcolor{tab_others}99.9/99.9/98.8&\cellcolor{tab_others}\textbf{100.}/\textbf{100.}/99.4&\cellcolor{tab_others}99.0/99.7/98.8& \cellcolor{ours!50}97.5/99.1/94.7 & \cellcolor{ours}98.4/99.6/95.7 \\
    \multicolumn{1}{c}{} & \multicolumn{1}{c}{Wood}  & 99.5/99.8/98.3&97.9/99.4/95.9&99.9/\textbf{100.}/99.2&99.3/99.8/98.4&99.1/99.7/97.5& \cellcolor{ours!50}99.6/99.9/99.2 & \cellcolor{ours}\textbf{100.}/\textbf{100.}/\textbf{100.} \\
    \hline
    \multicolumn{2}{c}{\cellcolor{tab_others}Mean} & \cellcolor{tab_others}98.5/99.3/97.5&\cellcolor{tab_others}95.8/98.9/96.7&\cellcolor{tab_others}\textbf{99.3}/\textbf{99.8}/98.5&\cellcolor{tab_others}98.6/99.5/97.6&\cellcolor{tab_others}99.2/\textbf{99.8}/\textbf{98.8}& \cellcolor{ours!50}98.7/99.6/98.1 & \cellcolor{ours}99.2/\textbf{99.8}/98.6 \\
    \bottomrule
    \end{tabular}%
  }
  \label{tab:sgmvtecsp}%
\end{table*}

\begin{table*}[t!]
  \centering
  \caption{Comparison with SoTA methods on \textbf{MVTec-AD} dataset for \textbf{single-class }anomaly localization with AU-ROC/AP/F1\_max/AU-PRO metrics.}
  \resizebox{1\linewidth}{!}{
    \begin{tabular}{ccccccccc}
    \toprule
    \multicolumn{2}{c}{Method~$\rightarrow$} & RD4AD~\cite{deng2022anomaly} & UniAD~\cite{you2022unified} & SimpleNet~\cite{liu2023simplenet}  & DeSTSeg~\cite{zhang2023destseg}  & PatchCore~\cite{roth2022towards} & \cellcolor{ours!50}MambaAD & \cellcolor{ours}MambaADv2 \\
    \cline{1-2}
    \multicolumn{2}{c}{Category~$\downarrow$} & CVPR'22 & NeurlPS'22 & CVPR'23 & CVPR'23 & CVPR'22& \cellcolor{ours!50}NeurIPS'24 & \cellcolor{ours}(Ours) \\
    \hline
    \multicolumn{1}{c}{\multirow{10}[1]{*}{\begin{turn}{-90}Objects\end{turn}}} & \multicolumn{1}{c}{Bottle} &  98.6/75.9/74.1/95.8&98.3/73.6/70.7/94.4&98.0/70.4/72.7/88.8&\textbf{99.3}/\textbf{91.0}/\textbf{84.3}/94.0&98.5/77.7/75.2/94.9& \cellcolor{ours!50}98.8/79.7/76.6/96.1 & \cellcolor{ours}99.0/82.3/79.2/\textbf{97.2} \\
    \multicolumn{1}{c}{} & \multicolumn{1}{c}{\cellcolor{tab_others}Cable} & \cellcolor{tab_others}96.8/51.3/57.3/88.8&\cellcolor{tab_others}97.4/54.8/56.9/87.7&\cellcolor{tab_others}97.4/\textbf{66.8}/60.0/87.7&\cellcolor{tab_others}95.8/55.0/52.8/82.4&\cellcolor{tab_others}\textbf{98.4}/66.3/\textbf{65.1}/92.5& \cellcolor{ours!50}98.0/54.4/61.7/93.4 & \cellcolor{ours}98.2/57.0/64.3/\textbf{94.5} \\
    \multicolumn{1}{c}{} & \multicolumn{1}{c}{Capsule} & 98.9/48.1/49.9/\textbf{95.3}&98.0/35.2/40.1/86&98.9/42.5/49.1/92.8&98.8/\textbf{52.4}/\textbf{57.9}/71.6&\textbf{99.0}/44.7/50.9/91.7& \cellcolor{ours!50}98.6/45.0/47.5/94.3 & \cellcolor{ours}98.8/47.7/50.1/\textbf{95.3} \\
    \multicolumn{1}{c}{} & \multicolumn{1}{c}{\cellcolor{tab_others}Hazelnut} & \cellcolor{tab_others}98.8/59.6/60.8/96.5&\cellcolor{tab_others}98.3/55.0/56.1/92.8&\cellcolor{tab_others}97.9/46.2/50.3/78.9&\cellcolor{tab_others}\textbf{99.4}/\textbf{84.5}/\textbf{79.0}/90.6&\cellcolor{tab_others}98.7/53.5/59.2/89.7& \cellcolor{ours!50}99.0/61.8/63.9/95.8 & \cellcolor{ours}99.2/64.4/66.5/\textbf{96.8} \\
    \multicolumn{1}{c}{} & \multicolumn{1}{c}{Metal Nut} & 96.9/75.3/79.5/94.1&94.4/55.8/68.8/77.8&98.8/91.6/86.6/84.0&\textbf{99.2}/\textbf{95.0}/\textbf{88.7}/\textbf{95.0}&98.3/86.9/85.1/91.4& \cellcolor{ours!50}97.1/78.3/80.6/93.6 & \cellcolor{ours}97.3/80.9/83.2/94.7 \\
    \multicolumn{1}{c}{} & \multicolumn{1}{c}{\cellcolor{tab_others}Pill} & \cellcolor{tab_others}97.6/66.6/66.7/95.9&\cellcolor{tab_others}95.2/45.9/51.9/92.2&\cellcolor{tab_others}98.5/79.5/72.6/92.5&\cellcolor{tab_others}\textbf{98.9}/\textbf{87.9}/\textbf{79.6}/56.6&\cellcolor{tab_others}97.8/77.8/72.9/94.1& \cellcolor{ours!50}97.3/64.5/66.0/95.8 & \cellcolor{ours}97.5/67.2/68.7/\textbf{96.8} \\
    \multicolumn{1}{c}{} & \multicolumn{1}{c}{Screw} & 99.5/44.6/47.2/96.9&98.4/15.5/23.6/91.3&99.3/35.0/38.8/95.2&97.3/54.2/53.3/53.6&99.5/36.5/40.1/96.6& \cellcolor{ours!50}99.5/52.2/51.7/97.6 & \cellcolor{ours}\textbf{99.7}/\textbf{54.9}/\textbf{54.3}/\textbf{98.7} \\
    \multicolumn{1}{c}{} & \multicolumn{1}{c}{\cellcolor{tab_others}Toothbrush} & \cellcolor{tab_others}99.0/51.8/60.9/92.0&\cellcolor{tab_others}98.5/42.6/52.1/84.6&\cellcolor{tab_others}98.5/41.7/52.9/92.4&\cellcolor{tab_others}\textbf{99.5}/\textbf{78.2}/\textbf{73.7}/90.8&\cellcolor{tab_others}98.6/38.3/52.3/92.3& \cellcolor{ours!50}99.0/51.1/60.7/92.3 & \cellcolor{ours}99.2/53.8/63.3/\textbf{93.3} \\
    \multicolumn{1}{c}{} & \multicolumn{1}{c}{Transistor} & 90.6/52.8/54.9/80.8&\textbf{98.8}/\textbf{79.5}/\textbf{75.8}/\textbf{94.5}&96.8/67.4/62.1/91.1&88.1/63.6/60.0/80.9&96.2/66.4/61.5/89.8& \cellcolor{ours!50}96.8/69.7/67.5/91.6 & \cellcolor{ours}97.0/72.3/70.2/92.7 \\
    \multicolumn{1}{c}{} & \multicolumn{1}{c}{\cellcolor{tab_others}Zipper}  & \cellcolor{tab_others}98.7/53.0/59.0/94.6&\cellcolor{tab_others}96.5/35.5/43.4/89.3&\cellcolor{tab_others}98.9/64.3/65.4/95.5&\cellcolor{tab_others}\textbf{99.2}/\textbf{85.7}/\textbf{77.4}/84.3&\cellcolor{tab_others}98.9/62.9/66.1/94.3& \cellcolor{ours!50}98.1/55.6/58.8/94.8 & \cellcolor{ours}98.3/58.3/61.4/\textbf{95.8} \\
    \hline
    \multicolumn{1}{c}{\multirow{5}[1]{*}{\begin{turn}{-90}Textures\end{turn}}} & \multicolumn{1}{c}{Carpet} & 99.1/60.1/60.9/95.6&98.5/52.1/53.3/94.9&98.0/43.2/47.8/87.9&98.2/\textbf{77.5}/\textbf{72.5}/94.9&99.1/64.1/63.0/94.8& \cellcolor{ours!50}99.2/63.0/63.4/97.0 & \cellcolor{ours}\textbf{99.4}/65.7/66.0/\textbf{98.1} \\
    \multicolumn{1}{c}{} & \multicolumn{1}{c}{\cellcolor{tab_others}Grid}  & \cellcolor{tab_others}99.3/45.2/47.7/96.3&\cellcolor{tab_others}94.3/24.3/31.5/85.8&\cellcolor{tab_others}98.8/34.7/39.3/93.9&\cellcolor{tab_others}\textbf{99.5}/\textbf{62.5}/\textbf{63.5}/93.8&\cellcolor{tab_others}98.8/31.0/35.3/94.2& \cellcolor{ours!50}99.2/48.9/48.8/97.3 & \cellcolor{ours}99.4/51.5/51.4/\textbf{98.3} \\
    \multicolumn{1}{c}{} & \multicolumn{1}{c}{Leather} & 99.3/44.7/46.7/97.9&99.1/38.5/41.9/98.1&99.2/41.7/45.6/95.7&\textbf{99.8}/\textbf{74.9}/\textbf{69.6}/96.9&99.3/46.3/46.7/96.8& \cellcolor{ours!50}99.3/47.5/49.5/98.6 & \cellcolor{ours}99.5/50.1/52.1/\textbf{99.7} \\
    \multicolumn{1}{c}{} & \multicolumn{1}{c}{\cellcolor{tab_others}Tile}  & \cellcolor{tab_others}95.1/48.6/59.7/85.1&\cellcolor{tab_others}90.2/40.4/48.7/77.1&\cellcolor{tab_others}97.2/69.8/68.8/89.7&\cellcolor{tab_others}\textbf{99.4}/\textbf{94.6}/\textbf{87.1}/89.1&\cellcolor{tab_others}95.8/55.0/64.7/\textbf{90.0}& \cellcolor{ours!50}93.1/42.8/52.3/80.3 & \cellcolor{ours}93.3/45.5/54.9/81.4 \\
    \multicolumn{1}{c}{} & \multicolumn{1}{c}{Wood} & 94.8/48.6/49.5/91.6&93.5/38.6/43.9/85.2&94.0/45.6/48.1/82.5&\textbf{98.4}/\textbf{84.5}/\textbf{77.5}/90.9&95.0/50.2/50.9/85.3& \cellcolor{ours!50}93.9/44.2/47.7/92.1 & \cellcolor{ours}94.2/46.9/50.4/\textbf{93.2} \\
    \hline
    \multicolumn{2}{c}{\cellcolor{tab_others}Mean} & \cellcolor{tab_others}97.5/55.1/58.3/93.1&\cellcolor{tab_others}96.6/45.8/50.6/88.8&\cellcolor{tab_others}98.0/56.0/57.3/89.9&\cellcolor{tab_others}\textbf{98.1}/\textbf{76.1}/\textbf{71.8}/84.4&\cellcolor{tab_others}\textbf{98.1}/57.2/59.3/92.6& \cellcolor{ours!50}97.8/57.2/59.8/94.0 & \cellcolor{ours}98.0/59.9/62.4/\textbf{95.1} \\
    \bottomrule
    \end{tabular}
  }
  \label{tab:sgmvtecpx}%
\end{table*}

\section*{More Quantitative Results for Each Category on The VisA Dataset for Single-class Anomaly Detection.}
Tab. \ref{tab:sgvisasp} and Tab. \ref{tab:sgvisapx} respectively present the results of image-level anomaly detection and pixel-level anomaly localization quantitative outcomes for single-class anomaly detection of the VisA dataset. 

\begin{table*}[t!]
  \centering
   \caption{Comparison with SoTA methods on \textbf{VisA} dataset for \textbf{single-class} anomaly detection with AU-ROC/AP/F1\_max metrics.}
  \resizebox{1.\linewidth}{!}{
    \begin{tabular}{cccccccc}
    \toprule
    Method~$\rightarrow$ & RD4AD~\cite{deng2022anomaly} & UniAD~\cite{you2022unified} & SimpleNet~\cite{liu2023simplenet}  & DeSTSeg~\cite{zhang2023destseg}  & PatchCore~\cite{roth2022towards} & \cellcolor{ours!50}MambaAD & \cellcolor{ours}MambaADv2 \\
    \cline{1-1}
    Category~$\downarrow$ & CVPR'22 & NeurlPS'22 & CVPR'23 & CVPR'23 & CVPR'22& \cellcolor{ours!50}NeurIPS'24 & \cellcolor{ours}(Ours) \\
    \hline
    pcb1  & 95.9/95.6/92.8&91.9/90.9/87.2&\textbf{98.9}/\textbf{98.9}/\textbf{96.0}&94.9/92.5/92.0&97.7/97.1/94.9& \cellcolor{ours!50}96.0/95.1/93.8 & \cellcolor{ours}97.0/96.2/95.4 \\
    \cellcolor{tab_others}pcb2  & \cellcolor{tab_others}96.5/96.2/92.1&\cellcolor{tab_others}92.2/92.3/84.8&\cellcolor{tab_others}\textbf{98.8}/\textbf{98.8}/\textbf{93.9}&\cellcolor{tab_others}96.6/94.7/93.4&\cellcolor{tab_others}97.7/98.2/93.3& \cellcolor{ours!50}97.1/96.8/91.9 & \cellcolor{ours}98.1/97.9/93.5 \\
    pcb3  & 96.2/96.1/91.1&90.9/91.3/86.2&\textbf{98.9}/98.8/\textbf{95.1}&97.0/97.3/91.7&98.5/98.5/93.5& \cellcolor{ours!50}97.8/98.0/92.7 & \cellcolor{ours}98.8/\textbf{99.1}/94.4 \\
    \cellcolor{tab_others}pcb4  & \cellcolor{tab_others}\textbf{100.}/\textbf{100.}/\textbf{99.5}&\cellcolor{tab_others}98.8/98.6/96.1&\cellcolor{tab_others}99.3/99.0/97.6&\cellcolor{tab_others}99.3/99.3/94.8&\cellcolor{tab_others}99.8/99.8/97.5& \cellcolor{ours!50}99.7/99.7/97.5 & \cellcolor{ours}\textbf{100.}/\textbf{100.}/99.1 \\
    \hline
    macaroni1 & 92.1/90.2/85.4&87.5/83.3/79.6&\textbf{94.8}/\textbf{92.7}/\textbf{89.0}&82.8/75.1/77.2&91.3/85.9/87.0& \cellcolor{ours!50}91.2/88.4/84.1 & \cellcolor{ours}92.3/89.6/85.7 \\
    \cellcolor{tab_others}macaroni2 & \cellcolor{tab_others}\textbf{86.1}/80.3/\textbf{80.2}&\cellcolor{tab_others}83.1/\textbf{82.5}/76.4&\cellcolor{tab_others}83.4/77.0/70.4&\cellcolor{tab_others}72.8/66.1/69.2&\cellcolor{tab_others}75.2/60.6/59.4& \cellcolor{ours!50}81.8/78.9/77.7 & \cellcolor{ours}82.9/80.1/79.4 \\
    capsules & 89.7/94.2/87.7&79.1/88.7/79.0&90.1/93.3/86.4&85.3/91.5/83.6&77.7/83.2/74.4& \cellcolor{ours!50}91.4/94.6/90.5 & \cellcolor{ours}\textbf{92.4}/\textbf{95.7}/\textbf{92.1} \\
    \cellcolor{tab_others}candle & \cellcolor{tab_others}94.6/94.7/89.4&\cellcolor{tab_others}96.5/96.5/89.3&\cellcolor{tab_others}92.3/85.6/86.9&\cellcolor{tab_others}94.6/95.1/87.2&\cellcolor{tab_others}92.2/85.3/84.4& \cellcolor{ours!50}97.7/97.7/92.1 & \cellcolor{ours}\textbf{98.7}/\textbf{98.9}/\textbf{93.7} \\
    \hline
    cashew & \textbf{96.9}/98.6/93.4&91.9/96.1/90.3&94.1/96.2/90.9&73.8/87.2/81.3&91.2/93.5/89.2& \cellcolor{ours!50}95.7/97.6/93.7 & \cellcolor{ours}96.7/\textbf{98.7}/\textbf{95.3} \\
    \cellcolor{tab_others}chewinggum &\cellcolor{tab_others}98.8/99.4/96.3&\cellcolor{tab_others}97.4/98.8/94.8&\cellcolor{tab_others}98.6/99.3/96.9&\cellcolor{tab_others}97.1/98.8/95.9&\cellcolor{tab_others}97.9/98.8/95.4& \cellcolor{ours!50}99.0/99.5/96.1 & \cellcolor{ours}\textbf{100.}/\textbf{100.}/\textbf{97.7} \\
    fryum &94.8/97.7/91.7&85.5/92.6/86.0&96.4/98.0/93.5&91.7/96.0/88.6&95.0/96.7/89.8& \cellcolor{ours!50}96.5/98.3/92.5 & \cellcolor{ours}\textbf{97.5}/\textbf{99.4}/\textbf{94.1} \\
    \cellcolor{tab_others}pipe\_fryum &\cellcolor{tab_others}99.7/99.8/98.5&\cellcolor{tab_others}91.9/95.4/90.8&\cellcolor{tab_others}99.8/99.9/99.0&\cellcolor{tab_others}99.3/99.6/97.5&\cellcolor{tab_others}99.1/99.5/97.5& \cellcolor{ours!50}99.2/99.6/97.6 & \cellcolor{ours}\textbf{100.}/\textbf{100.}/\textbf{99.2} \\
    \hline
    Mean  & 95.1/95.2/91.5&90.6/92.3/86.7&95.4/94.8/91.3&90.4/91.1/87.7&92.8/91.4/88.0& \cellcolor{ours!50}95.3/95.3/91.7 & \cellcolor{ours}\textbf{96.2}/\textbf{96.3}/\textbf{93.3} \\
    \bottomrule
    \end{tabular}
  }
  \label{tab:sgvisasp}%
\end{table*}

\begin{table*}[t!]
  \centering
  \caption{Comparison with SoTA methods on \textbf{VisA} dataset for \textbf{single-class} anomaly localization with AU-ROC/AP/F1\_max/AU-PRO metrics.}
  \resizebox{1\linewidth}{!}{
    \begin{tabular}{cccccccc}
    \toprule
    Method~$\rightarrow$ & RD4AD~\cite{deng2022anomaly} & UniAD~\cite{you2022unified} & SimpleNet~\cite{liu2023simplenet}  & DeSTSeg~\cite{zhang2023destseg}  & PatchCore~\cite{roth2022towards} & \cellcolor{ours!50}MambaAD & \cellcolor{ours}MambaADv2 \\
    \cline{1-1}
    Category~$\downarrow$ & CVPR'22 & NeurlPS'22 & CVPR'23 & CVPR'23 & CVPR'22& \cellcolor{ours!50}NeurIPS'24 & \cellcolor{ours}(Ours) \\
    \midrule
    pcb1  & 99.7/71.2/68.4/94.8&99.2/53.0/56.2/86.7&99.6/87.7/78.4/89.6&99.4/61.5/62.8/76.9&99.8/\textbf{91.8}/\textbf{85.9}/94.7& \cellcolor{ours!50}99.8/72.5/67.4/94.9 & \cellcolor{ours}\textbf{100.}/74.8/69.9/\textbf{95.7} \\
    \cellcolor{tab_others}pcb2  & \cellcolor{tab_others}98.8/16.2/24.0/90.1&\cellcolor{tab_others}97.1/ \;5.8/13.0/80.8&\cellcolor{tab_others}98.2/11.9/22.2/87.7&\cellcolor{tab_others}98.9/\textbf{20.0}/\textbf{30.1}/\textbf{91.5}&\cellcolor{tab_others}98.7/12.5/22.8/89.9& \cellcolor{ours!50}98.9/10.1/19.5/90.7 & \cellcolor{ours}\textbf{99.2}/12.5/22.0/\textbf{91.5} \\
    pcb3  & 99.2/26.4/29.6/92.0&98.2/14.9/23.0/80.3&99.3/45.5/45.4/92.5&98.7/31.3/33.1/40.4&99.4/\textbf{48.1}/\textbf{47.3}/89.2& \cellcolor{ours!50}99.3/25.7/27.9/92.8 & \cellcolor{ours}\textbf{99.5}/28.1/30.4/\textbf{93.6} \\
    \cellcolor{tab_others}pcb4  &\cellcolor{tab_others}98.4/39.7/40.9/89.1&\cellcolor{tab_others}97.5/20.8/29.4/82.0&\cellcolor{tab_others}97.5/40.6/45.5/80.6&\cellcolor{tab_others}96.7/27.5/37.5/66.2&\cellcolor{tab_others}98.2/42.1/45.8/82.2& \cellcolor{ours!50}98.7/47.6/47.0/90.9 & \cellcolor{ours}\textbf{98.9}/\textbf{50.0}/\textbf{49.5}/\textbf{91.7} \\
    \midrule
    macaroni1 & 99.6/21.0/\textbf{31.1}/95.4&99.0/\;6.8/13.5/95.8&99.6/ \;7.0/11.1/\textbf{98.5}&98.7/17.3/25.3/53.4&99.7/ \;7.8/13.5/98.4& \cellcolor{ours!50}99.6/18.7/27.8/96.8 & \cellcolor{ours}\textbf{99.8}/\textbf{21.1}/30.3/97.6 \\
    \cellcolor{tab_others}macaroni2 & \cellcolor{tab_others}\textbf{99.5}/10.9/\textbf{18.7}/95.9&\cellcolor{tab_others}98.2/ \;3.8/10.3/95.3&\cellcolor{tab_others}98.4/ \;3.9/ \;8.8/94.8&\cellcolor{tab_others}99.0/ \;8.6/18.2/68.8&\cellcolor{tab_others}98.8/ \;4.9/ \;7.6/94.5& \cellcolor{ours!50}99.3/10.8/16.2/95.9 & \cellcolor{ours}\textbf{99.5}/\textbf{13.2}/\textbf{18.7}/\textbf{96.7} \\
    capsules &\textbf{99.6}/61.1/59.5/92.8&98.4/44.8/49.4/76.6&98.8/52.8/56.2/89.6&98.9/49.6/51.0/90.0&99.4/65.2/\textbf{65.6}/89.8& \cellcolor{ours!50}99.3/65.2/62.5/93.1 & \cellcolor{ours}99.5/\textbf{67.6}/65.0/\textbf{93.9} \\
    \cellcolor{tab_others}candle & \cellcolor{tab_others}98.9/22.6/33.7/94.5&\cellcolor{tab_others}98.9/16.7/26.2/94.9&\cellcolor{tab_others}98.4/13.1/22.2/90.2&\cellcolor{tab_others}99.0/\textbf{31.1}/\textbf{36.7}/78.7&\cellcolor{tab_others}\textbf{99.3}/18.7/27.2/\textbf{96.7}& \cellcolor{ours!50}98.7/19.9/30.5/95.1 & \cellcolor{ours}98.9/22.3/33.0/95.9 \\
    \midrule
    cashew & 95.4/53.0/56.2/89.8&\textbf{99.2}/64.6/64.3/85.5&98.9/62.3/62.6/78.5&99.0/\textbf{80.1}/\textbf{74.9}/89.2&98.7/58.8/60.0/\textbf{91.8}& \cellcolor{ours!50}97.5/58.9/59.3/86.9 & \cellcolor{ours}97.7/61.2/61.8/87.7 \\
    \cellcolor{tab_others}chewinggum & \cellcolor{tab_others}98.8/54.5/\textbf{59.1}/82.6&\cellcolor{tab_others}98.7/48.0/48.5/81.2&\cellcolor{tab_others}98.4/19.5/35.7/\textbf{86.0}&\cellcolor{tab_others}98.8/24.6/38.8/82.9&\cellcolor{tab_others}\textbf{98.9}/43.0/44.2/81.1& \cellcolor{ours!50}98.4/56.5/56.2/80.6 & \cellcolor{ours}98.7/\textbf{58.9}/58.7/81.4 \\
    fryum & 96.6/44.2/49.5/90.3&97.1/43.4/51.5/73.8&91.1/37.9/46.0/85.0&95.0/\textbf{51.8}/52.3/70.2&92.6/38.3/45.9/89.1& \cellcolor{ours!50}97.0/48.1/52.9/91.1 & \cellcolor{ours}\textbf{97.2}/50.5/\textbf{55.5}/\textbf{91.9} \\
    \cellcolor{tab_others}pipe\_fryum &\cellcolor{tab_others}99.0/51.0/55.5/93.7&\cellcolor{tab_others}99.3/56.9/63.7/90.7&\cellcolor{tab_others}98.9/59.6/60.4/91.5&\cellcolor{tab_others}\textbf{99.7}/\textbf{89.2}/\textbf{82.6}/50.7&\cellcolor{tab_others}98.9/58.7/58.8/\textbf{96.4}& \cellcolor{ours!50}98.9/51.1/54.7/94.0 & \cellcolor{ours}99.1/53.4/57.2/94.8 \\
    \midrule
    Mean  & 98.6/39.3/43.8/91.7&98.4/31.6/37.4/85.3&98.1/36.8/41.2/88.7&98.5/41.1/45.3/71.6&98.5/40.8/43.7/91.1& \cellcolor{ours!50}98.8/40.4/43.5/91.9 & \cellcolor{ours}\textbf{99.0}/\textbf{42.8}/\textbf{46.0}/\textbf{92.7} \\
    \bottomrule
    \end{tabular}
  }
  \label{tab:sgvisapx}%
\end{table*}
